\documentclass[preprint,12pt,authoryear]{elsarticle}
\usepackage{amssymb}
\usepackage{hyperref} 
\usepackage{lineno}
\usepackage{soul}
\usepackage{setspace}
\usepackage{subfig}
\usepackage{bm}
\usepackage{amssymb}
\usepackage{amsmath}
\usepackage{upgreek}
\usepackage{bigints}
\usepackage{algorithm2e}
\usepackage{algpseudocode}
\RestyleAlgo{ruled}
\usepackage{float}
\usepackage{xcolor}
\usepackage[left=2.7cm, right=2.7cm, top=3.3cm, bottom=3.3cm]{geometry}

\journal{Advances in Water Resources}
\begin{document}
\begin{frontmatter}

\title{Surrogate Model for Geological CO$_2$ Storage and Its Use in Hierarchical MCMC History Matching}

\author[inst1]{Yifu Han}
\affiliation[inst1]{organization={Department of Energy Science and Engineering}, 
            addressline={Stanford University}, 
            city={Stanford},
            state={CA},
            postcode={94305}, 
            country={USA}}

\author[inst2]{Fran\c cois P. Hamon}
\affiliation[inst2]{organization={TotalEnergies, E\&P Research and Technology},
            addressline={1201 Louisiana Street},
            city={Houston},
            state={TX},
            postcode={77002},
            country={USA}}

\author[inst1]{Su Jiang}
\author[inst1]{Louis J. Durlofsky}

\begin{abstract}
Deep-learning-based surrogate models show great promise for use in geological carbon storage operations. In this work we target an important application -- the history matching of storage systems characterized by a high degree of (prior) geological uncertainty. Toward this goal, we extend the recently introduced recurrent R-U-Net surrogate model to treat geomodel realizations drawn from a wide range of geological scenarios. These scenarios are defined by a set of metaparameters, which include the horizontal correlation length, mean and standard deviation of log-permeability, permeability anisotropy ratio, and constants in the porosity-permeability relationship. An infinite number of realizations can be generated for each set of metaparameters, so the range of prior uncertainty is large. The surrogate model is trained with flow simulation results, generated using the open-source simulator GEOS, for 2000 random realizations. The flow problems involve four wells, each injecting 1~Mt~CO$_2$/year, for 30~years. The trained surrogate model is shown to provide accurate predictions for new realizations over the full range of geological scenarios, with median relative error of 1.3\% in pressure and 4.5\% in saturation. The surrogate model is incorporated into a hierarchical Markov chain Monte Carlo history matching workflow, where the goal is to generate history matched geomodel realizations and posterior estimates of the metaparameters. We show that, using observed data from monitoring wells in synthetic `true' models, geological uncertainty is reduced substantially. This leads to posterior 3D pressure and saturation fields that display much closer agreement with the true-model responses than do prior predictions.

\end{abstract}


\end{frontmatter}

\section{Introduction}
\label{Introduction}
Geological carbon storage, if implemented at the gigatonne scale, has the potential to significantly reduce atmospheric CO$_2$ emissions. There are, however, many challenges associated with modeling and optimizing large-scale carbon storage operations. For example, because significant uncertainty exists in the detailed distributions of aquifer properties, model calibration is an essential step in the workflow. This calibration (also referred to as history matching or data assimilation), however, requires numerous time-consuming flow simulations, which must be repeated as new data are collected. Very fast data-driven surrogate models, used in place of these high-fidelity flow simulations, have been shown to provide accurate solutions at reduced computational cost for idealized cases. Their extension and application for more realistic scenarios, however, remains an open issue. 

In recent work, a deep-learning-based surrogate model that uses residual U-Nets within a recurrent neural network architecture was developed to predict flow and surface displacement in carbon storage models~\citep{tang2022deep}. This surrogate model treated 3D multi-Gaussian permeability realizations drawn from a single (specified) geological scenario. This means the geomodels were all characterized by the same scenario-defining quantities. Examples of scenario-defining quantities, also referred to as metaparameters, are the mean and standard deviation of porosity and log-permeability, the (spatial) correlation lengths in three orthogonal directions, the ratio of vertical to horizontal permeability, etc. In practice these metaparameters are themselves uncertain, so it may be important to include some of them as variables in the history matching procedure. In this study, we extend the recurrent R-U-Net surrogate model to treat geomodel realizations drawn from a range of geological scenarios. The surrogate model is then used with a hierarchical Markov chain Monte Carlo history matching procedure to reduce uncertainty in the metaparameters and thus in predicted pressure and saturation fields.

A number of deep-learning-based surrogate models have been developed to estimate flow quantities in carbon storage applications.~\citet{mo2019deep} introduced a data-driven deep convolutional encoder-decoder network to predict pressure and saturation for 2D Gaussian permeability fields.~\citet{wen2021ccsnet} presented a temporal convolutional neural network  to predict pressure and saturation for single-well axisymmetric injection scenarios.~\citet{wen2023real} later proposed a nested Fourier neural operator (FNO) for 3D problems. This architecture enables predictions of saturation and pressure in cases with permeability and porosity heterogeneity and anisotropy, under different reservoir conditions and injection-well configurations. \citet{yan2022robust} developed an FNO-based surrogate model to predict saturation and pressure, in both injection and post-injection periods, for new permeability and porosity fields and operational settings. \citet{yan2022gradient} proposed a physics-informed deep neural network model for CO$_2$ storage problems. They added flux terms into the loss function (minimized during training) to improve the resulting pressure fields. \citet{tang2022deep} extended earlier recurrent R-U-Net surrogate models~\citep{tang2020deep,tang2021deep} to treat coupled flow and geomechanics. Their surrogate provided predictions for flow variables in the storage aquifer, along with vertical displacement at the Earth's surface (for new realizations drawn from a single geological scenario).

A variety of history matching procedures have been applied in previous studies. \citet{chen2020reducing} incorporated pressure and saturation data from monitoring wells to reduce uncertainty in predictions of saturation and pressure at plugged/abandoned wells. \citet{tavakoli2013comparison} used pressure at the injection well and saturation at the monitoring well to reduce uncertainty in the location of the saturation plume. \citet{xiao2022model} used saturation fields interpreted from time-lapse seismic data to history match permeability. \citet{liu2020petrophysical} integrated injection well bottom-hole pressure, saturation at the monitoring wells, and time-lapse seismic data to reduce uncertainty in porosity and permeability fields and thus plume location. \citet{jung2015detection} used pressure data from both the storage aquifer and the caprock to detect potential leakage pathways. \citet{gonzalez2015detection} applied a restart ensemble Kalman filter algorithm with pressure measurements above the storage aquifer to identify caprock discontinuities. \citet{jahandideh2021inference} considered coupled flow and geomechanics, and used microseismic data to history match permeability and geomechanical parameters. It is important to note that none of studies mentioned above considered uncertainty in the geological scenario, i.e., in the metaparameters. This could be extremely expensive with these approaches because they are all simulation-based.

Surrogate models have been widely used for history matching and uncertainty quantification in many types of subsurface flow problems. \citet{wang2021efficient} developed a theory-guided neural network approach for efficient Monte Carlo-based uncertainty quantification in subsurface hydrology problems. \citet{zhou2021markov} proposed a surrogate-based adaptive MCMC method to reduce uncertainty in parameters characterizing the initial contaminant plume in groundwater problems. \citet{jiang2023use} proposed a transfer-learning framework that used low and high-fidelity simulation data to train their surrogate model. The surrogate was then applied in an ensemble-based history matching procedure in the context of oil reservoir simulation. Surrogate models have also been used for history matching in geological carbon storage problems. \citet{mo2019deep} introduced a surrogate model for uncertainty quantification in this setting. \citet{tang2022deep} incorporated their surrogate model with a rejection sampling-based history matching procedure to reduce uncertainty in aquifer pressure and surface displacement. Similarly,~\citet{2022deep} considered ensemble-based history matching, with surrogate-based flow predictions, to history match porosity and permeability. 

The main goals of this study are two-fold. First, we extend the recurrent R-U-Net surrogate model to treat geological realizations drawn from multiple geological scenarios, with each scenario characterized by a distinct set of metaparameters. Second, we use the surrogate model within a hierarchical and dimension-robust MCMC-based data assimilation workflow. We generate posterior (history matched) predictions for both the metaparameters and geological realizations. To our knowledge, this is the first hierarchical MCMC treatment involving multiple geological scenarios applied within the context of geological carbon storage. The metaparameters considered in this study include horizontal correlation length, mean and standard deviation of the log-permeability field, permeability anisotropy ratio, and constants in the porosity-permeability relationship. These are sampled from broad prior distributions to provide training realizations. Flow simulations, performed using the open-source simulator GEOS~\citep{bui2021multigrid}, provide the 3D pressure and saturation fields required for training. New input channels are introduced in the recurrent R-U-Net model to handle multiple geological scenarios. The MCMC method involves a noncentered preconditioned Crank-Nicolson method within a Gibbs algorithm~\citep{chen2018dimension}. In this approach, metaparameters and geological realizations are sampled sequentially at each iteration. Results for synthetic `true' models demonstrate the ability of the overall framework to provide substantial uncertainty reduction in metaparameters and flow predictions.

This paper proceeds as follows. The metaparameters and procedure to construct geomodels are presented in Section~\ref{Geomodels}. In Section~\ref{Surrogate Model}, we first discuss flow simulations using GEOS. The extended recurrent R-U-Net surrogate model is then described. In Section~\ref{MCMC}, our hierarchical MCMC-based history matching procedure is presented. Flow predictions for 3D models, and comparisons to high-fidelity simulation results, are provided in Section~\ref{Surrogate-performance}. The problem setup involves 4~megatonnes/year of CO$_2$ injection through four vertical wells. MCMC-based history matching results are presented in Section~\ref{History-matching}. We conclude in Section~\ref{Conclusions} with a summary and suggestions for future work. 

Additional results are provided in Supplementary Information (SI), available online. The results presented in SI include an assessment of surrogate model accuracy for second-order statistics, an evaluation of surrogate model performance and timing relative to an FNO method \citep{wen2023real}, a global sensitivity analysis \citep{sobol2001global, saltelli2010variance} quantifying the impact of the various metaparameters on monitoring well data, and additional history matching results for a different synthetic true model and for different prior distributions of the metaparameters.

\section{Geomodels and Metaparameters}
\label{Geomodels}
\begin{figure}[!ht]
\centering  
\includegraphics[width=12cm]{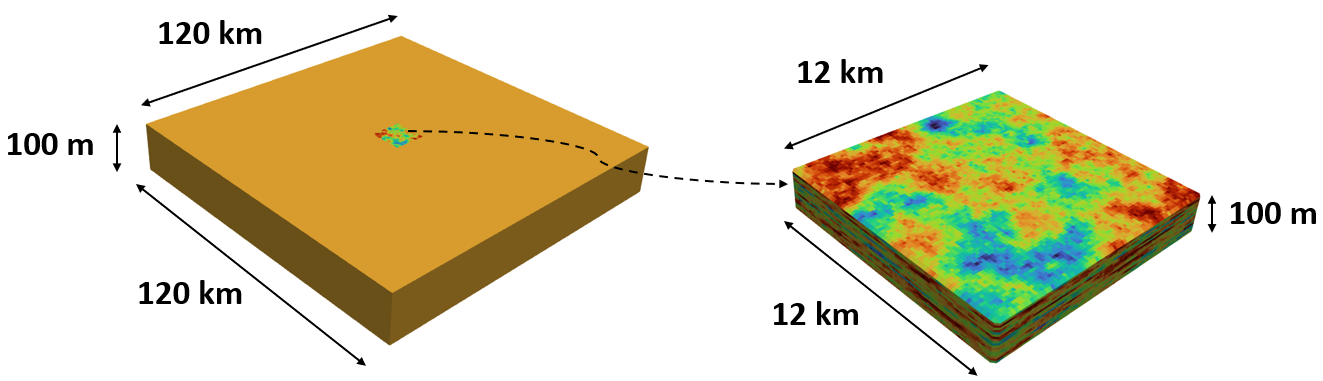}
\caption{Model domains for flow simulations. Full domain including surrounding region shown on the left, central storage aquifer shown on the right.}
\label{Model_Setup}
\end{figure}

The flow simulations are performed on an overall domain consisting of a storage aquifer and a surrounding region, as shown in Fig.~\ref{Model_Setup}. The dimensions of the full system are 120~km $\times$ 120~km $\times$ 100~m. The storage aquifer, of size 12~km $\times$ 12~km $\times$ 100~m, is at the center of the model. The storage aquifer is represented by 80 $\times$ 80 $\times$ 20 cells and the overall domain by 100 $\times$ 100 $\times$ 20 cells. The cells in the storage aquifer are 150~m $\times$ 150~m $\times$ 5~m. The cells in the surrounding domain increase in size (in $x$ and $y$) with distance from the storage aquifer. 

The geomodels of the full system are denoted by $\textbf{m}_f \in$ $\mathbb{R}^{n_c \times n_p}$, where $n_c$ = 200,000 is the number of cells in the overall domain, and $n_p$ = 3 is the number of rock properties in each cell (horizontal permeability, porosity, and permeability anisotropy ratio). The rock properties in the surrounding domain are taken to be constant. The geomodel realizations of the storage aquifer are drawn from a range of geological scenarios characterized by a set of six metaparameters. The metaparameters $\boldsymbol{\uptheta}_{\mathrm{meta}} \in \mathbb{R}^{6}$ are given by

\begin{equation} \label{metaparameters}
\boldsymbol{\uptheta}_{\mathrm{meta}} = [l_h, \mu_{\log k}, \sigma_{\log k}, a_r, d, e],
\end{equation}
\noindent where $k$ indicates the horizontal permeability, $l_h$ denotes the horizontal correlation length, $\mu_{\log k}$ and $\sigma_{\log k}$ are the mean and standard deviation of the log-permeability field, $a_r$ is the permeability anisotropy ratio ($a_r = k_v/k$, where $k_v$ is vertical permeability), and $d$ and $e$ are coefficients relating porosity to permeability. For any set of metaparameters, an infinite number of realizations can be generated.

The geomodels used for training and testing are generated using Gaussian sequential simulation in the geological modeling software SGeMS~\citep{remy2009applied}. We use an exponential variogram model characterized by horizontal ($l_x$ and $l_y$) and vertical ($l_z$) correlation lengths. In this study, the vertical correlation length ($l_v=l_z$) is set to 10~m (2~cells), and a discrete set of values for $l_h=l_x=l_y$ is considered. We fix $l_v$ while considering a range of $l_h$ because $l_v$ can often be estimated directly from well-log data. The value of $l_h$, by contrast, can be more challenging to determine. In this study, in surrogate model training and testing, we consider $l_h$ values of 1500~m, 2250~m, 3000~m, 3750~m and 4500~m, which correspond to 10, 15, 20, 25 and 30~cells. However, in numerical tests in which monitoring wells were placed relatively close to injection wells, measured data were found not to be informative regarding the value of $l_h$. Therefore, in our history matching examples, we consider the last five metaparameters in Eq.~\ref{metaparameters}, and take $l_h$ (and $l_v$) as fixed. These could also be considered as metaparameters (and included in the history matching) if necessary.

The permeability fields of the storage aquifer are denoted by $\textbf{k}_s \in \mathbb{R}^{n_s}$, where $n_s$ = 128,000 is the number of cells in the storage aquifer. These fields are constructed from standard multi-Gaussian fields 
$\textbf{y}^{\mathrm{G}} \in$ $\mathbb{R}^{n_s}$ and the metaparameters, $\mu_{\mathrm{log}k}$ and $\sigma_{\mathrm{log}k}$. During the history matching (online) computations, these multi-Gaussian fields are generated using principal component analysis (PCA). This is faster and more convenient than using geological modeling software. Specifically, $O(10^5)$ geomodels must be generated and evaluated with the MCMC-based history matching procedure applied here. Constructing a standard multi-Gaussian realization using SGeMS requires about 0.64~seconds (including startup) on a single AMD EPYC-7543 CPU. Generating a PCA realization, by contrast, requires only 0.01~seconds with the same CPU (speedup factor of 64). In addition, the use of PCA eliminates the need to embed SGeMS in the online workflow. Given the PCA representation and metaparameters, the full geomodel can be formed. 

The PCA basis matrix, denoted $\Phi$, is constructed by performing singular value decomposition on a set of 1000 centered SGeMS realizations. The left singular vectors and singular values provide the basis matrix $\Phi$. The metaparameters act to shift and rescale the standard multi-Gaussian fields provided by PCA. New PCA realizations of standard multi-Gaussian fields, denoted $\textbf{y}^{\mathrm{pca}} \in \mathbb{R}^{n_s}$, can be generated through application of 

\begin{equation} \label{pca}
\textbf{y}^{\mathrm{pca}} = \Phi \bm{\xi} + \bar{\textbf{y}}, 
\end{equation} 
\noindent where $\Phi \in \mathbb{R}^{n_s \times n_d}$ is the basis matrix truncated to $n_d$ columns, $\bar{\textbf{y}} \in \mathbb{R}^{n_s}$ is the mean of the SGeMS realizations used to generate $\Phi$, and $\bm{\xi} \in \mathbb{R}^{n_d}$ is the low-dimensional standard-normal variable. Here $n_d$ is set to 900, which preserves about 95\% of the total `energy' present in the original set of 1000 SGeMS realizations (energy here is computed as the sum of the squared singular values). For more details on PCA basis matrix construction, please see~\citet{liu20213d}.

For a given cell $i$ in the storage aquifer, the permeability $(k_s)_i$ is given by

\begin{equation} \label{log_permeability}
(k_{s})_i = \exp \left( \sigma_{\log k} \cdot (y^{\mathrm{G}})_i + \mu_{\log k} \right), \ \  i = 1, 2, \dots, n_s.
\end{equation}

The porosity field in the storage aquifer, denoted $\boldsymbol{\phi}_s \in \mathbb{R}^{n_s}$, is generated directly from the permeability field via the metaparameters $d$ and $e$. The porosity-permeability relationship is expressed as

\begin{equation} \label{porosity}
(\phi_s)_i = d \cdot \mathrm{log}\left((k_s)_i\right) + e, \ \  i = 1, 2, \dots, n_s.
\end{equation}

The geomodel for the storage aquifer, $\textbf{m}_s \in$ $\mathbb{R}^{n_s \times n_p}$, is defined by the permeability field, the porosity field, and the permeability anisotropy ratio. These models are expressed as

\begin{equation} \label{geomodel of storage aquifer}
\textbf{m}_s = [\textbf{k}_s, \bm{\phi}_s, {\bf{a}}_r], 
\end{equation}
\noindent where ${\bf{a}}_r \in \mathbb{R}^{n_s}$ defines the permeability anisotropy ratio in every storage aquifer cell. In this study, we treat the anisotropy ratio as an uncertain metaparameter, which is taken to be constant throughout the storage aquifer.

\section{Deep-learning-based Surrogate Model}
\label{Surrogate Model}
The high-fidelity flow simulations are performed using open-source simulator GEOS \citep{bui2021multigrid}. GEOS is able to model coupled flow and geomechanics, though the simulation times are considerably longer if geomechanical effects are included (which is why we consider only flow in this study). For the governing equations for CO$_2$-water flow simulations, please refer to~\citet{tang2022deep}.

The flow simulations can be represented as

\begin{equation} \label{forward}
[\textbf{P}_f, \enspace \textbf{S}_f] = f(\mathbf{m}_f),
\end{equation}

\noindent where $f$ denotes the high-fidelity flow simulation, and $\textbf{P}_f \in \mathbb{R}^{n_c \times n_{ts}}$ and $\textbf{S}_f \in \mathbb{R}^{n_c \times n_{ts}}$ are the pressure and saturation in each cell of the overall domain at $n_{ts}$ simulation time steps. The flow simulations are performed for a wide range of geomodels $\mathbf{m}_f$, with the well locations and operational settings the same in all runs. Specifically, in this study supercritical CO$_2$ is injected through four vertical wells, with each well injecting 1~megatonne (Mt) CO$_2$ per year for 30~years.

We now describe the extended recurrent R-U-Net surrogate model used to approximate the flow variables for new geomodels. The surrogate models are trained to predict saturation and pressure fields only in the storage aquifer (not in the full domain), and only at particular time steps. Separate surrogate models are used for the pressure and saturation fields. In analogy to Eq.~\ref{forward}, the surrogate model predictions can be expressed as 

\begin{equation} \label{surrogate_forward}
[\hat{\textbf{P}}_s, \enspace \hat{\textbf{S}}_s] = \hat{f}(\mathbf{m}_s; \textbf{w}),
\end{equation}

\noindent where $\hat{f}$ denotes the surrogate model, $\hat{\textbf{P}}_s \in \mathbb{R}^{n_s \times n_{t}}$ and $\hat{\textbf{S}}_s \in \mathbb{R}^{n_s \times n_{t}}$ are surrogate model predictions for pressure and saturation in the storage aquifer, $n_t$ is the number of surrogate model time steps, and $\textbf{w}$ represents the deep neural network parameters determined during training. In this work we take $n_t = 10$. This is considerably less than $n_{ts}$, though it is sufficient to describe the key dynamics of the storage operation.

\begin{figure}[!ht]
\centering   
\subfloat[3D residual U-Net architecture]{\label{fig:a}\includegraphics[width = 132mm]{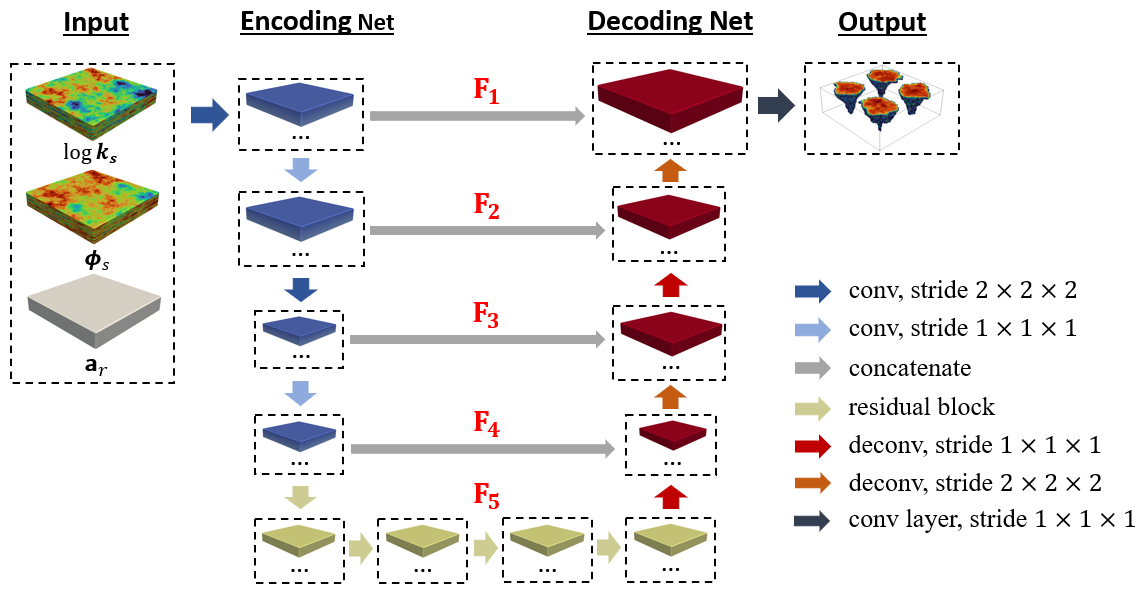}}\\[5ex]
\subfloat[Recurrent R-U-Net architecture]{\label{fig:b}\includegraphics[width = 132mm]{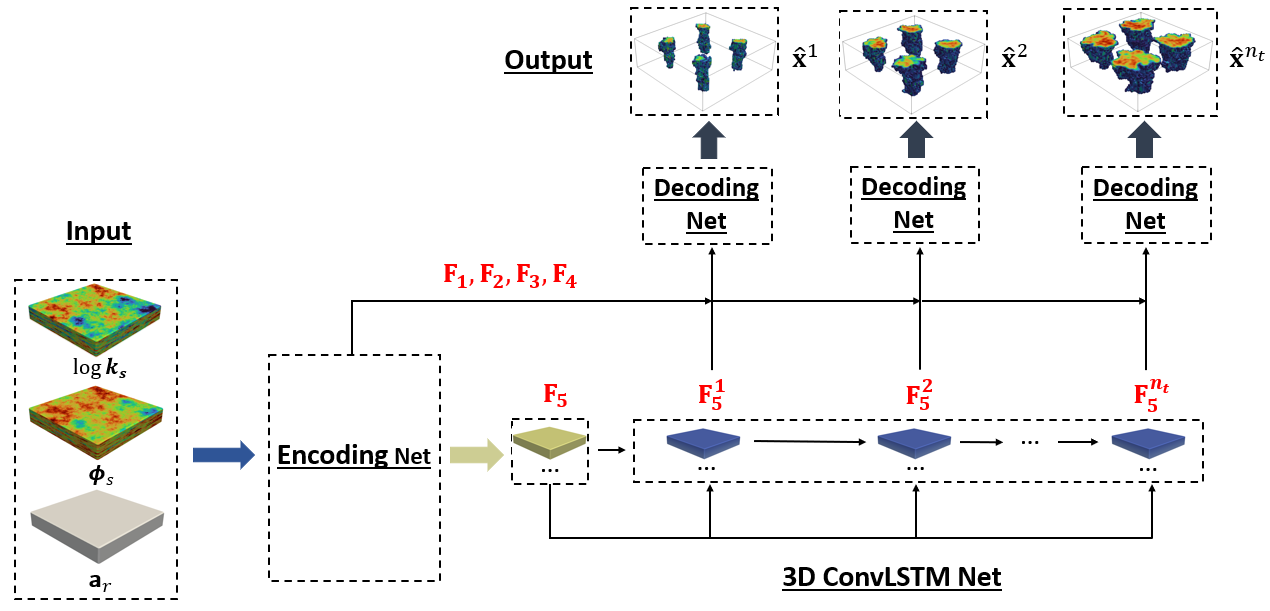}}
\caption{Schematic of the 3D extended recurrent R-U-Net architecture. 3D R-U-Net architecture for pressure and saturation prediction at a specific time step is shown in (a). Incorporation of 3D R-U-Net within recurrent neural network shown in (b). Geomodels are defined by three input channels.}
\label{nn_model}
\end{figure}

The surrogate model, which represents an extension of the recurrent R-U-Net developed by \citet{tang2022deep}, contains two main components -- a residual U-Net and a recurrent neural network. The residual U-Net, shown in Fig.~\ref{nn_model}(a), consists of encoding and decoding networks. The encoding network accepts three input channels representing the geomodel. These are transformed to a sequence of low-dimensional latent feature maps, represented as \textbf{F}$_1$, \textbf{F}$_2$, $\dots$, \textbf{F}$_5$. In the decoding network, these latent feature maps are combined with upsampled feature maps through a concatenation process. This provides predictions for the pressure and saturation fields at a specific time step.

To capture the temporal dynamics, a convolutional long short-term memory (convLSTM) recurrent neural network is incorporated. This network, shown in Fig.~\ref{nn_model}(b), sequentially applies a time sequence of the most compressed (global) feature maps, denoted \textbf{F}$_5^1$, $\dots$, \textbf{F}$_5^{n_t}$, through the convLSTM module. The decoding network of the overall surrogate model combines the feature maps from the encoding network (\textbf{F}$_1$ -- \textbf{F}$_4$) with each of the \textbf{F}$_5$ feature maps to provide pressure and saturation predictions at $n_t$ time steps. The latent feature maps \textbf{F}$_1$ -- \textbf{F}$_5$ capture the spatial characteristics of the geomodel with varying levels of fidelity. These range from local features in \textbf{F}$_1$ to global features in \textbf{F}$_5$. The convLSTM layer is applied to the most compressed features \textbf{F}$_5$ to capture global temporal evolution. The spatial features represented by the \textbf{F}$_1$ -- \textbf{F}$_4$ feature maps are incorporated into the decoding networks to improve the accuracy of spatial predictions. By treating the feature maps in this way, a reduction in the number of network parameters is achieved. This leads to faster surrogate model training and prediction, while retaining a high level of accuracy.

The original surrogate model developed by \citet{tang2022deep} treated geomodels drawn from a single geological scenario that were fully characterized by a single parameter in each cell. Thus only a single input channel was required. Here we consider a range of geological scenarios with geomodels characterized by three values in each grid block ($k_i$, $\phi_i$, $(a_r)_i$). Thus, three input channels (evident in Fig.~\ref{nn_model}) and a modified architecture are required. Input channels that differ from those used here could also be considered. The log-permeability field is related to the porosity field through metaparameters $d$ and $e$. Deep neural networks can capture and leverage correlations between input features, and layers within the networks can identify and utilize patterns in the input data. Thus it is possible that other input channels/formats, such as the use of log-permeability, $d$ and $e$ in place of log-permeability and porosity, might lead to improved accuracy or efficiency. It will be useful to test different approaches in future work.

In the network shown in Fig.~\ref{nn_model}, in the first convolutional layer, filters are applied to process all three input channels simultaneously. This provides latent feature maps \textbf{F}$_1$ that contain the combined features from the three input fields. These \textbf{F}$_1$ maps are (again) downsampled in the encoding network into a sequence of low-dimensional latent feature maps \textbf{F}$_2$, $\dots$, \textbf{F}$_5$. The recurrent component of the network is essentially unchanged, again generating a time sequence of latent feature maps. Details of the extended recurrent R-U-Net architecture are provided in Table~\ref{nn-architecture}. Recall that, in our storage aquifer models, $n_x=n_y=80$ and $n_z=20$, where $n_x$, $n_y$ and $n_z$ are the number of cells in each coordinate direction.

\begin{table}[!ht]
\footnotesize
\begin{center}
\caption{Architecture of the extended recurrent R-U-Net}
\label{nn-architecture}
\renewcommand{\arraystretch}{1.15} 
\begin{tabular}{ c c c } 
\hline
\textbf{Network} & \textbf{Layer}  & \textbf{Output}\\ 
\hline
Encoder & Input & (n$_{x}$, n$_{y}$, n$_{z}$, 3) \\ 
 & conv, 16 filters of size $3 \times 3 \times 3$, stride 2 & ($\frac{n_x}{2}, \frac{n_y}{2}, \frac{n_z}{2}$, 16) \\ 
 & conv, 32 filters of size $3 \times 3 \times 3$, stride 1 & ($\frac{n_x}{2}, \frac{n_y}{2}, \frac{n_z}{2}$, 32) \\
 & conv, 32 filters of size $3 \times 3 \times 3$, stride 2 & ($\frac{n_x}{4}, \frac{n_y}{4}, \frac{n_z}{4}$, 32) \\
 & conv, 64 filters of size $3 \times 3 \times 3$, stride 1 & ($\frac{n_x}{4}, \frac{n_y}{4}, \frac{n_z}{4}$, 64) \\
 & residual block, 64 filters of size $3 \times 3 \times 3$, stride 1 & ($\frac{n_x}{4}, \frac{n_y}{4}, \frac{n_z}{4}$, 64) \\
 & residual block, 64 filters of size $3 \times 3 \times 3$, stride 1 & ($\frac{n_x}{4}, \frac{n_y}{4}, \frac{n_z}{4}$, 64) \\
\hline
ConvLSTM & convLSTM3D, 64 filters of size $3 \times 3 \times 3$, stride 1 & ($\frac{n_x}{4}, \frac{n_y}{4}, \frac{n_z}{4}$, 64, n$_{t}$) \\
\hline
Decoder & residual block, 64 filters of size $3 \times 3 \times 3$, stride 1 & ($\frac{n_x}{4}, \frac{n_y}{4}, \frac{n_z}{4}$, 64, n$_{t}$) \\
 & residual block, 64 filters of size $3 \times 3 \times 3$, stride 1 & ($\frac{n_x}{4}, \frac{n_y}{4}, \frac{n_z}{4}$, 64, n$_{t}$) \\
 & deconv, 64 filters of size $3 \times 3 \times 3$, stride 1 & ($\frac{n_x}{4}, \frac{n_y}{4}, \frac{n_z}{4}$, 64, n$_{t}$) \\
 & deconv, 32 filters of size $3 \times 3 \times 3$, stride 2 & ($\frac{n_x}{2}, \frac{n_y}{2}, \frac{n_z}{2}$, 32, n$_{t}$) \\
 & deconv, 32 filters of size $3 \times 3 \times 3$, stride 1 & ($\frac{n_x}{2}, \frac{n_y}{2}, \frac{n_z}{2}$, 32, n$_{t}$) \\
 & deconv, 16 filters of size $3 \times 3 \times 3$, stride 2 & (n$_{x}$, n$_{y}$, n$_{z}$, 16, n$_{t}$) \\
 & conv layer, 1 filters of size $3 \times 3 \times 3$, stride 1 & (n$_x$, n$_y$, n$_z$, 1, n$_{t}$) \\
\hline
\end{tabular}
\end{center}
\end{table}

We now describe the training procedure. As noted earlier, separate networks are trained for pressure and saturation. Scaling and normalization is required prior to training. The log-permeability, porosity, and permeability anisotropy ratio input fields are normalized by scaling based on their respective maximum values. Saturation values are already in the (0, 1) range, so normalization is not required. A (time-dependent) min-max normalization procedure is applied for pressure. The normalized pressure is computed using

\begin{equation} \label{normalization}
(\bar{p}_s)_{i,j}^t = \frac{(p_s)_{i,j}^t - (p_s)_{i,j}^0}{(p_s)_{\mathrm{max}}^t - (p_s)_{\mathrm{min}}^t}, 
\end{equation}

\noindent where $(p_s)_{i,j}^t$, $(\bar{p}_s)_{i,j}^t$ and $(p_s)_{i,j}^0$ are the pressure, normalized pressure and initial hydrostatic pressure of cell $j$ in geomodel $i$, at time step $t$, and $(p_s)_{\mathrm{max}}^t$ and $(p_s)_{\mathrm{min}}^t$ are the maximum and minimum pressure in the storage aquifer at time step $t$ over all pressure fields in all training runs.

The training procedure entails determining optimal network parameters, denoted $\textbf{w}^{\ast}$, that minimize the difference between flow simulation results and surrogate model predictions. Flow simulations are performed on geomodels sampled from a range of geological scenarios, as explained above. The minimization problem can be stated as

\begin{equation} \label{minimization}
\textbf{w}^{\ast} = \operatorname*{argmin}_{\textbf{w}} \frac{1}{n_{smp}} \frac{1}{n_t} \sum_{i=1}^{n_{smp}}\sum_{t=1}^{n_t}\|\hat{\textbf{x}}_{i}^{t} - \textbf{x}_{i}^{t}\|_2^2,
\end{equation}

\noindent where $n_{smp}$ is the number of training samples (runs), and $\textbf{x}_{i}^{t}$ and $\hat{\textbf{x}}_{i}^{t}$ represent the normalized pressure or saturation from high-fidelity flow simulation and the surrogate model at time step $t$ for training sample $i$, respectively. In this study we set $n_{smp}=2000$. The training procedure for pressure requires 8~hours (corresponding to 300~epochs) to converge, while training for saturation requires about 13~hours (500~epochs) on a single Nvidia A100 GPU. For both networks a batch size of 4 is used, and the initial learning rate is set to 0.003.

\section{Hierarchical MCMC History Matching Procedure}
\label{MCMC}
The geomodels of the storage aquifer constructed in the history matching workflow are based on PCA representations of standard multi-Gaussian realizations and metaparameters. As discussed in Section~\ref{Geomodels}, the measured data considered in the history matching are not sensitive to horizontal correlation length $l_h$. In addition, in the MCMC procedure we now describe, it is simpler to neglect $l_h$ and consider only the last five metaparameters appearing in Eq.~\ref{metaparameters}. More specifically, the treatment of multiple $l_h$ values would require multiple (separate) PCA representations along with modifications to the hierarchical MCMC algorithm. Therefore, in the history matching, we set $\boldsymbol{\uptheta}_{\mathrm{meta}} = [\mu_{\log k}, \sigma_{\log k}, a_r, d, e]$.

The probability density function for $\boldsymbol{\uptheta}_{\mathrm{meta}}$ and the PCA latent variables $\bm{\xi}$, conditioned to observation data, is given by Bayes' theorem:
\begin{equation} \label{Bayes}
  p({\boldsymbol{\uptheta}_{\mathrm{meta}}, \bm{\xi} | \bm{\mathrm{d}}_{\mathrm{obs}}}) = \frac{p(\boldsymbol{\uptheta}_{\mathrm{meta}}, \bm{\xi}) p(\bm{\mathrm{d}}_{\mathrm{obs}}|{\boldsymbol{\uptheta}}_{\mathrm{meta}}, \bm{\xi})} {p(\bm{\mathrm{d}}_{\mathrm{obs}})}.
  \end{equation}

\noindent Here $\bm{\mathrm{d}}_{\mathrm{obs}}$ represents the observation data, $p({\boldsymbol{\uptheta}_{\mathrm{meta}}, \bm{\xi}|\bm{\mathrm{d}}_{\mathrm{obs}}})$ is the posterior probability density function of metaparameters and latent variables, $p(\boldsymbol{\uptheta}_{\mathrm{meta}}, \bm{\xi})$ is the prior probability density function, $p(\bm{\mathrm{d}}_{\mathrm{obs}} | {\boldsymbol{\uptheta}}_{\mathrm{meta}}, \bm{\xi})$ is the likelihood function, and ${p(\bm{\mathrm{d}}_{\mathrm{obs}})}$ is a normalization constant. The determination of the normalization constant would require integrating over the entire parameter space. This expensive calculation is not required in MCMC sampling, as the algorithm enables the estimation of $p({\boldsymbol{\uptheta}_{\mathrm{meta}}, \bm{\xi}|\bm{\mathrm{d}}_{\mathrm{obs}}})$ without the need to compute the normalization constant. 

MCMC sampling is a rigorous method for obtaining samples from posterior distributions. It involves simulating a Markov chain to generate a sequence of samples. With enough samples, the posterior distributions can be approximated accurately. The estimation of posterior distributions for both the metaparameters and their associated random fields is referred to as a hierarchical Bayes problem~\citep{malinverno2004expanded}. In one such application, \citet{xiao2021bayesian} presented a preconditioned Crank-Nicolson hierarchical MCMC method to generate posteriors of metaparameters and multi-Gaussian log-conductivity fields for hydrology problems.

In this study, we apply the noncentered preconditioned Crank-Nicolson within Gibbs algorithm developed by \citet{chen2018dimension}. This hierarchical and dimension-robust MCMC method uses the Metropolis-within-Gibbs~\citep{gamerman2006markov} treatment to generate posterior distributions for metaparameters and latent variables that are independent (as is the case here). The starting set of metaparameters ${\boldsymbol{\uptheta}}_{\mathrm{meta}}^{1}$ is sampled from the prior distributions. Each component of the initial set of latent variables $\bm{\xi}^1 \in \mathbb{R}^{n_d}$ is sampled independently from the standard normal distribution $\mathcal{N}$(0, 1).

The details of the hierarchical and dimension-robust MCMC-based history matching workflow are shown in Algorithm~\ref{mcmc-procedure}. At iteration $k+1$ of the Markov chain, we generate a new sample of latent variables $\bm{\xi}^{'}$ by adding a random perturbation $\boldsymbol{\epsilon} \in \mathbb{R}^{n_d}$ to the latent variables $\bm{\xi}^{k}$ at the previous iteration $k$. This new sample of latent variables $\bm{\xi}^{'}$ is given by

\begin{equation} \label{xi}
\bm{\xi}^{'} = (1 - \beta^2) \bm{\xi}^{k} + \beta \boldsymbol{\epsilon},
\end{equation}

\noindent where each component of $\boldsymbol{\epsilon} \in \mathbb{R}^{n_d}$ is sampled independently from $\mathcal{N}$(0, 1), and $\beta$ is a tuning coefficient that controls the magnitude of the update in the new sample. In accordance with~\cite{chen2018dimension}, $\beta$ is adjusted to achieve a reasonable acceptance rate. In this work, we set $\beta= 0.15$, resulting in an acceptance rate of about 20\%.

The Metropolis-Hastings~\citep{hastings1970monte} acceptance criterion is then applied. The probability of acceptance for the proposed latent variable $\bm{\xi}^{'}$ is given by

\begin{equation} \label{latent_probability}  
  \alpha\left(\bm{\xi}^{k}, \bm{\xi}^{'}\right)  = \mathrm{min}\left( 1,  \frac{p(\bm{\mathrm{d}}_{\mathrm{obs}}|{\boldsymbol{\uptheta}}_{\mathrm{meta}}^{k}, \bm{\xi}^{'})}{p(\bm{\mathrm{d}}_{\mathrm{obs}}|{{\boldsymbol{\uptheta}}_{\mathrm{meta}}^{k}}, \bm{\xi}^{k})} \right),
\end{equation}

\noindent where $p(\bm{\mathrm{d}}_{\mathrm{obs}} | {{\boldsymbol{\uptheta}}_{\mathrm{meta}}^{k}}, \bm{\xi}^{k})$ and $p(\bm{\mathrm{d}}_{\mathrm{obs}} | {{\boldsymbol{\uptheta}}_{\mathrm{meta}}^{k}}, \bm{\xi}^{'})$ are the likelihoods of the current ($\bm{\xi}^{k}$) and the proposed ($\bm{\xi}^{'}$) latent variables, conditioned to the metaparameters ${{\boldsymbol{\uptheta}}_{\mathrm{meta}}^{k}}$ (from the last iteration $k$). The likelihoods are computed from the mismatch between observed and predicted data (the detailed expression will be given in Section~\ref{History matching setup}).

Next, a new sample of metaparameters ${\boldsymbol{\uptheta}}_{\mathrm{meta}}^{'}$ is proposed by adding a random perturbation to each metaparameter in the current sample ${\boldsymbol{\uptheta}}_{\mathrm{meta}}^{k}$. With the Metropolis-within-Gibbs treatment, the proposal distribution is a multivariate Gaussian distribution centered on the current sample, i.e., $\mathcal{N}$(${\boldsymbol{\uptheta}}_{\mathrm{meta}}^{k}$, $C_{\uptheta}$), where $C_{\uptheta} \in \mathbb{R}^{5 \times 5}$ is the covariance matrix of the proposal distribution for the metaparameters. Large standard deviations enable fast exploration of the metaparameter space at the cost of low acceptance rates~\citep{gelman1997weak, andrieu2008tutorial}. Small standard deviations, by contrast, can increase the acceptance rate but slow the exploration of the metaparameter space. This could also result in highly correlated samples and poor convergence~\citep{gelman1997weak}. 

In this study, for metaparameter $i$ ($i = 1, \dots, 5$), we specify the standard deviation as $\sigma_i=(({{\uptheta}}_{\mathrm{meta}})_{i, \mathrm{max}}-({{\uptheta}}_{\mathrm{meta}})_{i, \mathrm{min}})/16$, where $({{\uptheta}}_{\mathrm{meta}})_{i, \mathrm{max}}$ and $({{\uptheta}}_{\mathrm{meta}})_{i, \mathrm{min}}$ are the maximum and minimum values of the prior range. This approach provides acceptance rates that fall within the range of 10\% to 40\%~\citep{gelman1996efficient}. The probability of acceptance for the proposed metaparameters is given by

\begin{equation} \label{probability}  
  \alpha\left({\boldsymbol{\uptheta}}_{\mathrm{meta}}^{k}, {\boldsymbol{\uptheta}}_{\mathrm{meta}}^{'}\right)  = \mathrm{min}\left( 1, \frac{p({\boldsymbol{\uptheta}}_{\mathrm{meta}}^{'})}{p({\boldsymbol{\uptheta}}_{\mathrm{meta}}^{k})} \frac{p(\bm{\mathrm{d}}_{\mathrm{obs}}|{\boldsymbol{\uptheta}}_{\mathrm{meta}}^{'}, \bm{\xi}^{k+1})}{p(\bm{\mathrm{d}}_{\mathrm{obs}}|{{\boldsymbol{\uptheta}}_{\mathrm{meta}}^{k}}, \bm{\xi}^{k+1})} \right),
\end{equation} 

\noindent where $p(\bm{\mathrm{d}}_{\mathrm{obs}} | {{\boldsymbol{\uptheta}}_{\mathrm{meta}}^{k}}, \bm{\xi}^{k+1})$ and $p(\bm{\mathrm{d}}_{\mathrm{obs}} | {{\boldsymbol{\uptheta}}_{\mathrm{meta}}^{'}}, \bm{\xi}^{k+1})$ are the likelihood of the current (${{\boldsymbol{\uptheta}}_{\mathrm{meta}}^{k}}$) and proposed  (${{\boldsymbol{\uptheta}}_{\mathrm{meta}}^{'}}$) metaparameters conditioned to $\bm{\xi}^{k+1}$, and $p({{\boldsymbol{\uptheta}}_{\mathrm{meta}}^{k}})$ and $p({{\boldsymbol{\uptheta}}_{\mathrm{meta}}^{'}})$ are their prior probabilities. In this study, we consider both uniform and Gaussian prior distributions of the metaparameters. History matching results for Gaussian-distributed priors are presented in SI.

By iteratively repeating this hierarchical sampling procedure, a set of samples that approximates the posterior distributions can be generated. MCMC may require tens of thousands of function evaluations to evaluate the likelihoods, which would be computationally prohibitive if high-fidelity flow simulations were used. With the surrogate model used instead of high-fidelity simulation, however, the computational cost is manageable. 

\begin{algorithm}
\DontPrintSemicolon 
\small
\caption{Hierarchical MCMC-based history matching procedure}\label{mcmc-procedure}
 1.~Set iteration counter $k$ = 1, and initialize the Markov chain with $\bm{\xi}^1$ and ${\boldsymbol{\uptheta}}_{\mathrm{meta}}^{1}$.\;
 \Repeat {\normalfont Convergence criterion is satisfied.}{
  2.~Propose a new sample for the PCA latent variables ($\bm{\xi}^{'}$) using Eq.~\ref{xi}.\; 
  \;
  3.~Construct PCA realization of multi-Gaussian field $\textbf{y}^{\mathrm{pca}}(\bm{\xi}^{'})$ using Eq.~\ref{pca}, and geomodel $\textbf{m}_s\left({\boldsymbol{\uptheta}}_{\mathrm{meta}}^{k}, \textbf{y}^{\mathrm{pca}}(\bm{\xi}^{'})\right)$ based on the metaparameters ${\boldsymbol{\uptheta}}_{\mathrm{meta}}^{k}$, and newly constructed PCA realization $\textbf{y}^{\mathrm{pca}}(\bm{\xi}^{'})$ using Eqs.~\ref{log_permeability}, ~\ref{porosity} and~\ref{geomodel of storage aquifer}.\;
  \;
  4.~Predict pressure and saturation at $n_t$ time steps using the surrogate model, through application of ${\mathbf d}^{'} = \hat{f}\left(\textbf{m}_s\left({\boldsymbol{\uptheta}}_{\mathrm{meta}}^{k}, \textbf{y}^{\mathrm{pca}}(\bm{\xi}^{'})\right)\right)$. Compute likelihood $p(\bm{\mathrm{d}}_{\mathrm{obs}} | {{\boldsymbol{\uptheta}}_{\mathrm{meta}}^{k}}, \bm{\xi}^{'})$ for the proposed latent variables based on mismatch between observed and predicted data (expressions given later). \;
  \;
  5.~Set $\bm{\xi}^{k+1} = \bm{\xi}^{'}$ with probability, $\alpha(\bm{\xi}^{k}, \bm{\xi}^{'})$, computed using Eq.~\ref{latent_probability}. \;
  \;
  6.~Propose a new sample for the metaparameters (${{\uptheta}}_{\mathrm{meta}}^{'}$) from a multivariate Gaussian proposal distribution, $\mathcal{N}$(${\boldsymbol{\uptheta}}_{\mathrm{meta}}^{k}$, $C_{\uptheta}$), centered on current sample ${\boldsymbol{\uptheta}}_{\mathrm{meta}}^{k}$. \;
  \;
  7.~Construct geomodel $\textbf{m}_s\left({\boldsymbol{\uptheta}}_{\mathrm{meta}}^{'}, \textbf{y}^{\mathrm{pca}}(\bm{\xi}^{k+1})\right)$ based on the newly proposed metaparameters ${\boldsymbol{\uptheta}}_{\mathrm{meta}}^{'}$ and latent variables $\bm{\xi}^{k+1}$ using Eqs.~\ref{log_permeability}, ~\ref{porosity} and~\ref{geomodel of storage aquifer}. \;
  \;
  8.~Predict pressure and saturation at $n_t$ time steps using the surrogate model, through application of ${\mathbf d}^{'} = \hat{f}\left(\textbf{m}_s\left({\boldsymbol{\uptheta}}_{\mathrm{meta}}^{'}, \textbf{y}^{\mathrm{pca}}(\bm{\xi}^{k+1})\right)\right)$. Compute likelihood $p(\bm{\mathrm{d}}_{\mathrm{obs}} | {{\boldsymbol{\uptheta}}_{\mathrm{meta}}^{'}}, \bm{\xi}^{k+1})$ for the proposed metaparameters. \;
  \;
  9.~Set ${\boldsymbol{\uptheta}}_{\mathrm{meta}}^{k+1}={\boldsymbol{\uptheta}}_{\mathrm{meta}}^{'}$ with probability,  $\alpha({\boldsymbol{\uptheta}}_{\mathrm{meta}}^{k}, {\boldsymbol{\uptheta}}_{\mathrm{meta}}^{'})$, computed using Eq.~\ref{probability}. \;
  \;
  Increment the counter $k = k + 1$.\;
  \;
 }
\end{algorithm}

MCMC sampling results in the initial `burn-in' period must be discarded to ensure that the initial state and transient behavior of the Markov chain do not bias the predictions. In this work, the burn-in period is taken to be 5000 iterations, which falls within the ranges suggested by~\citet{kruschke2015markov} and~\citet{ fang2020deep} for practical applications. In addition, a convergence criterion based on the average relative change of the bin-wise posterior probability density~\citep{ARMA-2022-0774} is incorporated into the algorithm. The bin-wise posterior probability density refers to the posterior probability density computed on discrete bins (10~bins in this study) or intervals of the metaparameter values. The posterior distributions of metaparameters are considered to be converged when the average relative change of the bin-wise posterior probability density falls below a threshold of 0.01.

\section{Surrogate Model Evaluation} 
\label{Surrogate-performance}
In this section, we first describe the geomodel and simulation setup. We then present aggregate error statistics for (new) test cases, spanning multiple geological scenarios, using the extended recurrent R-U-Net surrogate model. Detailed results for specific cases are then presented. 

\subsection{Geomodel and simulation setup}
\label{Model Setup}

Some of the geomodel properties considered in this study are based on In Salah ~\citep{white2014geomechanical, li2016coupled}, an onshore storage project in Algeria that was active from 2004-2011. Here, however, we consider a larger overall system with much higher CO$_2$ injection rates, as our interest is in modeling future industrial-scale operations. More specifically, at In Salah, 3.8~Mt of CO$_2$ were injected over a seven-year injection period~\citep{li2016coupled} (an average of 0.54~Mt/year), while in this study we consider injection rates of 4~Mt/year, with injection proceeding over 30~years.

As noted in Section~\ref{Geomodels}, the overall system is of physical dimensions 120~km $\times$ 120~km $\times$ 100~m. The full model is represented on a grid containing 100 $\times$ 100 $\times$ 20 cells (total of 200,000 cells). The storage aquifer, located in the center of the model, is 12~km $\times$ 12~km $\times$ 100~m, which corresponds to 80 $\times$ 80 $\times$ 20 cells (128,000 cells). The cell size within the storage aquifer is constant, while cell size increases with distance from the storage aquifer in the surrounding region. This setup provides computational efficiency gains relative to using a uniform grid over the entire 120~km $\times$ 120~km $\times$ 100~m region.
 
The geomodels for the storage aquifer are constructed from the standard multi-Gaussian fields and metaparameters sampled from their prior distributions. The mean and standard deviation of the log-permeability field, as well as the coefficients relating porosity and permeability, are sampled from uniform prior distributions. The permeability anisotropy ratio is sampled from a log-uniform prior distribution. The values of rock physics parameters, as well as ranges for prior distributions of the metaparameters, are presented in Table~\ref{rock physics}. Cutoff values for porosity (0.05 and 0.4) and permeability ($10^{-4}$~md and $10^{4}$~md) are applied to avoid nonphysical values. The porosity and permeability of the domain surrounding the storage aquifer are set to 0.1 and 10~md, consistent with the average porosity and permeability of the In Salah storage aquifer~\citep{li2016coupled}. 

\begin{table}[!ht]
\begin{center}
\footnotesize
\caption{Parameters used in the flow simulations}
\label{rock physics}
\renewcommand{\arraystretch}{1.2} 
\begin{tabular}{ c c } 
\hline
\textbf{Storage aquifer parameters} & \textbf{Values or ranges} \\ 
\hline
Mean of log-permeability, $\mu_{\log k}$  & $\mu_{\log k} \sim$ $U$(1.5, 4): (4.5, 54.6) md \\ 
Standard deviation of log-permeability, $\sigma_{\log k}$ & $\sigma_{\log k} \sim$ $U$(1, 2.5) \\ 
Parameter $d$ in $\log k$ -- $\phi$ correlation & $d \sim U(0.02, 0.04)$ \\ 
Parameter $e$ in $\log k$ -- $\phi$ correlation & $e \sim U(0.05, 0.1)$ \\ 
Horizontal correlation length, $l_h$ & $l_h \in \bigl\{10, 15, 20, 25, 30\bigl\}$~cells \\
Vertical correlation length, $l_v$ & 2~cells \\
Permeability anisotropy ratio, $a_r$  & $\log_{10}(a_r)$ $\sim$ $U$(-2, 0) \\ 
\hline
\textbf{Surrounding region parameters} & \textbf{Values} \\ 
\hline
Permeability & 10 md \\ 
Porosity & 0.1 \\ 
\hline
\textbf{Relative permeability and capillary pressure} & \textbf{Values} \\ 
\hline
Irreducible water saturation, $\emph{S$_{wi}$}$ & 0.2 \\
Residual CO$_2$ saturation, $\emph{S$_{gr}$}$ & 0 \\
Water exponent for Corey model, $\emph{n$_{w}$}$ & 6 \\
CO$_2$ exponent for Corey model, $\emph{n$_{g}$}$ & 5 \\
Relative permeability of CO$_2$ at $\emph{S$_{wi}$}$, $\emph{k$_{rg}$}$($\emph{S$_{wi}$}$) & 0.95 \\
Capillary pressure exponent $\lambda$ & 0.67 \\ 
\hline
\end{tabular}
\end{center}
\end{table}

\begin{figure}[!ht]
\centering
\subfloat[Relative permeability curves]{\label{fig:a}\includegraphics[width = 70mm]{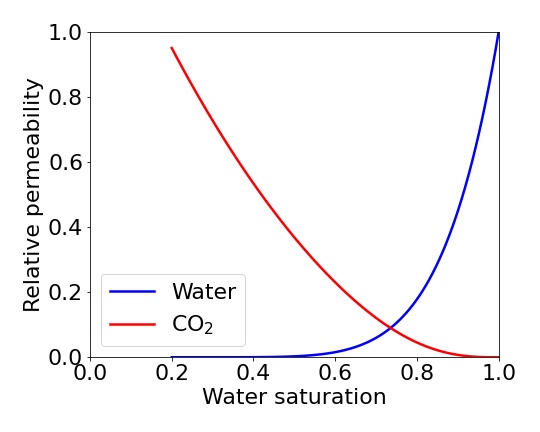}}
\hspace{2.5mm}
\subfloat[Capillary pressure curve]{\label{fig:b}\includegraphics[width = 70mm]{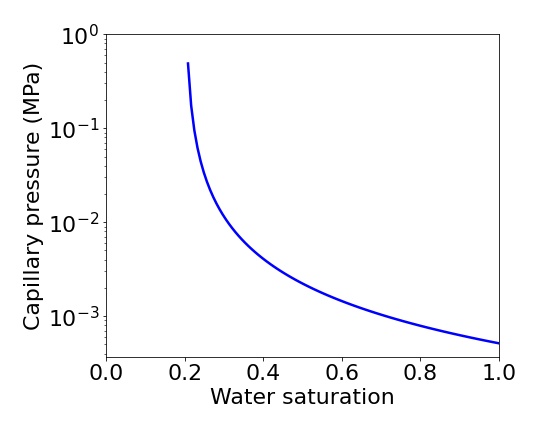}}
\caption{Two-phase flow curves. Capillary pressure curve in (b) is for $\phi=0.1$ and $k=10$~md.}
\label{Relative_Permeability}
\end{figure}

The relative permeability and capillary pressure curves used for our flow simulations are based on measured data for Berea sandstone~\citep{krevor2012} (the curves reported for In Salah led to numerical problems in our simulations). The CO$_2$-water relative permeability curves are shown in Fig.~\ref{Relative_Permeability}a. These curves correspond to Corey coefficients of $n_w$ = 6, $n_g$ = 5, and $k_{rg}({S_{wi}})$ = 0.95, where $n_w$ and $n_g$ are the relative permeability exponents and $k_{rg}({S_{wi}})$ is the relative permeability of CO$_2$ at irreducible water saturation ($S_{wi}$). The Brooks-Corey~\citep{saadatpoor2010new} model, with exponent ($\lambda$) of 0.67, is used to generate the capillary pressure curves. The Leverett J-function is used to provide the capillary pressure curve for each cell. The capillary pressure curve for porosity of 0.1 and permeability of 10~md is shown in Fig.~\ref{Relative_Permeability}b.

\begin{figure}[!ht]
\centering  
\includegraphics[width=7.5cm]{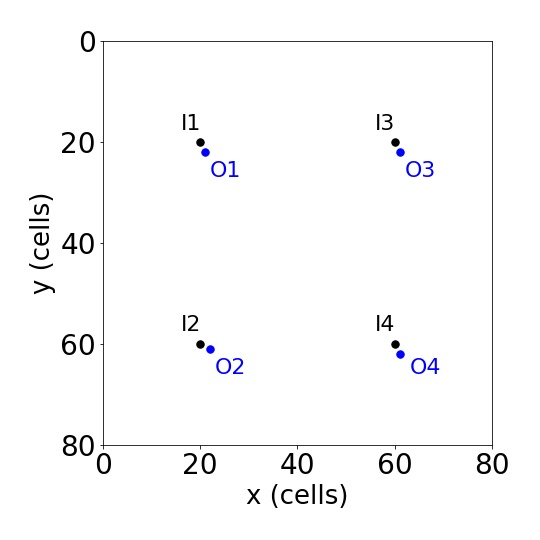}
\caption{Locations of injection wells (I1 -- I4) and observation wells (O1 -- O4) in the $80 \times 80 \times 20$ storage aquifer. All wells are fully penetrating.}
\label{wells}
\end{figure}

Consistent with the conditions at In Salah, we set the initial hydrostatic pressure in the middle layer of the overall domain to 19~MPa. Temperature is constant throughout the domain at 90~$^{\circ}$C. At initial conditions, the viscosity of the supercritical CO$_2$ is 0.047~cp, and the viscosity of brine is 0.38~cp. No-flow conditions are prescribed at the boundaries of the full domain. Four fully-penetrating vertical wells, denoted I1 -- I4, operate within the storage aquifer. The injection well locations, along with the observation well locations (O1 -- O4), are shown in Fig.~\ref{wells}. Each well injects 1~Mt CO$_2$ per year for 30~years. In the absence of a detailed well model, the injection rate into each cell intersected by the well is specified to be proportional to the well-block permeability. This treatment approximates the injection profile that would result from the use of a general well model.

\subsection{Surrogate model error statistics and predictions}
\label{error statistics}

The process for generating the training and test data is as follows. We generate 2000 standard multi-Gaussian realizations with SGeMS. For these models, five different values of $l_h$ (1500, 2250, 3000, 3750 and 4500~m) are considered. 
The geomodel realizations are then constructed from the multi-Gaussian fields and the five metaparameters in Eq.~\ref{metaparameters} (sampled from their prior distributions), as described in Section~\ref{Geomodels}. Flow simulation is then performed for each realization using GEOS. The pressure and saturation fields in the storage aquifer are collected at 10 discrete time steps (specifically 1.5, 3, 4.5, 7.5, 10.5, 13.5, 18, 22.5, 27 and 30~years after the start of injection). Separate surrogate models are trained for saturation and pressure. A test set of $n_e$ = 500 new geomodels is then generated and simulated using the same procedures. This set is used to evaluate the performance of the surrogate model, as we now describe.

We first present error quantities for CO$_2$ saturation and pressure. The relative error in CO$_2$ saturation, for test sample $i$ ($i = 1, \dots, 500$), is denoted $\delta_S^i$. This is computed as

\begin{equation} \label{error_s}
\delta_S^i = \frac{1}{n_sn_t} \sum_{j=1}^{n_s} \sum_{t=1}^{n_t} \frac{| (\hat{S}_s)_{i,j}^t - (S_s)_{i,j}^t |}{(S_s)_{i,j}^t + \epsilon}, 
\end{equation}

\noindent where ${n_s}$ = 128,000 is the total number of cells in the storage aquifer, $n_t$ = 10 is the number of output time steps considered, and $(S_s)_{i,j}^t$ and $(\hat{S}_s)_{i,j}^t$ are CO$_2$ saturation predictions from GEOS and the surrogate model, respectively, for test case $i$, cell $j$ and time step $t$. A constant $\epsilon$ = 0.025 is included to prevent division by zero (or a very small value). 

The relative error for pressure for sample $i$ ($i = 1, \dots, 500$), denoted $\delta_p^i$, is defined similarly, though in this case we normalize by the difference between the maximum and minimum pressure, i.e.,

\begin{equation} \label{error_p}
\delta_p^i = \frac{1}{n_sn_t} \sum_{j=1}^{n_s} \sum_{t=1}^{n_t} \frac{| (\hat{p}_s)_{i,j}^t - (p_s)_{i,j}^t |}{(p_s)_{i,\mathrm{max}}^t - (p_s)_{i,\mathrm{min}}^t}. 
\end{equation}

\noindent Here $(p_s)_{i,j}^t$ and $(\hat{p}_s)_{i,j}^t$ are pressure predictions from GEOS and the surrogate model, for test case $i$, cell $j$ and time step $t$, and $(p_s)_{i,\mathrm{max}}^t$ and $(p_s)_{i,\mathrm{min}}^t$ are the maximum and minimum pressure for all cells in test case $i$ at time step $t$. Note that we use a different definition of relative error for pressure than for saturation. This is because the initial pressure is already large (19~MPa in the middle layer), so computing relative error with $(p_s)_{i,j}^t$ in the denominator of Eq.~\ref{error_p} may understate the scale of the error. Our approach is more conservative and results in larger errors, but we believe these values better characterize surrogate model performance.

\begin{figure}[!ht]
\centering   
\subfloat[Saturation relative error]{\label{fig:a}\includegraphics[width = 73mm]{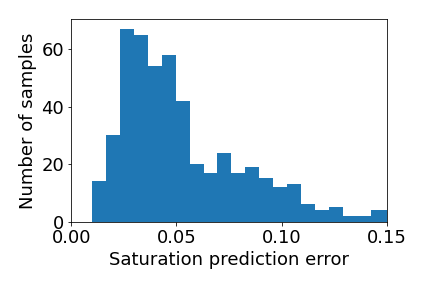}}
\hspace{2mm}
\subfloat[Pressure relative error]{\label{fig:b}\includegraphics[width = 73mm]{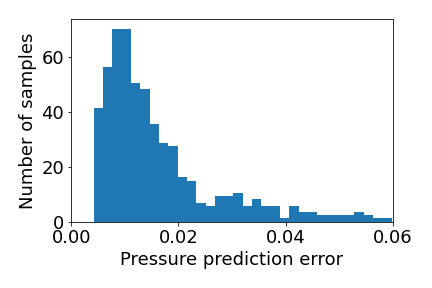}}
\caption{Histograms of relative errors for the 500 test cases.}
\label{errors}
\end{figure}

Histograms of the test-case relative errors for saturation and pressure are displayed in Fig.~\ref{errors}. The mean and median relative errors for saturation over the 500 test cases are 5.4\% and 4.5\%, respectively, and for pressure they are 1.9\% and 1.3\%. These errors are comparable to (but slightly larger than) those achieved in \citet{tang2022deep}, where average saturation and pressure errors of 5.3\% and 0.3\% were reported for test cases drawn from a single geological scenario. In the saturation error calculation in \citet{tang2022deep}, $\epsilon=0.01$ was used, which would tend to inflate the errors in that study compared to those reported here. In any event, the small relative errors achieved here demonstrate the applicability of the surrogate model for geomodels drawn from a wide range of geological scenarios. 

\begin{figure}[!ht]
\centering   
\subfloat[Porosity field ($l_h=1500$~m, $\mu_{\phi}$=0.13, $\sigma_{\phi}$=0.06)]{\label{fig:a}\includegraphics[width = 50mm]{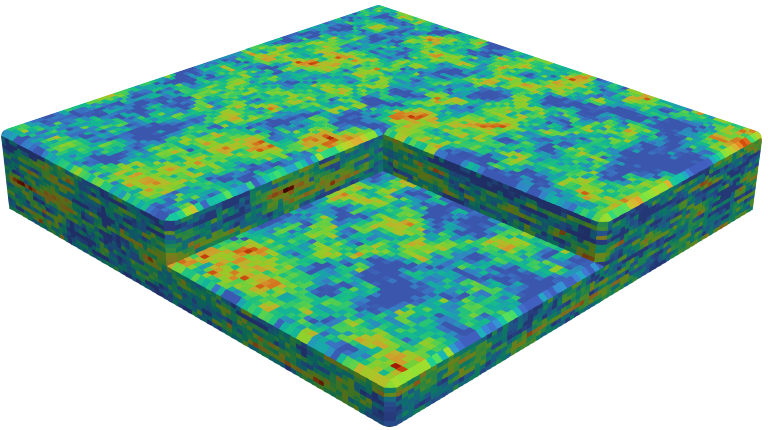}}
\includegraphics[width = 8mm]{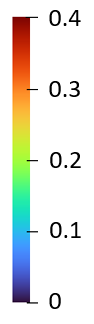}
\hspace{9mm}
\subfloat[Log-permeability field ($l_h=1500$~m, $\mu_{\log k}$=2.27, $\sigma_{\log k}$=2.06)]{\label{fig:b}\includegraphics[width = 50mm]{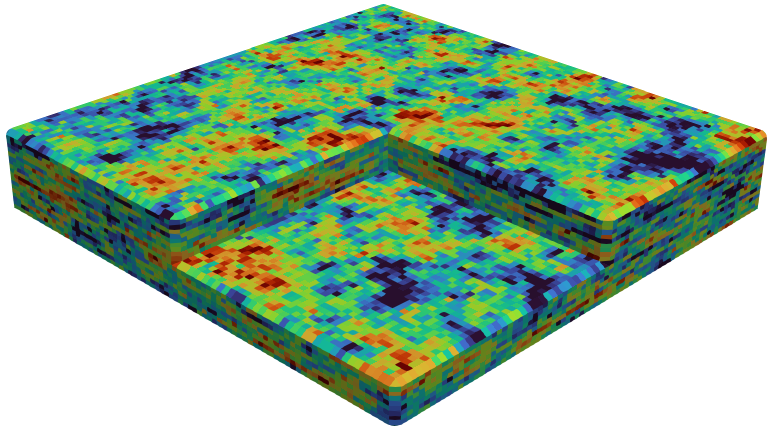}}
\includegraphics[width = 8mm]{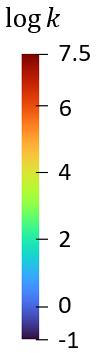}
\caption{Porosity and log-permeability fields of test-case realization corresponding to median relative saturation error.}
\label{Representative}
\end{figure}

\begin{figure}[!ht]
\centering   
\subfloat[4.5 years (sim)]{\label{fig:a}\includegraphics[width=45mm]{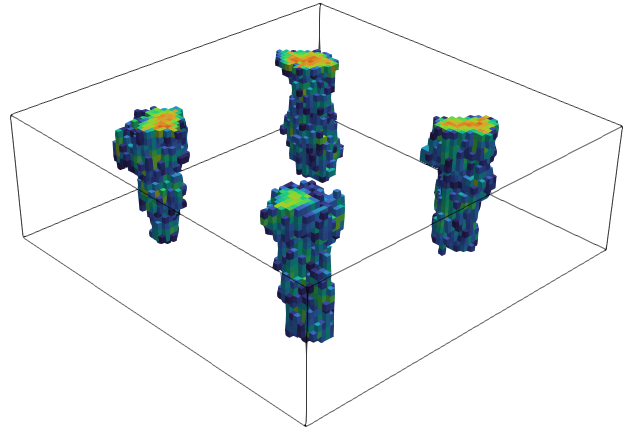}}
\hspace{3mm}
\subfloat[13.5 years (sim)]{\label{fig:b}\includegraphics[width=45mm]{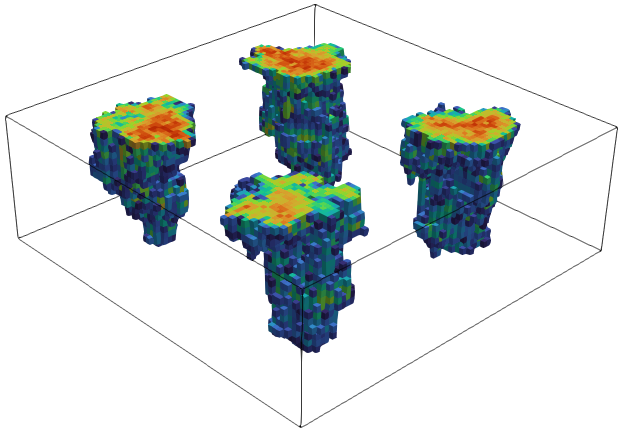}}
\hspace{3mm}
\subfloat[30 years (sim)]{\label{fig:c}\includegraphics[width=45mm]{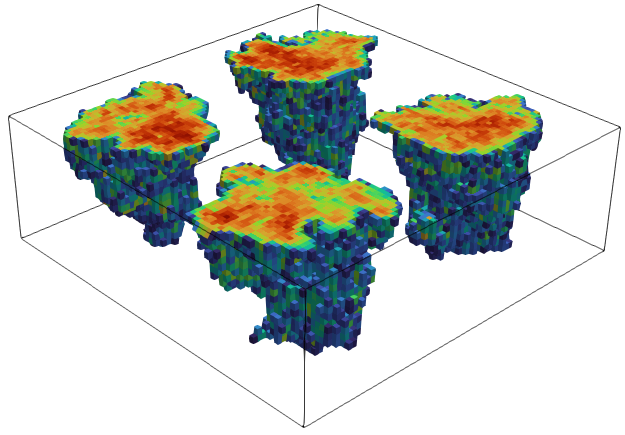}}
\includegraphics[width=7mm]{Figure/Test_Case/Sw_Scale.png}\\[1ex]
\subfloat[4.5 years (surr)]{\label{fig:d}\includegraphics[width=45mm]{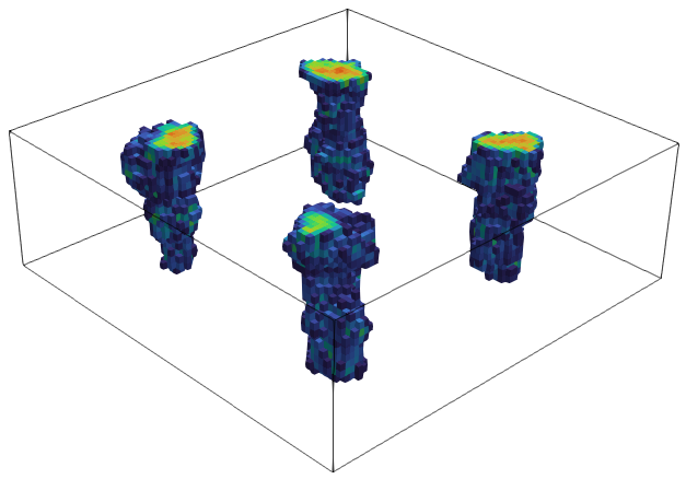}}
\hspace{3mm}
\subfloat[13.5 years (surr)]{\label{fig:e}\includegraphics[width=45mm]{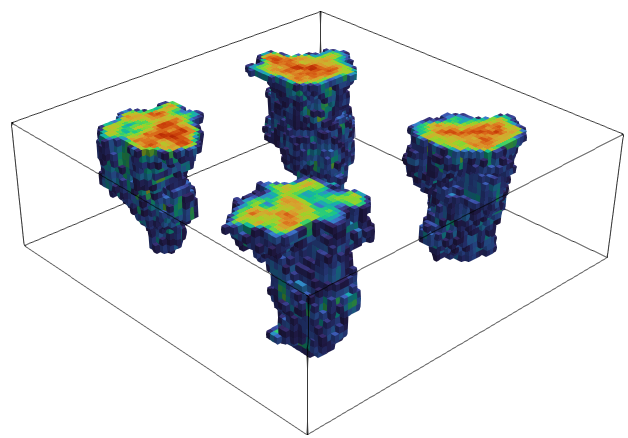}}
\hspace{3mm}
\subfloat[30 years (surr)]{\label{fig:f}\includegraphics[width=45mm]{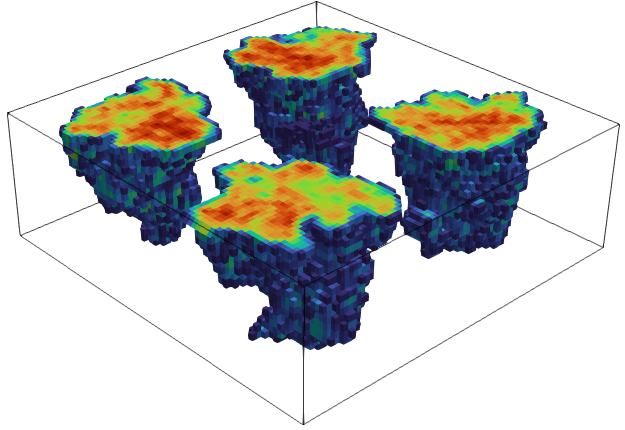}}
\includegraphics[width=7mm]{Figure/Test_Case/Sw_Scale.png}\\[1ex]
\caption{CO$_2$ saturation in the storage aquifer from GEOS flow simulation (upper row) and the extended recurrent R-U-Net surrogate model (lower row) for representative test case at three time steps.}
\label{S2:saturation}
\end{figure}

\begin{figure}[!ht]
\centering   
\subfloat[4.5 years (sim)]{\label{fig:a}\includegraphics[width=41mm]{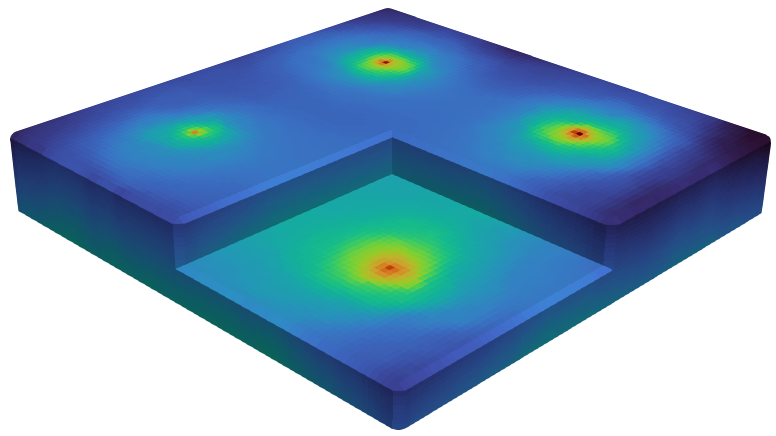}}
\includegraphics[width=7.75mm]{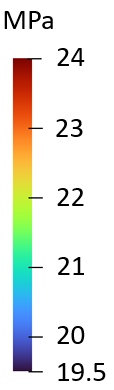}
\hspace{2mm}
\subfloat[13.5 years (sim)]{\label{fig:b}\includegraphics[width=41mm]{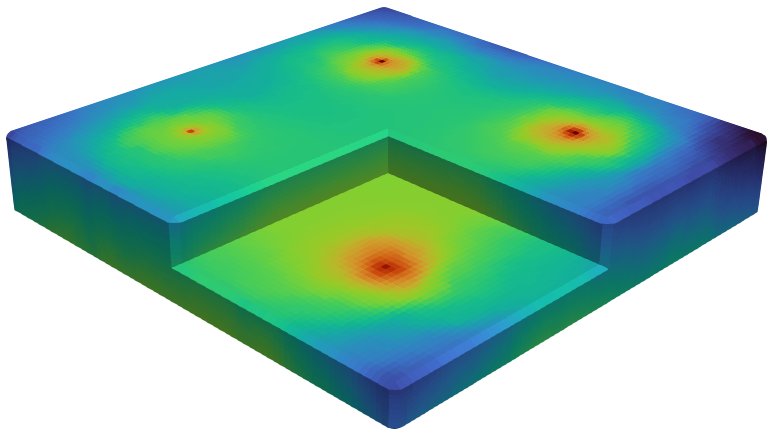}}
\includegraphics[width=6.6mm]{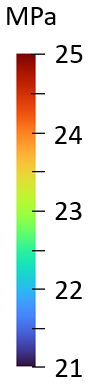}
\hspace{2mm}
\subfloat[30 years (sim)]{\label{fig:c}\includegraphics[width=41mm]{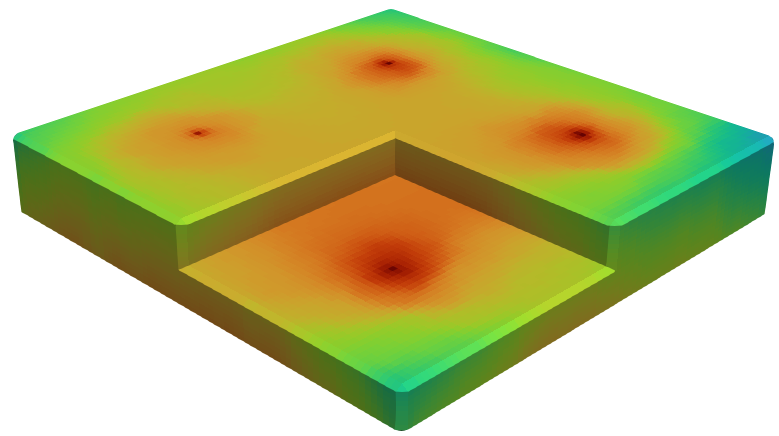}}
\includegraphics[width=7.8mm]{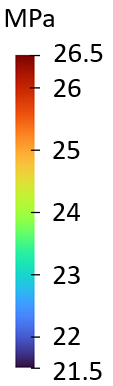}\\[1ex]
\subfloat[4.5 years (surr)]{\label{fig:d}\includegraphics[width=41mm]{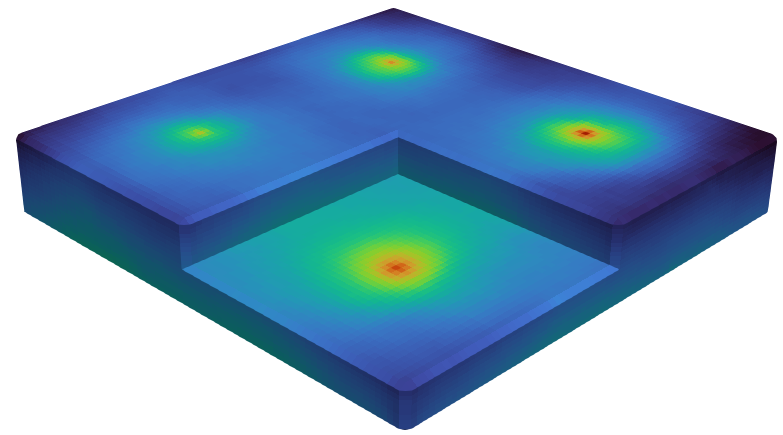}}
\includegraphics[width=7.75mm]{Figure/Test_Case/P50_P_1_Scale.PNG}
\hspace{2mm}
\subfloat[13.5 years (surr)]{\label{fig:e}\includegraphics[width=41mm]{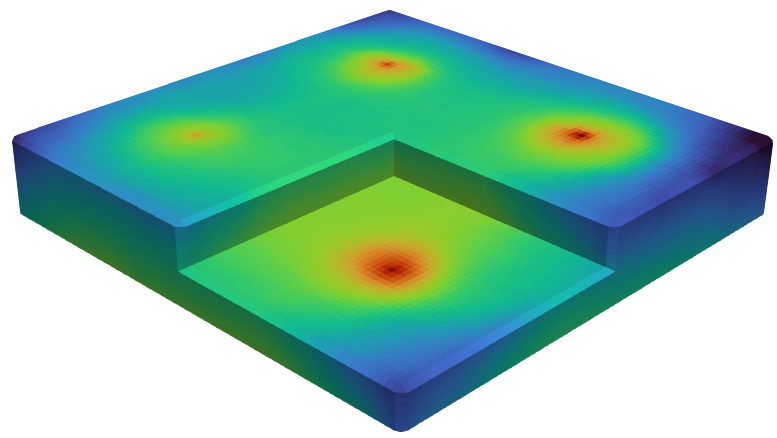}}
\includegraphics[width=6.6mm]{Figure/Test_Case/P50_P_2_Scale.PNG}
\hspace{2mm}
\subfloat[30 years (surr)]{\label{fig:f}\includegraphics[width=41mm]{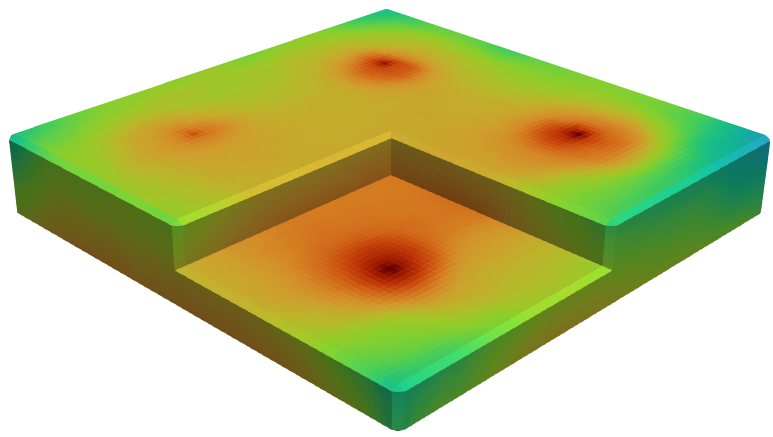}}
\includegraphics[width=7.8mm]{Figure/Test_Case/P50_P_3_Scale.PNG}
\caption{Pressure in the storage aquifer from GEOS flow simulation (upper row) and the extended recurrent R-U-Net surrogate model (lower row) for representative test case at three time steps.}
\label{P50:pressure}
\end{figure}

We next present detailed results for the case corresponding to the median relative saturation error. The geomodel for this case is shown in Fig.~\ref{Representative}. This realization corresponds to the following metaparameters: $l_h=1500$~m, $\mu_{\log k}=2.27$, $\sigma_{\log k}=2.06$, $a_r=0.4$, $d=0.034$ and $e=0.055$. The saturation and pressure fields for this test case at 4.5, 13.5, and 30~years are displayed in Figs.~\ref{S2:saturation} and~\ref{P50:pressure}. The upper rows in the figures show the reference GEOS results, while the lower rows display the surrogate model results. There is close visual correspondence between the flow simulation and surrogate model results for both quantities. The saturation plumes display irregular shapes (most notably at 30~years) due to the highly heterogeneous porosity and log-permeability fields. The plume radii at the end of injection are larger in the upper layers than the lower layers due to the relatively high permeability anisotropy ratio.

The pressure fields for the same test case are shown in Fig.~\ref{P50:pressure}. The surrogate predictions are again very close to the GEOS results. At early time, pressure buildup is mainly around the injection wells (Fig.~\ref{P50:pressure}a and d). The no-flow boundaries, although distant, eventually impact the pressure field. This is evident in the significantly increased pressure, over the full storage aquifer, after 30~years (Fig.~\ref{P50:pressure}c and f). 

\subsection{Surrogate predictions for geomodels from various scenarios}
\label{Surrogate predictions}

The saturation plumes and pressure fields for geomodel realizations drawn from different geological scenarios can vary substantially. Vertical permeability, for example, impacts vertical CO$_2$ migration, which is driven by gravity and density difference (the density of supercritical CO$_2$ and brine are 610~kg/m$^3$ and 975~kg/m$^3$, respectively, at initial conditions). The anisotropy ratio can therefore have a significant effect on the shape and extent of the plume. In addition, mean porosity and mean log-permeability can have large impacts on the pressure field. These quantities all correspond to metaparameters that vary over substantial ranges in this study.

\begin{figure}[!ht]
\centering   
\subfloat[Realization~1 ($l_h$=1500~m, $\mu_{\log k}$=3.78, $\sigma_{\log k}$=1.42, $a_r$=0.56, $d$=0.02, $e$=0.084)]{\label{fig:a}\includegraphics[width = 45mm]{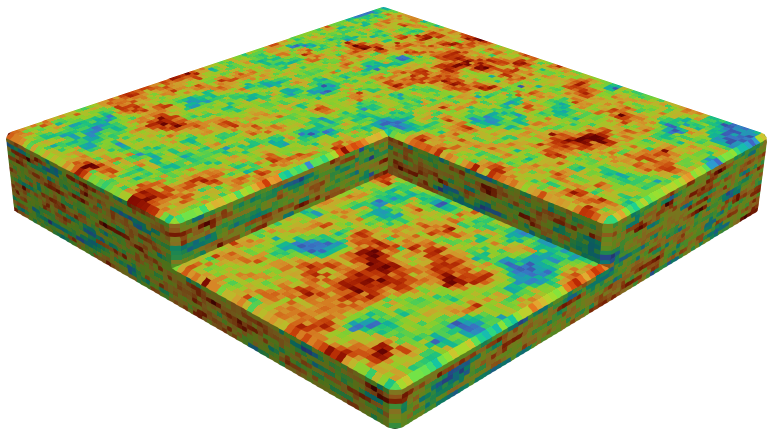}}
\hspace{3mm}
\subfloat[Realization~2 ($l_h$=1500~m, $\mu_{\log k}$=2.44, $\sigma_{\log k}$=1.37, $a_r$=0.14, $d$=0.025, $e$=0.074)]{\label{fig:b}\includegraphics[width = 45mm]{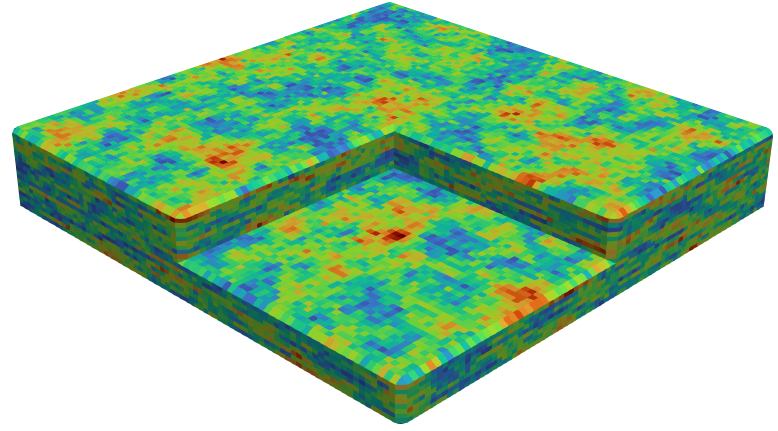}}
\hspace{3mm}
\subfloat[Realization 3 ($l_h$=2250~m, $\mu_{\log k}$=2.92, $\sigma_{\log k}$=1.77, $a_r$=0.01, $d$=0.029, $e$=0.09)]{\label{fig:b}\includegraphics[width = 45mm]{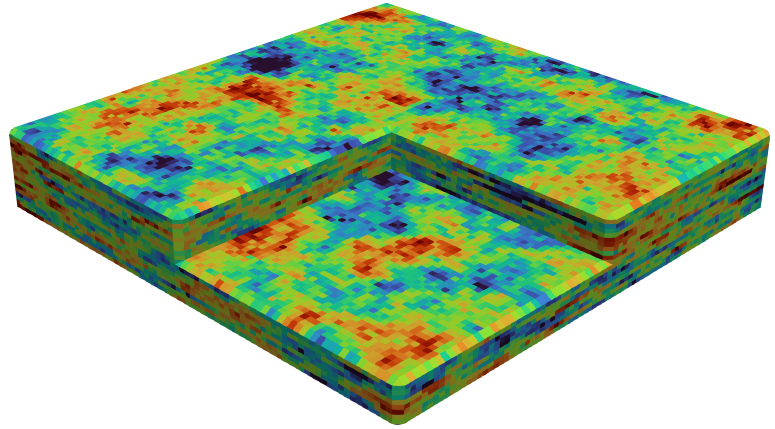}}
\includegraphics[width = 7mm]{Figure/True_Model_2/Range_k.PNG}
\caption{Log-permeability ($\log k$) fields for geomodels drawn from three geological scenarios. Geomodels in (a) and (b) are the `true' log-permeability fields used for history matching in Section~\ref{History-matching} and SI.}
\label{True_Models}
\end{figure}

The log-permeability fields for the three realizations considered, each from a different geological scenario, are displayed in Fig.~\ref{True_Models}. The realizations vary significantly in appearance, and they also vary in $a_r=k_v/k$, which is not evident in the figure because $\log k$ is the quantity displayed.

\begin{figure}[!ht]
\centering   
\subfloat[Realization 1 (sim)]{\label{fig:a}\includegraphics[width=45mm]{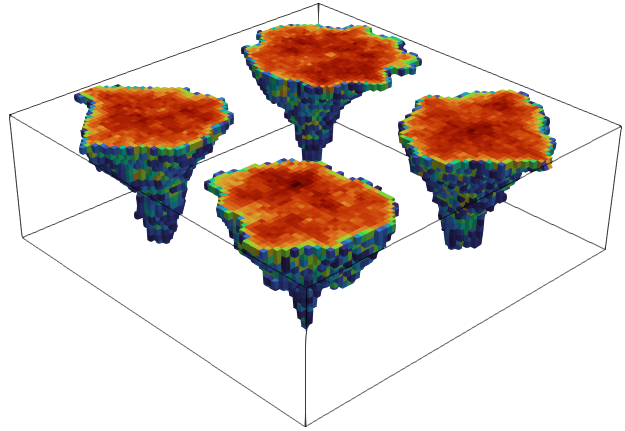}}
\hspace{3mm}
\subfloat[Realization 2 (sim)]{\label{fig:b}\includegraphics[width=45mm]{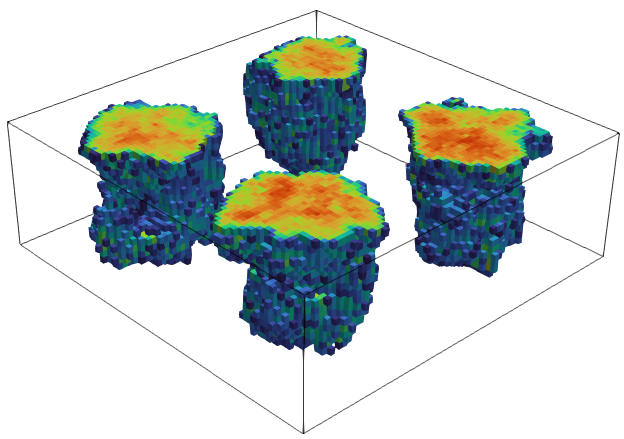}}
\hspace{3mm}
\subfloat[Realization 3 (sim)]{\label{fig:c}\includegraphics[width=45mm]{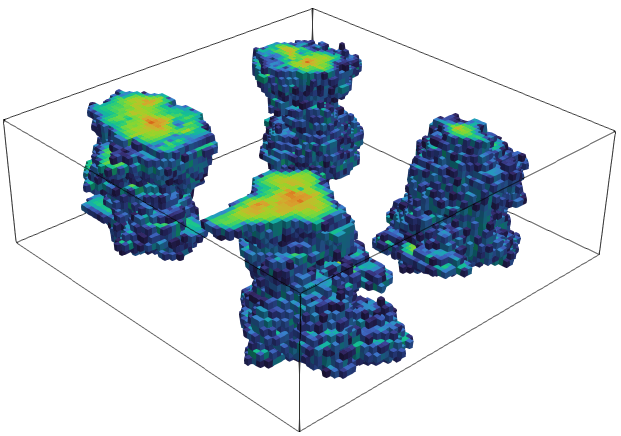}}
\includegraphics[width=8mm]{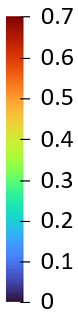}\\[1ex]
\subfloat[Realization 1 (surr)]{\label{fig:d}\includegraphics[width=45mm]{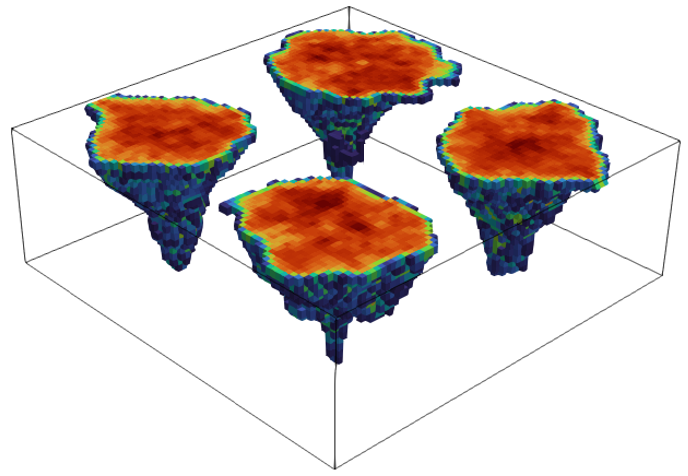}}
\hspace{3mm}
\subfloat[Realization 2 (surr)]{\label{fig:e}\includegraphics[width=45mm]{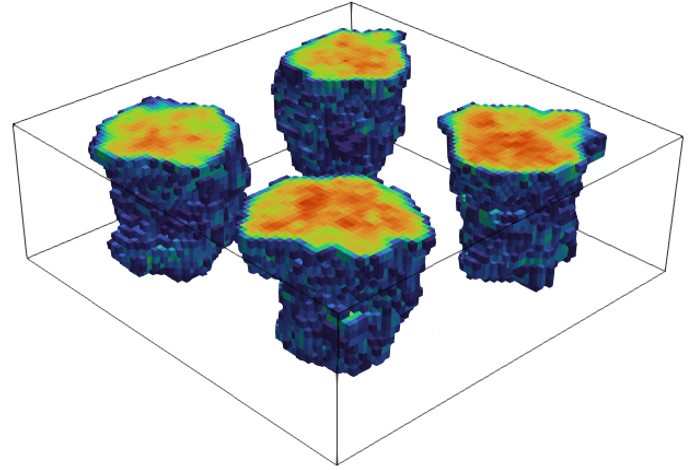}}
\hspace{3mm}
\subfloat[Realization 3 (surr)]{\label{fig:f}\includegraphics[width=45mm]{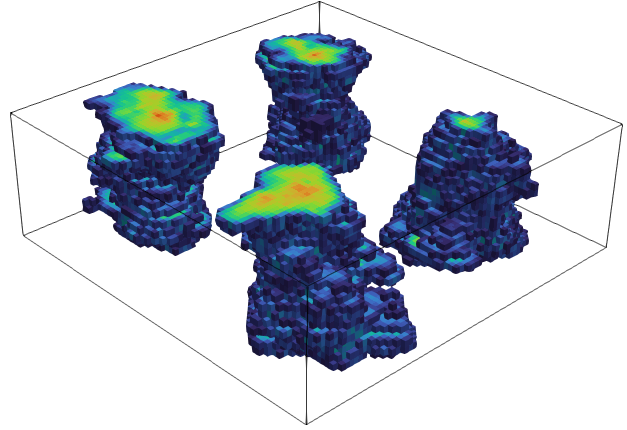}}
\includegraphics[width=8mm]{Figure/Test_Case/Sw_Scale.PNG}\\[1ex]
\caption{CO$_2$ saturation in the storage aquifer after 30~years of injection from GEOS (upper row) and the surrogate model (lower row) for geomodels drawn from three geological scenarios. Corresponding geological realizations shown in Fig.~\ref{True_Models}.}
\label{saturations}
\end{figure}

The CO$_2$ saturation plumes at the end of injection for these three geomodels are shown in Fig.~\ref{saturations}. The relative saturation errors are 1.6\%, 6.1\% and 3.3\%, respectively. Of particular interest here is the general agreement between the GEOS and surrogate model results, which requires the surrogate model to capture a range of plume shapes and extents. For Realization~1, $a_r=0.56$. With this relatively large anisotropy ratio vertical flow is not impeded, leading to conical-shaped saturation plumes (Fig.~\ref{saturations}a and d). Because CO$_2$ migrates vertically due to gravity effects, the plume radii in the top layer of the model are large. For Realizations~2 and 3, $a_r=0.14$ and 0.01, respectively. Vertical flow is thus more restricted, and the plume shapes shift from conical (Fig.~\ref{saturations}a and d) to cylindrical (Fig.~\ref{saturations}b and e) to irregular (Fig.~\ref{saturations}c and f). As a result of geological heterogeneity,  the saturation plumes associated with the four injection wells display clear differences in all cases. Importantly, these effects are captured by the surrogate model.

\begin{figure}[!ht]
\centering   
\subfloat[Realization 1 (sim)]{\label{fig:a}\includegraphics[width=44mm]{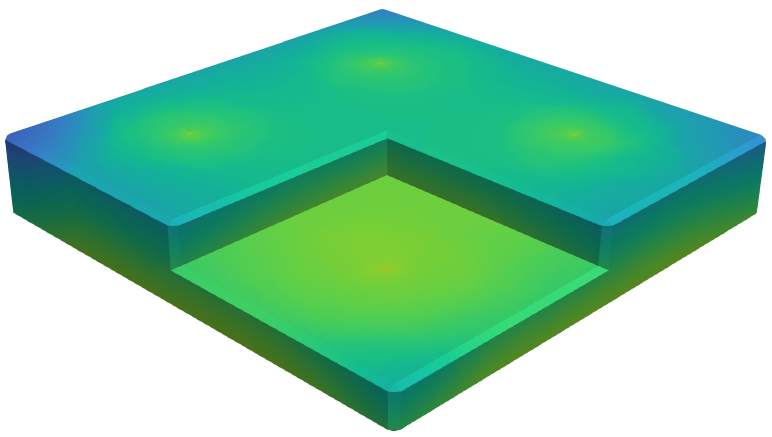}}
\hspace{3mm}
\subfloat[Realization 2 (sim)]{\label{fig:b}\includegraphics[width=44mm]{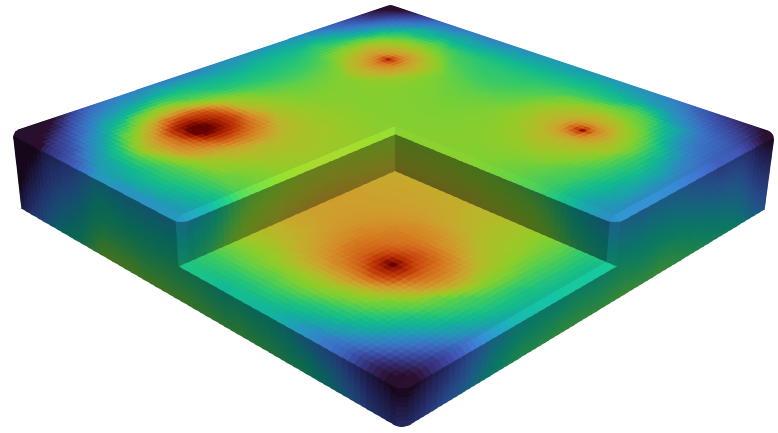}}
\hspace{3mm}
\subfloat[Realization 3 (sim)]{\label{fig:c}\includegraphics[width=44mm]{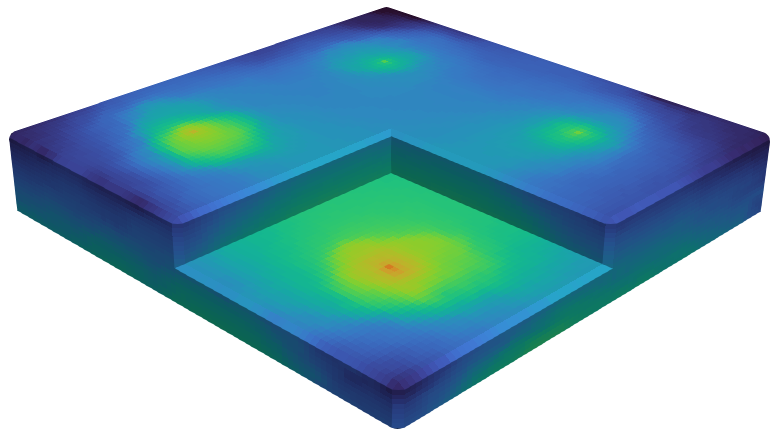}}
\includegraphics[width=8mm]
{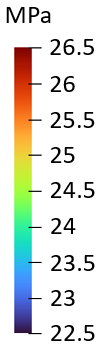}\\[1ex]
\subfloat[Realization 1 (surr)]{\label{fig:d}\includegraphics[width=44mm]{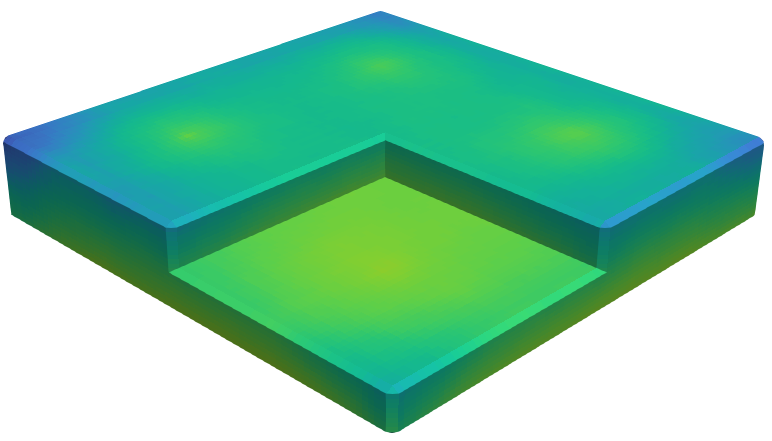}}
\hspace{3mm}
\subfloat[Realization 2 (surr)]{\label{fig:e}\includegraphics[width=44mm]{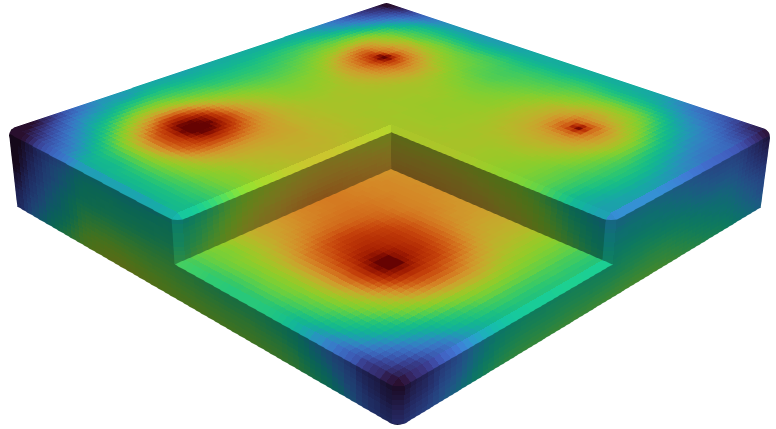}}
\hspace{3mm}
\subfloat[Realization 3 (surr)]{\label{fig:f}\includegraphics[width=44mm]{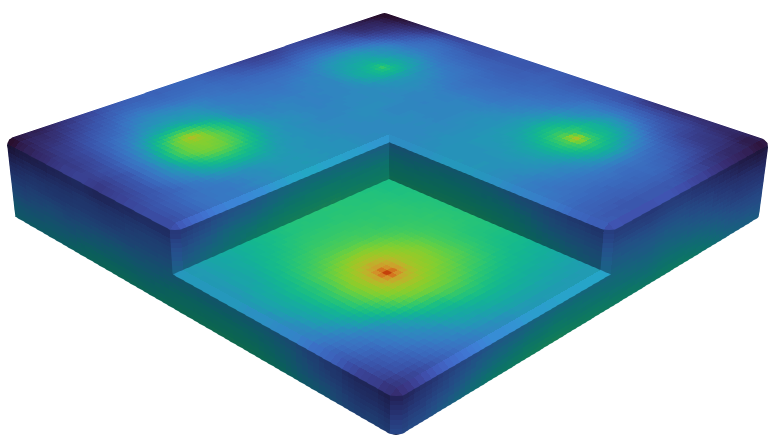}}
\includegraphics[width=8mm]
{Figure/Test_Case/P_Scale.PNG}
\caption{Pressure fields in the storage aquifer after 30~years of injection from GEOS (upper row) and the surrogate model (lower row) for geomodels drawn from three geological scenarios. Corresponding geological realizations shown in Fig.~\ref{True_Models}.}
\label{pressures_1}
\end{figure}

The pressure fields at the end of the injection period for these geomodels are shown in Fig.~\ref{pressures_1}. The correspondence between surrogate model and GEOS results is again very close despite the variation in solution character. The surrogate model relative pressure errors for Realizations~1, 2 and 3 are 1.3\%, 2.2\% and 1.0\%, respectively. The highest pressure buildup is observed in Realization~2, which is reasonable because this model corresponds to the lowest $\mu_{\log k}$ (2.44). The pressure fields for Realizations~1 and 3 are different in character, which could be due to the different metaparameters and/or to differences in near-well properties, with Realization~1 showing a more uniform response.

As noted earlier, vertical observation wells (denoted O1 -- O4), shown in Fig.~\ref{wells}, are positioned near each of the four injection wells. These observation wells provide the saturation and pressure data that will be used in history matching. To evaluate the accuracy of surrogate model predictions for this type of data, we now assess the overall (statistical) correspondence between GEOS and surrogate model results for monitoring well data. Specifically, we consider P$_{10}$, P$_{50}$ and P$_{90}$ (10th, 50th and 90th percentile) results, for both saturation and pressure, at particular locations through time. 

\begin{figure}[!ht]
\centering
\subfloat[Saturation at O1 in layer 1]{\label{fig:a}\includegraphics[width = 70mm]{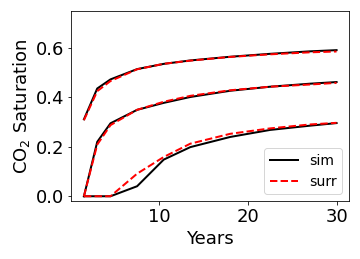}}
\hspace{2mm}
\subfloat[Saturation at O2 in layer 5]{\label{fig:b}\includegraphics[width = 70mm]{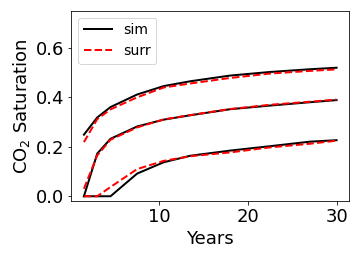}}
\\[1ex]
\subfloat[Saturation at O3 in layer 10]{\label{fig:c}\includegraphics[width = 70mm]{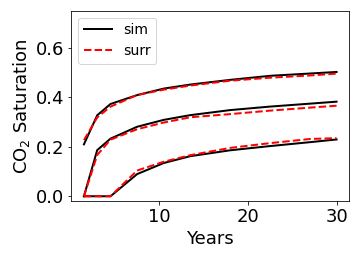}}
\hspace{2mm}
\subfloat[Saturation at O4 in layer 20]{\label{fig:d}\includegraphics[width = 70mm]{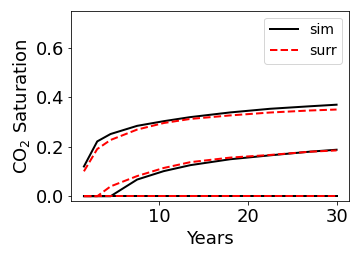}}
\\[1ex]
\caption{Saturation ensemble statistics from GEOS (black solid curves) and surrogate model (red dashed curves) at four observation locations. The upper, middle and lower curves correspond to P$_{90}$, P$_{50}$ and P$_{10}$ results over the full ensemble of 500 test cases.}
\label{statistics_s}
\end{figure}

\begin{figure}[!ht]
\centering
\subfloat[Pressure at O1 in layer 1]{\label{fig:a}\includegraphics[width = 70mm]{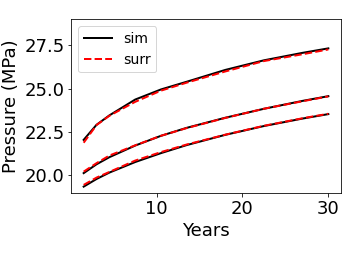}}
\hspace{2mm}
\subfloat[Pressure at O2 in layer 5]{\label{fig:b}\includegraphics[width = 70mm]{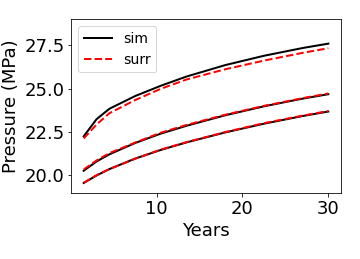}}
\\[1ex]
\subfloat[Pressure at O3 in layer 10]{\label{fig:c}\includegraphics[width = 70mm]{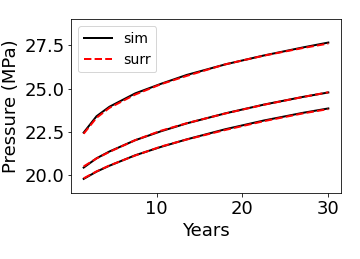}}
\hspace{2mm}
\subfloat[Pressure at O4 in layer 20]{\label{fig:d}\includegraphics[width = 70mm]{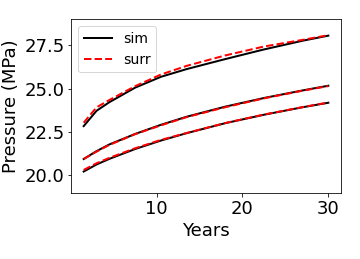}}
\\[1ex]
\caption{Pressure ensemble statistics from GEOS (black solid curves) and surrogate model (red dashed curves) at four observation locations. The upper, middle and lower curves correspond to P$_{90}$, P$_{50}$ and P$_{10}$ results over the full ensemble of 500 test cases.}
\label{statistics_p}
\end{figure}

The ensemble statistical results (P$_{10}$, P$_{50}$ and P$_{90}$ curves) for saturation and pressure are shown in Figs.~\ref{statistics_s} and~\ref{statistics_p}. The solid black curves represent GEOS results, and the dashed red curves are surrogate model results. The upper curves are the P$_{90}$ results (at each of the 10~time steps), the middle curves are the P$_{50}$ results, and the lower curves the P$_{10}$ results. Although each monitoring well can provide data at all 20~layers, the results presented are for one particular layer. Close agreement is consistently observed between the two sets of results, indicating that, at least in a statistical sense, the surrogate model is able to provide accurate predictions at specific monitoring well locations. This capability is essential for history matching.

There are noticeable differences in the P$_{10}$--P$_{90}$ ranges for saturation in the top (Fig.~\ref{statistics_s}a) and bottom (Fig.~\ref{statistics_s}d) layers. This is due to gravity effects, which drive CO$_2$ upward. The clear increase in pressure with time, which results in large part from the no-flow outer boundary conditions, is evident in Fig.~\ref{statistics_p}. The change in pressure with time ($dp/dt$) becomes essentially constant, in space and time, after several years of injection. This is a characteristic of the pseudo-steady state flow regime, which is expected given the problem setup. 

The results in Figs.~\ref{statistics_s} and \ref{statistics_p} do not quantify the ability of the surrogate model to maintain correlations between predictions at different locations or times. Second-order statistics, as were considered by, e.g., \citet{zhong2019predicting} in their assessment of surrogate model performance, can also be evaluated. Comparisons between GEOS and surrogate model results for second-order-statistics quantities are presented in SI. There we show that the surrogate model is able to capture, e.g., the covariance between saturation in different layers along monitoring well~O1.

\section{History Matching using Surrogate Model} 
\label{History-matching}
We now apply the hierarchical MCMC-based history matching procedure in combination with the extended recurrent R-U-Net surrogate model to estimate the posterior distributions of the metaparameters. Saturation and pressure measurements in the monitoring wells provide the observed data. We first describe the setup for the history matching problem, including our treatment of error, and then present history matching results for a new (random) synthetic `true' model. History matching results for a second true model, along with results for Gaussian-distributed priors, are presented in SI.

\subsection{Problem setup for history matching}
\label{History matching setup}

The synthetic true models (here and in SI), denoted $\textbf{m}_{\mathrm{true}}$, correspond to newly generated samples from the (broad) prior distribution. For true model~1, we set $l_h=1500$~m (as noted earlier, $l_h$ is not included in the history matching). The true metaparameters are $\mu_{\log k}=3.78$, $\sigma_{\log k} = 1.42$, $a_r=0.56$, $d=0.02$ and $e=0.084$. The log-permeability field for true model~1 is shown in Fig.~\ref{True_Models}a. Flow simulation is performed on this model, and these simulation results, at observation-well locations, are taken as true data, denoted by $\textbf{d}_{\mathrm{true}}$. The observed data used in history matching, denoted $\textbf{d}_{\mathrm{obs}}$, are obtained by adding random error to $\textbf{d}_{\mathrm{true}}$. 

In this study, the error considered in the history matching workflow includes measurement error, model-resolution error, and surrogate-model error. The measurement error derives from the actual measurements of pressure and (interpreted) saturation-log values. Resolution error, which can also be viewed as upscaling error, arises from the mismatch between model resolution (grid cells are of dimensions 150~m $\times$ 150~m $\times$ 5~m) and the scale of the measured data (cm to m). Surrogate model error arises because we use the surrogate model, rather than the reference simulator GEOS, to provide flow predictions for the MCMC procedure. 

The observed data are measured at three time steps (1.5~years, 3~years, 4.5~years) after the start of injection. The observed saturation data include measurements, taken in all 20~layers, in each of the four observation wells (resulting in a total of 240 saturation measurements). Pressure data are measured at layer~1 in each of the observation wells, leading to a total of 12 pressure measurements. The observed data, with measurement error and model-resolution error, is expressed as

\begin{equation} \label{noise}
\textbf{d}_{\mathrm{obs}} = \textbf{d}_{\mathrm{true}} + \bm{\epsilon} = f(\textbf{m$_{\mathrm{true}}$}) + \bm{\epsilon},
\end{equation}
\noindent where $f(\textbf{m$_{\mathrm{true}}$})$ represents flow simulation results for the true model $\textbf{m}_{\mathrm{true}}$, and $\bm{\epsilon}$ denotes the measurement and model-resolution error. This quantity, which is taken to be normally distributed and uncorrelated, is sampled with $\textbf{0}$ mean and covariance $\widetilde{C}_{\mathrm{D}}$. This covariance involves two contributions, i.e., $\widetilde{C}_{\mathrm{D}} = {C_{\mathrm{D}}} + {C_{\mathrm{res}}}$, where ${C_{\mathrm{D}}}$ is the measurement error covariance and ${C_{\mathrm{res}}}$ is the covariance associated with model resolution.

Measurement error derives from the measurement device and data interpretation, and these can be readily quantified/estimated. Model resolution error, which is typically much larger than measurement error, is more difficult to assess. This error includes grid resolution and subgrid heterogeneity effects, though other modeling treatments and approximations, such as those associated with relative permeability, capillary pressure, hysteresis, dissolution functions, etc., also contribute and should be considered. Here we use the model error values suggested by \citet{jiang2023history}. In a limited grid resolution study for a similar problem, they found the standard deviation of the (combined) measurement and resolution errors to be around 0.1 (saturation units) for saturation, and 0.1~MPa for pressure. This pressure error corresponds to that used by \citet{sun2019data}, though the saturation error is larger. We considered a few different values (over a reasonable range) for these error parameters and did not observe significant differences in MCMC results. Accurate quantification of resolution/model error may be important in some settings and should be considered in future work.

The surrogate modeling error is more straightforward to estimate. This error, denoted $\bm{\epsilon}_{\mathrm{surr}}$, can be expressed as

\begin{equation} \label{model_error}
\bm{\epsilon}_{\mathrm{surr}} = f(\textbf{m$_{\mathrm{true}}$}) - \hat{f}(\textbf{m$_{\mathrm{true}}$}) ,
\end{equation}
\noindent where $f(\textbf{m$_{\mathrm{true}}$})$ and $\hat{f}(\textbf{m$_{\mathrm{true}}$})$ represent the GEOS and surrogate model predictions for saturation or pressure (for the true model). We estimate $\bm{\epsilon}_{\mathrm{surr}}$ by computing the surrogate model error, over the full ensemble of 500 test cases, at observation well locations. The mean and standard deviation of the surrogate model error for saturation at these locations are 0.001 and 0.055, and those for pressure are 0.001~MPa and 0.289~MPa. These errors do not involve absolute values, so error cancellation occurs (unlike in the computations in Eqs.~\ref{error_s} and \ref{error_p}). The mean saturation and pressure errors quantify the biases in surrogate model predictions. These biases can be neglected because they are small compared to the standard deviations. The standard deviations of the surrogate model errors are incorporated into the total error for history matching, given below.

As discussed in Section~\ref{MCMC}, surrogate predictions $\hat{f}(\textbf{m}_s)$ for new geomodels, constructed based on the proposed metaparameters and PCA latent variables, are used in the likelihood computation. The likelihood function is expressed as
\begin{equation} \label{likelihood}
  p(\bm{\mathrm{d}}_{\mathrm{obs}} | {\boldsymbol{\uptheta}}_{\mathrm{meta}}, \bm{\xi}) = \\
  c \exp\left({{-\frac{1}{2}  \left(\bm{\mathrm{d}}_{\mathrm{obs}}-\hat{f}\left(\textbf{m}_s\left({\boldsymbol{\uptheta}}_{\mathrm{meta}}, \bm{\xi}\right)\right) \right)}^T C_{\mathrm{tot}}^{-1} \left(\bm{\mathrm{d}}_{\mathrm{obs}}-\hat{f}\left(\textbf{m}_s\left({\boldsymbol{\uptheta}}_{\mathrm{meta}}, \bm{\xi}\right)\right) \right)} \right),
\end{equation} 
\noindent where $c$ is a normalization constant and $C_{\mathrm{tot}}$ is the total covariance, given by

\begin{equation} \label{likelihood}
  C_{\mathrm{tot}} = \widetilde{C}_{\mathrm{D}} + C_{\mathrm{surr}}.
\end{equation} 
Here $C_{\mathrm{surr}}$ represents the covariance of the surrogate model error. The errors discussed above were quantified in terms of standard deviation. For pressure, the corresponding $C_{\mathrm{tot}}$ is $[(0.1)^2 + (0.289)^2] = 0.094~{\text {MPa}}^2$. For saturation, $C_{\mathrm{tot}}=[(0.1)^2 + (0.055)^2] = 0.013$. 

\subsection{History matching results for true model~1}
\label{`true' model 1}

For true model~1 (shown in Fig.~\ref{True_Models}a), the MCMC history matching procedure requires 72,540 iterations to achieve convergence in the posterior distributions of the metaparameters. A total of 13,750 sets of metaparameters and geomodel realizations are accepted, which corresponds to an acceptance rate of about 19\%. The acceptance rate falls within the expected range of 10\% to 40\% for MCMC-based methods~\citep{gelman1996efficient}. Each surrogate model evaluation requires approximately 0.1~seconds on a single Nvidia A100 GPU. The high-fidelity (GEOS) flow simulations require about 5~minutes per run using 64 AMD EPYC-7543 CPU cores in parallel. Although the hardware environments for the two approaches are quite different, we achieve a speedup factor of 3000 in elapsed time with the surrogate model. Given the sequential nature of the MCMC procedure, the execution of this procedure would not be practical using high-fidelity flow simulation.

\begin{figure}[H]
\centering   
\subfloat[$\mu_{\mathrm{log}k}$]{\label{fig:a}\includegraphics[width = 50mm]{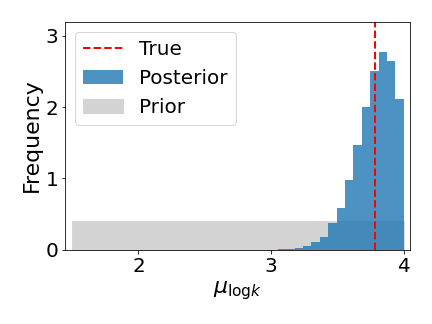}}
\hspace{2mm}
\subfloat[$\sigma_{\mathrm{log}k}$]{\label{fig:b}\includegraphics[width = 50mm]{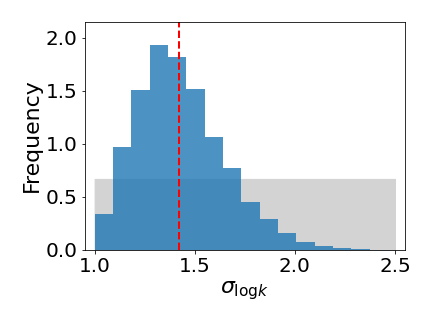}}
\hspace{2mm}
\subfloat[$\log_{10}(a_r)$]{\label{fig:d}\includegraphics[width = 50mm]{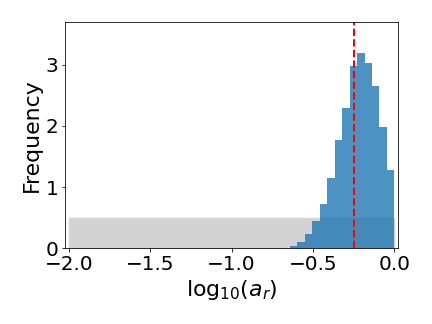}}\\
\subfloat[Parameter $d$]{\label{fig:d}\includegraphics[width = 50mm]{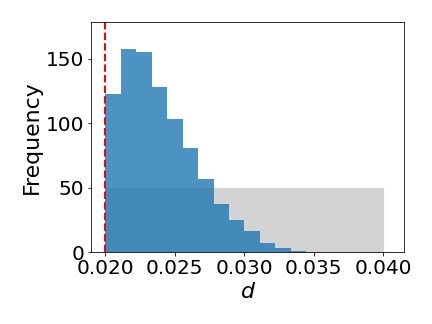}}
\hspace{5mm}
\subfloat[Parameter $e$]{\label{fig:e}\includegraphics[width = 50mm]{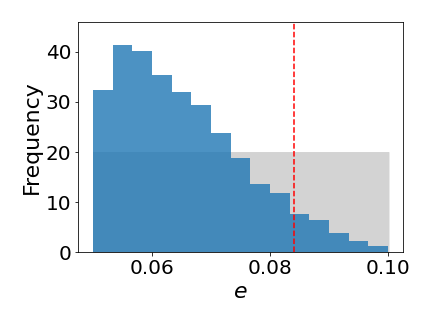}}\\
\subfloat[$\mu_{\phi}$]{\label{fig:d}\includegraphics[width = 50mm]{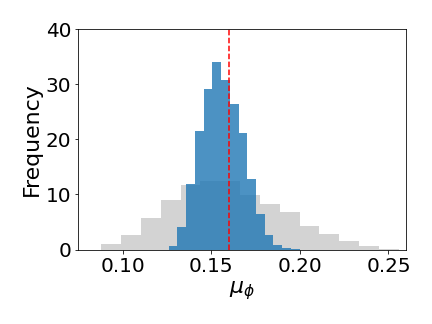}}
\hspace{5mm}
\subfloat[$\sigma_{\phi}$]{\label{fig:e}\includegraphics[width = 50mm]{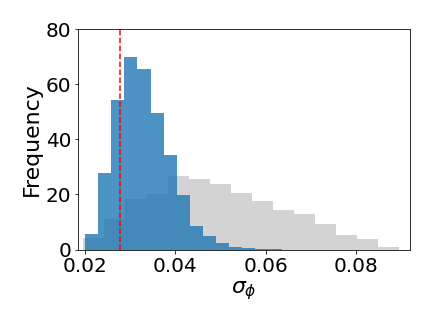}}
\caption{History matching results for the metaparameters for true model~1. Gray regions represent prior distributions, blue histograms are posterior distributions, and red vertical lines denote true values. Legend in (a) applies to all subplots.}
\label{meta_1}
\end{figure}

History matching results for the metaparameters for this case are displayed in Fig.~\ref{meta_1}. The grey regions indicate the prior distributions (which are uniform), the blue histograms represent the posterior distributions, and the red vertical dashed lines display the true values. The true values consistently fall within the posterior distributions. Substantial uncertainty reduction in the mean and standard deviation of the log-permeability field, and in the permeability anisotropy ratio, is clearly achieved (Fig.~\ref{meta_1}a, b and c). A degree of uncertainty reduction is observed in the metaparameter $d$ (Fig.~\ref{meta_1}d). The metaparameter $e$, however, is underestimated. The global sensitivity analysis (presented in SI) shows that saturation and pressure in the top layer of the storage aquifer are not sensitive to metaparameters $d$ and $e$. The results in Fig.~\ref{meta_1}d and e are consistent with this observation.

Recall that $d$ and $e$ relate porosity and permeability via an equation of the form $\phi = d \cdot \log k + e$ (Eq.~\ref{porosity}). It is thus instructive to consider the prior and posterior distributions for porosity directly. These are displayed in Fig.~\ref{meta_1}f and g. It is evident that, although history matching does not provide significant uncertainty reduction for $d$ and $e$ individually, it does reduce uncertainty in the mean ($\mu_\phi$) and standard deviation ($\sigma_\phi$) of porosity. Thus we conclude that it is these values, rather than the cell-by-cell relationship between porosity and permeability, for which the measured data are informative.

\begin{figure}[!ht]
\centering
\subfloat[Saturation at O1 in layer 1]{\label{fig:a}\includegraphics[width = 75mm]{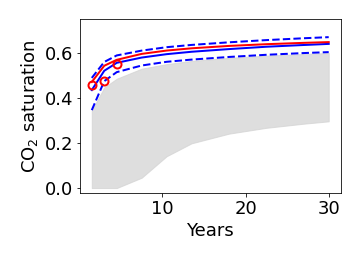}}
\hspace{2mm}
\subfloat[Saturation at O2 in layer 5]{\label{fig:b}\includegraphics[width = 75mm]{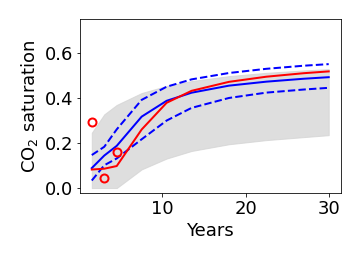}}
\\[1ex]
\subfloat[Saturation at O3 in layer 9]{\label{fig:c}\includegraphics[width = 75mm]{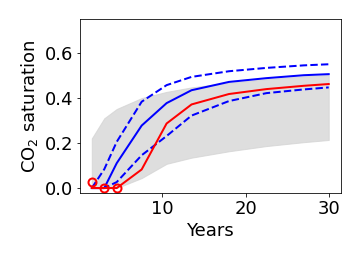}}
\hspace{2mm}
\subfloat[Saturation at O4 in layer 20]{\label{fig:d}\includegraphics[width = 75mm]{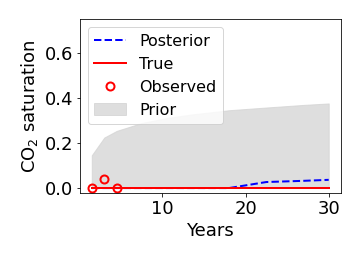}}
\\[1ex]
\caption{History matching results for saturation at four observation locations for true model~1. Grey regions represent the prior P$_{10}$--P$_{90}$ range, red circles and red lines denote observed and true data, and blue curves show the posterior P$_{10}$, P$_{50}$ and P$_{90}$ predictions from the surrogate model. Legend in (d) applies to all subplots.}
\label{Posterior_Sw}
\end{figure}

\begin{figure}[!ht]
\centering
\subfloat[Pressure at O1 in layer 1]{\label{fig:a}\includegraphics[width = 74mm]{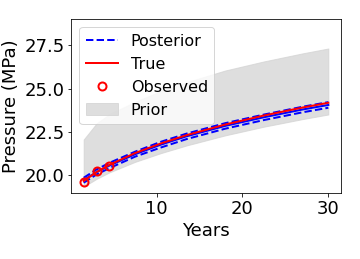}}
\hspace{2mm}
\subfloat[Pressure at O2 in layer 5]{\label{fig:b}\includegraphics[width = 74mm]{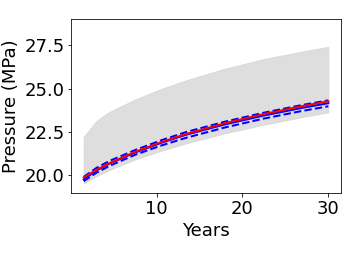}}
\\[1ex]
\subfloat[Pressure at O3 in layer 9]{\label{fig:c}\includegraphics[width = 74mm]{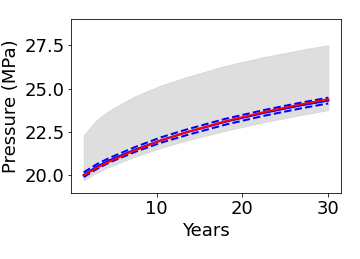}}
\hspace{2mm}
\subfloat[Pressure at O4 in layer 20]{\label{fig:d}\includegraphics[width = 74mm]{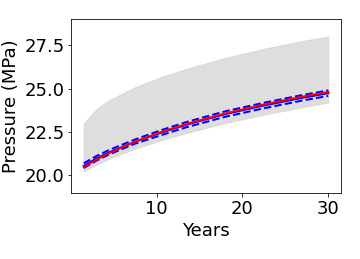}}
\\[1ex]
\caption{History matching results for pressure at four locations for true model~1. Grey regions represent the prior P$_{10}$--P$_{90}$ range, red circles and red lines denote observed and true data, and blue curves show the posterior P$_{10}$, P$_{50}$ and P$_{90}$ predictions from the surrogate model. Note that pressure data for history matching are only measured in layer~1. Legend in (a) applies to all subplots.}
\label{Posterior_P}
\end{figure}

History matching results for saturation and pressure at four observation locations are shown in Figs.~\ref{Posterior_Sw} and~\ref{Posterior_P}. In these figures, the grey regions represent the prior P$_{10}$--P$_{90}$ range, the red curves display the true data (simulation results without any error), and the red circles denote the observed data (which include measurement and resolution error). The blue solid curves represent the P$_{50}$ posterior results, while the blue dashed curves show the P$_{10}$ (lower) and P$_{90}$ (upper) posterior results. Significant uncertainty reduction in saturation is observed at different layers of the storage aquifer in Fig.~\ref{Posterior_Sw}. This is the case even though there is considerable data error (which is mostly due to model error), as is evident from the discrepancy between the red circles and the red curves. See, in particular, Fig.~\ref{Posterior_Sw}b. Pressure in the different layers also exhibits substantial uncertainty reduction (Fig.~\ref{Posterior_P}). This indicates that  pressure measurements in the top layer are highly informative in terms of the vertical pressure profile.

The shape and extent of the CO$_2$ saturation plumes, and the pressure fields within the storage aquifer, are of particular interest. We now display saturation plumes and pressure fields, after 30~years of injection, corresponding to prior and posterior models. Our approach for finding representative samples of these prior and posterior solutions is as follows. We first select, randomly, a total of 1000 prior geomodels (out of the 2000 geomodels used for training) and 1000 posterior geomodels (out of the 13,750 posterior geomodels accepted by MCMC). The saturation and pressure fields for these prior and posterior geomodels (at 30~years) are then generated using the surrogate model. We then apply a k-means clustering method to provide five clusters for each case. Finally, a k-medoids method is used to identify the `center' of each cluster. These five prior and five posterior (cluster center) solutions can be viewed as representative saturation and pressure fields. 

\begin{figure}[!ht]
\centering   
\subfloat[Prior 1]{\label{fig:a}\includegraphics[width=26mm]{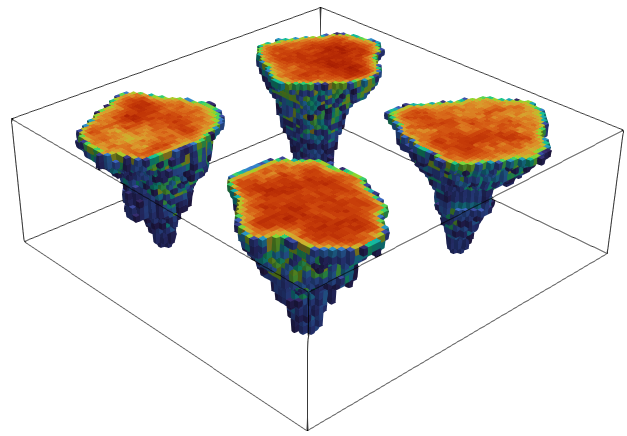}}
\hspace{1.5mm}
\subfloat[Prior 2]{\label{fig:b}\includegraphics[width=26mm]{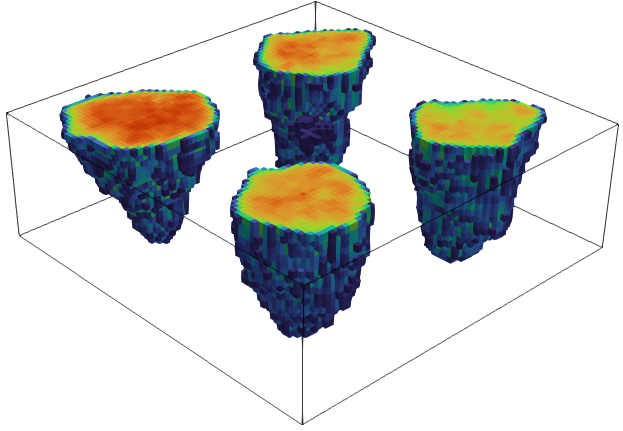}}
\hspace{1.5mm}
\subfloat[Prior 3]{\label{fig:c}\includegraphics[width=26mm]{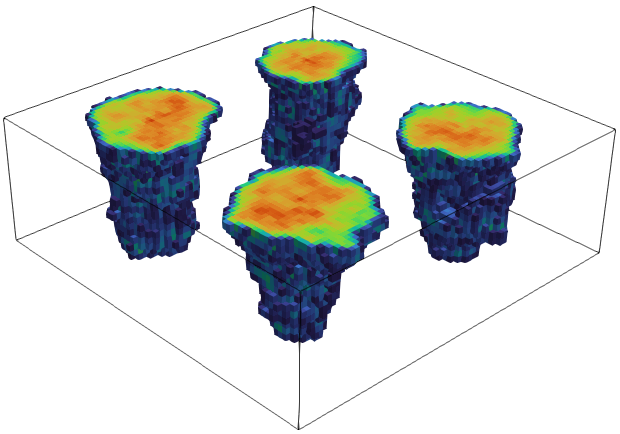}}
\hspace{1.5mm}
\subfloat[Prior 4]{\label{fig:b}\includegraphics[width=26mm]{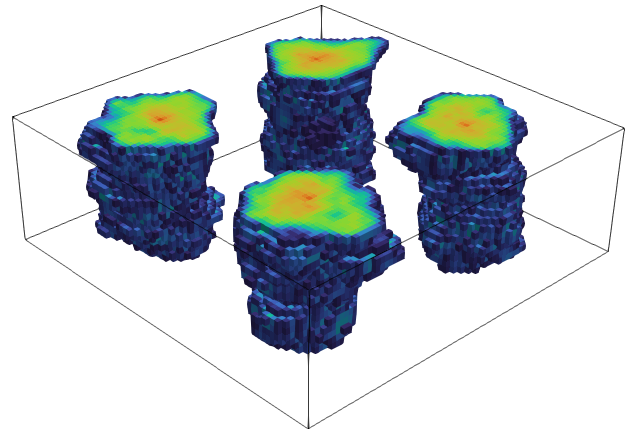}}
\hspace{1.5mm}
\subfloat[Prior 5]{\label{fig:c}\includegraphics[width=26mm]{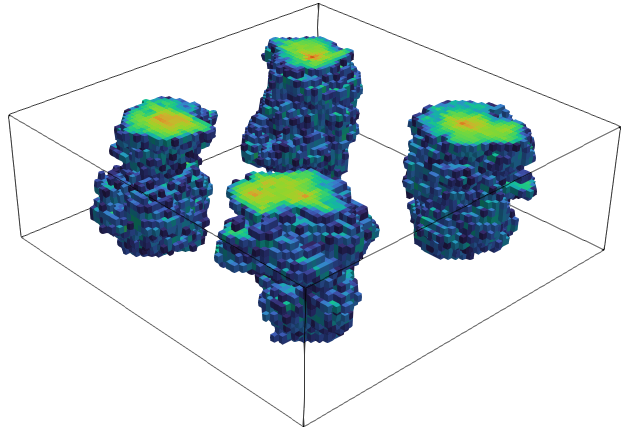}}
\includegraphics[width=4.5mm]{Figure/Test_Case/Sw_Scale.PNG}\\[1ex]
\subfloat[Posterior 1]{\label{fig:a}\includegraphics[width=26mm]{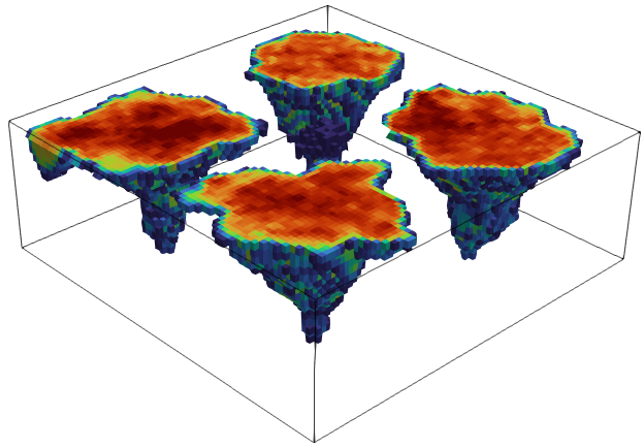}}
\hspace{1.5mm}
\subfloat[Posterior 2]{\label{fig:b}\includegraphics[width=26mm]{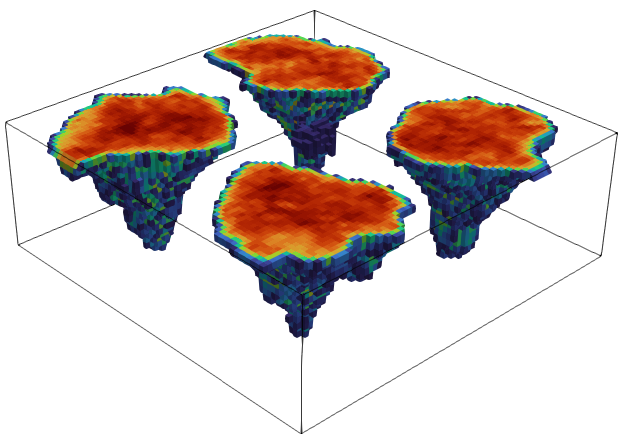}}
\hspace{1.5mm}
\subfloat[Posterior 3]{\label{fig:c}\includegraphics[width=26mm]{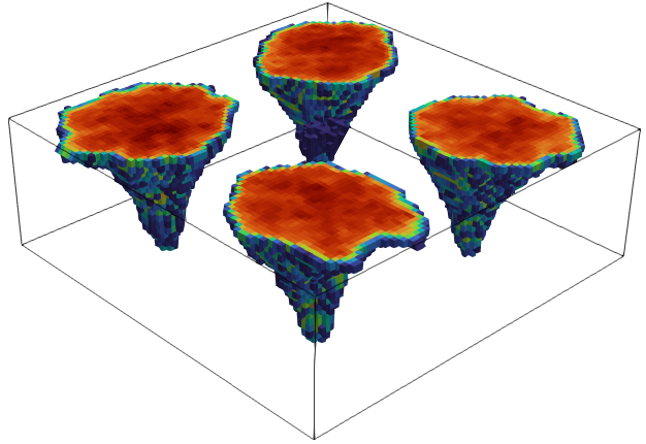}}
\hspace{1.5mm}
\subfloat[Posterior 4]{\label{fig:b}\includegraphics[width=26mm]{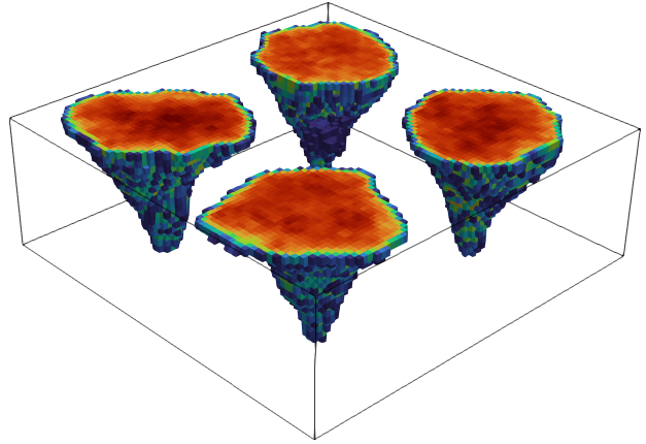}}
\hspace{1.5mm}
\subfloat[Posterior 5]{\label{fig:c}\includegraphics[width=26mm]{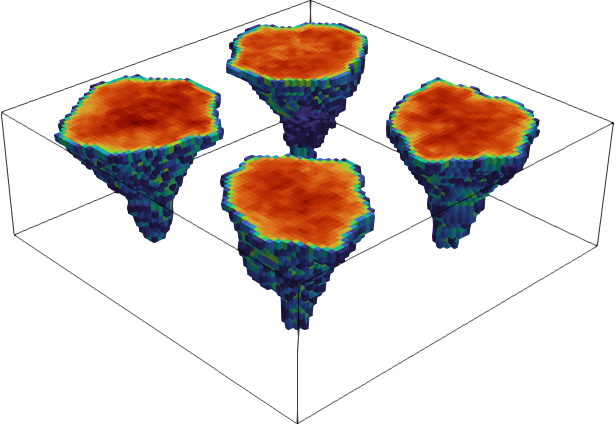}}
\includegraphics[width=4.5mm]{Figure/Test_Case/Sw_Scale.PNG}
\caption{Representative saturation plumes for prior geomodels (upper row) and posterior geomodels (lower row), for true model~1 at 30~years. True saturation field for this case shown in Fig.~\ref{saturations}a.}
\label{Prior_Posterior:Saturation}
\end{figure}

The resulting prior and posterior saturation fields, at 30~years, are shown in Fig.~\ref{Prior_Posterior:Saturation}. A high degree of variability, with significant differences in the shape and extent of the saturation plumes, is observed in the priors (top row). This is consistent with the variability shown earlier in Fig.~\ref{saturations}. The prior plume geometries vary from conical to cylindrical to irregular as we proceed from Fig.~\ref{Prior_Posterior:Saturation}a to e. The plume corresponding to true model~1 is shown in Fig.~\ref{saturations}a.
The plumes for the posterior geomodels (lower row of Fig.~\ref{Prior_Posterior:Saturation}) show much less variability than those for the prior geomodels, and they are relatively close to the true solution. This clearly illustrates the uncertainty reduction in plume geometry provided by our MCMC-based history matching procedure.

\begin{figure}[!ht]
\centering   
\subfloat[Prior 1]{\label{fig:a}\includegraphics[width=26mm]{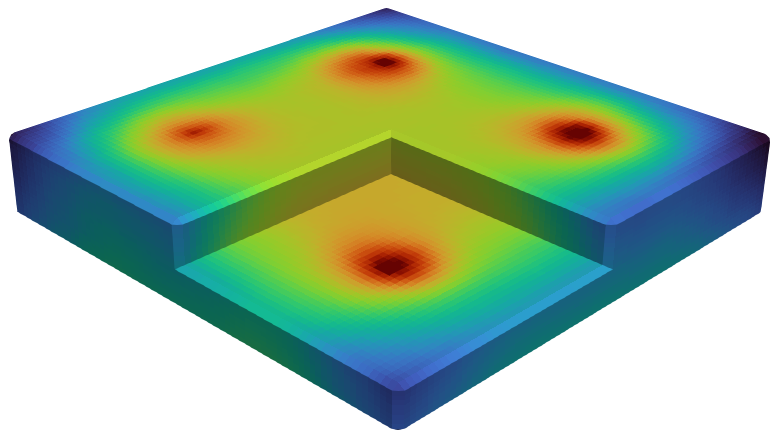}}
\hspace{1.5mm}
\subfloat[Prior 2]{\label{fig:b}\includegraphics[width=26mm]{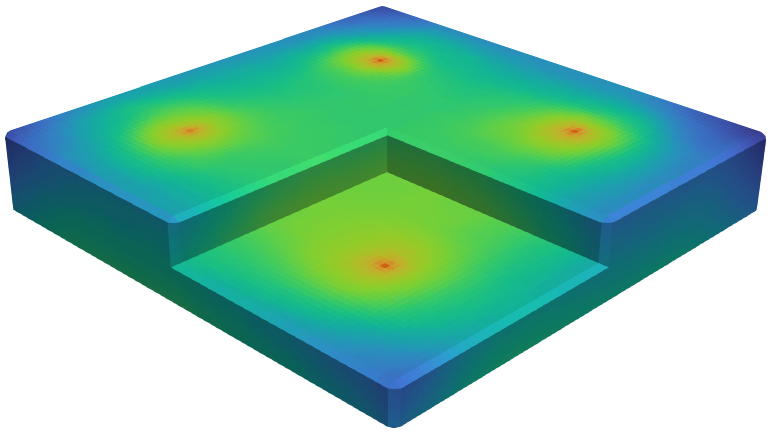}}
\hspace{1.5mm}
\subfloat[Prior 3]{\label{fig:c}\includegraphics[width=26mm]{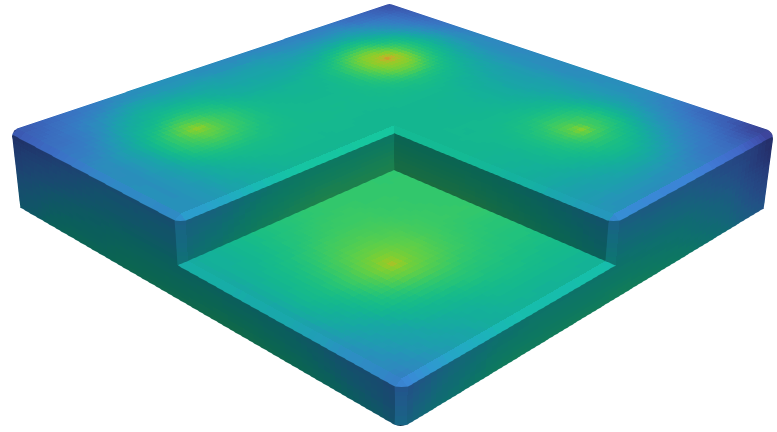}}
\hspace{1.5mm}
\subfloat[Prior 4]{\label{fig:b}\includegraphics[width=26mm]{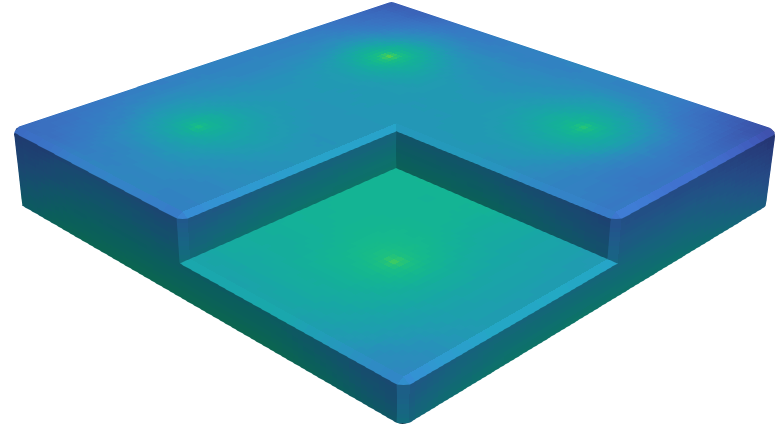}}
\hspace{1.5mm}
\subfloat[Prior 5]{\label{fig:c}\includegraphics[width=26mm]{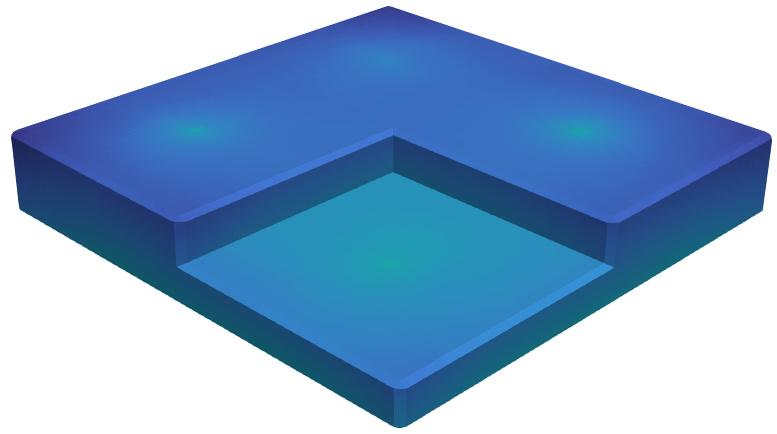}}
\includegraphics[width=4.5mm]{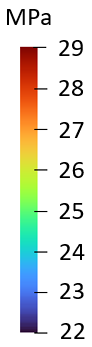}\\[1ex]
\subfloat[Posterior 1]{\label{fig:a}\includegraphics[width=26mm]{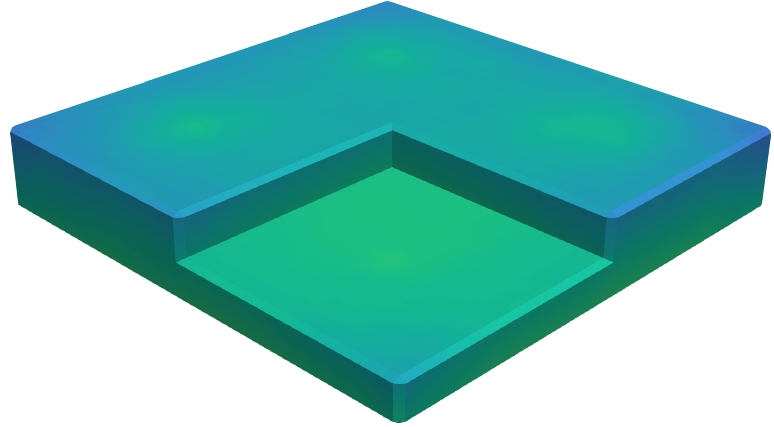}}
\hspace{1.5mm}
\subfloat[Posterior 2]{\label{fig:b}\includegraphics[width=26mm]{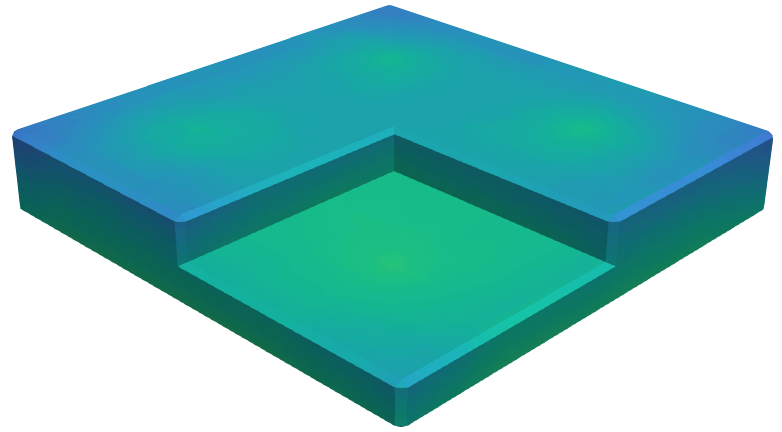}}
\hspace{1.5mm}
\subfloat[Posterior 3]{\label{fig:c}\includegraphics[width=26mm]{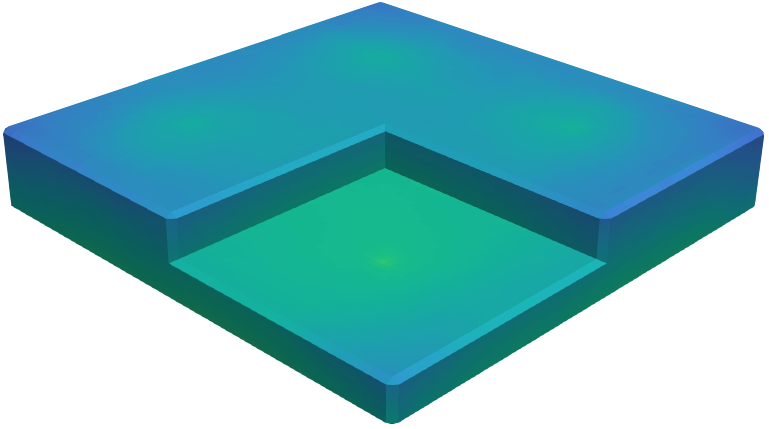}}
\hspace{1.5mm}
\subfloat[Posterior 4]{\label{fig:b}\includegraphics[width=26mm]{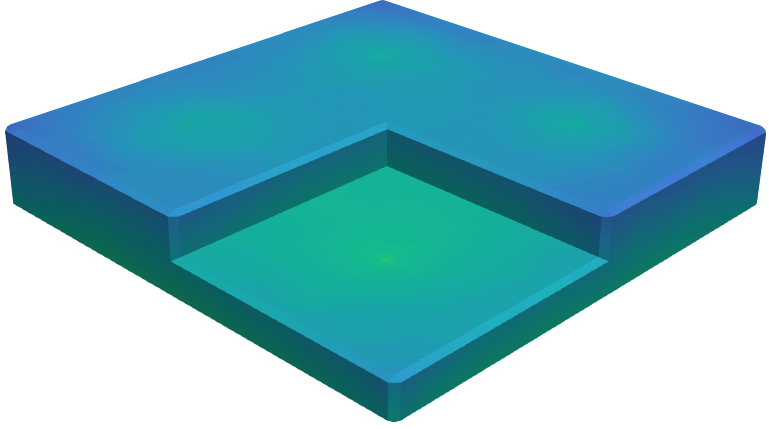}}
\hspace{1.5mm}
\subfloat[Posterior 5]{\label{fig:c}\includegraphics[width=26mm]{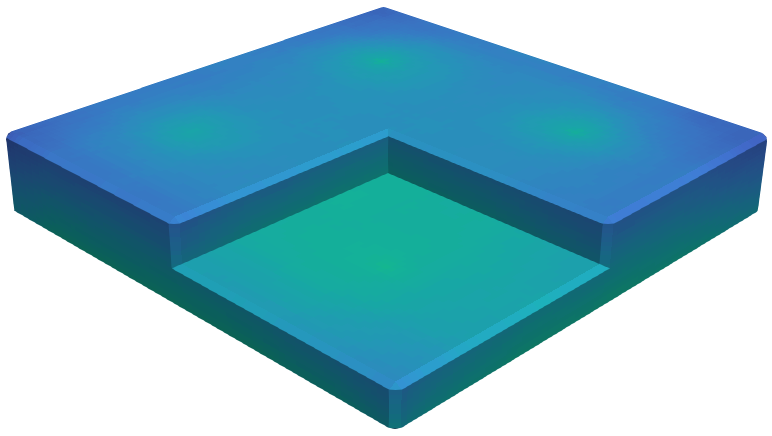}}
\includegraphics[width=4.5mm]{Figure/True_Model_1/P_Prior_Scale.PNG}
\caption{Representative pressure fields for prior geomodels (upper row) and posterior geomodels (lower row), for true model~1 at 30~years. True pressure field for this case shown in Fig.~\ref{pressures_1}a.}
\label{Prior_Posterior:Pressure}
\end{figure}

Analogous prior and posterior pressure solutions are shown in Fig.~\ref{Prior_Posterior:Pressure}. The true (simulated) pressure field for this case at 30~years is shown in Fig.~\ref{pressures_1}a. Again, we see a high degree of variability in the prior results (upper row of Fig.~\ref{Prior_Posterior:Pressure}), which is reduced considerably in the posterior results (lower row of Fig.~\ref{Prior_Posterior:Pressure}). The posterior pressure fields clearly resemble the true solution, again demonstrating the capabilities of the overall MCMC-based history matching procedure.

Additional history matching cases are presented in SI. There we show results for true model~2, along with results for both true models with Gaussian priors (rather than the uniform priors considered here). In all cases, the MCMC-based history matching procedure achieves substantial uncertainty reduction in the key metaparameters ($\mu_{\log k}$, $\sigma_{\log k}$ and $a_r$).

\section{Concluding Remarks} 
\label{Conclusions}
In this study, we extended the 3D recurrent R-U-Net surrogate model to handle geomodel realizations drawn from different geological scenarios. These scenarios are characterized by metaparameters, which in realistic cases are themselves uncertain. The metaparameters considered here are the horizontal correlation length, mean and standard deviation of log-permeability, permeability anisotropy ratio, and constants in the porosity-permeability relationship. For any combination of metaparameters, an infinite number of realizations can be generated. To accommodate this wide range of potential geomodels, the recurrent R-U-Net model was extended to include additional input channels. The GEOS flow simulator provided training data for the surrogate model. This training data comprised flow simulation results for 2000 random realizations, which were themselves characterized by different combinations of the six metaparameters. A history matching workflow based on a dimension-robust Markov chain Monte Carlo procedure was introduced to treat systems in which the geological scenario is uncertain. In the history matching, five metaparameters were considered to be uncertain (horizontal correlation length was fixed).

The flow setup involved a large-scale storage operation with injection of 4~Mt CO$_2$ per year, for 30~years, through four vertical wells. Surrogate model performance was assessed through error statistics and results for particular cases. For a set of 500 new (random) test cases, the surrogate model provided results with a median relative saturation error of 4.5\% and a median relative pressure error of 1.3\%. The ability of the model to provide accurate saturation and pressure predictions for geomodels from different geological scenarios was clearly demonstrated, as was the accuracy of the solutions at monitoring-well locations. 

History matching results were presented for a synthetic true model. The error considered in the history matching workflow included measurement error, model-resolution error, and surrogate-model error. The MCMC procedure required about 70,000 function evaluations to achieve convergence in the posterior distributions of the metaparameters. These computations are manageable using the surrogate model (which required 0.1~seconds per run on a Nvidia A100 GPU), but they would not be practical using high-fidelity flow simulation. Uncertainty reduction in key metaparameters was achieved using the MCMC-based procedure, and the resulting saturation plumes and pressure fields were shown to match the true model results much more closely than the prior simulations.

There are a number of topics that should be considered in future research in this area. The use of different input channels for the extended recurrent R-U-Net, involving both independent and correlated fields, could be tested. The current workflow can be extended to treat coupled flow and geomechanics problems in cases with uncertain metaparameters, including those associated with geomechanical properties such as Young's modulus. The (spatial) correlation lengths and variogram angles characterizing the log-permeability fields could also be included as metaparameters in the history matching. A multifidelity/multiphysics surrogate modeling framework could significantly accelerate the simulations and training required for coupled problems, and this should be investigated. It will also be of interest to assess the impact of different types and quality of observed data on posterior uncertainty. This will require appropriate error models, which will need to characterize a range of modeling and resolution errors. Finally, the application of the workflow to realistic cases should be pursued.

\section*{CRediT authorship contribution statement}
\textbf{Yifu Han}: Conceptualization, Methodology, Software, Visualization, Formal analysis, Writing -- original draft. \textbf{Fran\c cois P. Hamon}: GEOS software -- Coding and use. \textbf{Su Jiang}: Conceptualization, Software, Writing -- original draft. \textbf{Louis J. Durlofsky}: Supervision, Conceptualization, Resources, Formal analysis, Writing -- review \& editing.

\section*{Declaration of competing interest}
The authors declare that they have no known competing financial interests or personal relationships that could have appeared to influence the work reported in this paper.

\section*{Data availability}
The code used in this study will be made available on github when this paper is published. Please contact Yifu Han (yifu@stanford.edu) for earlier access.

\section*{Acknowledgements} 
We are grateful to the Stanford Center for Carbon Storage, Stanford Smart Fields Consortium, and TotalEnergies (through the FC-MAELSTROM project) for funding. We thank the SDSS Center for Computation for providing the computational resources used in this work. We also acknowledge timely assistance from GEOS developers at Lawrence Livermore National Laboratory, Stanford University, and TotalEnergies.

\section*{Supplementary Information} 
\section*{SI 1. SI overview}
\renewcommand{\thesection}{SI \arabic{section}} 
In this SI, we expand on the results presented in the main text. We first evaluate the correspondence in second-order ensemble statistics, over the full set of test cases, between GEOS simulation results and those from the surrogate model. The accuracy and timing of the extended recurrent R-U-Net surrogate model are then compared to those of an FNO model. Next, a variance-based sensitivity analysis is performed to assess the importance of the metaparameters and geological realizations in terms of pressure and saturation data (at particular locations and times). History matching results for a second true model are then presented. Finally, we consider cases with Gaussian-distributed priors (as opposed to the uniform priors used in earlier results). Posterior results for these priors are presented for true models~1 and 2.

\section*{SI 2. Surrogate model second-order statistics}
\renewcommand{\thesection}{SI \arabic{section}} 
\label{second_order}
In the main text, in Figs.~12 and 13, we compared ensemble statistics for saturation and pressure at observation locations between GEOS and the surrogate model. It is also of interest to evaluate surrogate model performance in terms of second-order statistics. \citet{zhong2019predicting}, for example, compared simulation and surrogate model results for the standard deviation of saturation at observation locations. Here, the covariance between different quantities of interest is considered. Specifically, we compute the covariance of saturation at observation well~O1 in layer~1 with saturation in all other layers at 30~years. The covariance between saturation at 30~years and pressure at each of the 10 time steps is also computed at O1 in layer~1. These second-order statistics, which can be important to capture for history matching, are computed for both GEOS and the surrogate model over the full set of 500 test cases.

Results are shown in Fig.~\ref{second_order}. The GEOS results are represented by the black solid curves and the surrogate model results by the red dashed curves. We see very close agreement between the two sets of results for covariance between saturation in different layers (Fig.~\ref{second_order}a) at 30~years, and reasonable correspondence in the pressure -- saturation covariance through time (Fig.~\ref{second_order}b). Some discrepancy is evident at early time in Fig.~\ref{second_order}b. Results of similar accuracy are observed at observation wells~O2, O3 and O4. The agreement in both the first-order statistics (P$_{10}$, P$_{50}$, P$_{90}$ responses, shown in the main paper) and the second-order statistics demonstrates that the surrogate model is able to provide accurate predictions while maintaining correlations between different variables.

\begin{figure}[!ht]
\centering   
\subfloat[Covariance between saturation in layer~1 and saturation in other layers at O1 (at 30~years)]
{\label{fig:a}\includegraphics[width = 75mm]{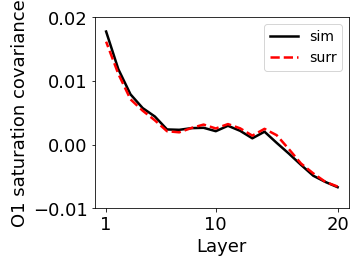}}
\hspace{3mm}
\subfloat[Covariance through time between saturation and pressure in layer~1 at O1]{\label{fig:d}\includegraphics[width = 80mm]{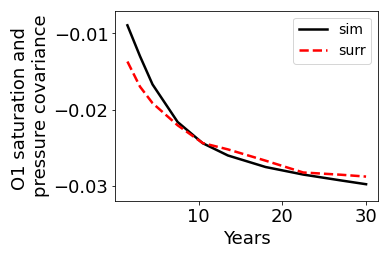}}\\
\caption{Test-case results for saturation and pressure-saturation covariance quantities at observation well~O1. Black solid curves represent GEOS results and red dashed curves show surrogate model results.}
\label{second_order}
\end{figure}

\section*{SI 3. Extended recurrent R-U-Net and FNO comparison}
\renewcommand{\thesection}{SI \arabic{section}} 
\label{fno}
\citet{wen2022u} showed that their Fourier neural operator (FNO)-based surrogate model provided more accurate pressure and saturation predictions than a convolutional neural network (CNN)-based surrogate model for a single-well axisymmetric injection scenario. In subsequent work, they introduced a 3D nested FNO surrogate model with local grid refinement around injection wells~\citep{wen2023real}. We now compare the extended recurrent R-U-Net surrogate model with the basic FNO model developed by \citet{wen2023real} for the specific 3D problem considered in this work. The nested, local grid refinement feature in their FNO method is not considered in this assessment. The same training and test samples are used to evaluate both methods. The normalization procedures described in Section~3 in the main paper for the input channels and for pressure and saturation are applied for both methods.

The optimized network parameters for the FNO surrogate model, denoted $\textbf{w}_{\mathrm{FNO}}^{\ast}$, are determined by minimizing the difference between simulation results and surrogate model predictions. The loss function, however, is different from that used for the extended recurrent R-U-Net. Specifically, in accordance with \citet{wen2022u,wen2023real}, the derivatives of pressure and saturation are included in the loss functions, as this was found to improve FNO performance. The resulting minimization problem can be expressed as
\begin{equation} \label{loss}
\begin{split}
\textbf{w}_{\mathrm{FNO}}^{\ast} &= \operatorname*{argmin}_{\textbf{w}_{\mathrm{FNO}}} \frac{1}{n_{smp}} \frac{1}{n_t} \sum_{i=1}^{n_{smp}}\sum_{t=1}^{n_t}\left(\frac{\|\hat{\textbf{x}}_{i}^{t} - \textbf{x}_{i}^{t}\|_2^2}{\|\textbf{x}_{i}^{t}\|_2^2} \right.
+ \lambda_1 \frac{\|\partial{\hat{\textbf{x}}_i^t} / \partial{x} - \partial{\textbf{x}_i^t} / \partial{x}\|_2^2}{\|\partial{\textbf{x}_i^t} / \partial{x}\|_2^2} \\
&\quad+ \lambda_2 \frac{\|\partial{\hat{\textbf{x}}_i^t} / \partial{y} - \partial{\textbf{x}_i^t} / \partial{y}\|_2^2}{\|\partial{\textbf{x}_i^t} / \partial{y}\|_2^2} 
+ \left. \lambda_3 \frac{\|\partial{\hat{\textbf{x}}_i^t} / \partial{z} - \partial{\textbf{x}_i^t} / \partial{z}\|_2^2}{\|\partial{\textbf{x}_i^t} / \partial{z}\|_2^2}\right).
\end{split}
\end{equation}
\noindent Here $n_{smp}$ = 2000 is the number of training samples, $n_t$ = 10 is the number of surrogate model time steps, $\textbf{x}_{i}^{t}$ and $\hat{\textbf{x}}_{i}^{t} \in \mathbb{R}^{n_x \times n_y \times n_z}$ represent the normalized pressure or saturation from high-fidelity flow simulation (GEOS) and the surrogate model at time step $t$ for training sample $i$, respectively, and $\partial{\textbf{x}}_i^t / \partial{x}$, $\partial{\textbf{x}}_i^t / \partial{y}$ and $\partial{\textbf{x}}_i^t / \partial{z}$ (and $\partial{\hat{\textbf{x}}_i^t} / \partial{x}$, $\partial{\hat{\textbf{x}}_i^t} / \partial{y}$ and $\partial{\hat{\textbf{x}}_i^t} / \partial{z}$) are the partial derivatives of $\textbf{x}_{i}^{t}$ (and $\hat{\textbf{x}}_{i}^{t}$). The quantities $\lambda_1$, $\lambda_2$ and $\lambda_3$ are weighting factors for the derivative terms. Recall that $n_x=n_y=80$ and $n_z=20$ for the storage aquifer models, where $n_x$, $n_y$ and $n_z$ are the number of cells in each direction.

Separate surrogate models are trained for pressure and saturation (this is also the case with the extended recurrent R-U-Net). For both networks a batch size of 32 is used. Consistent with \citet{wen2022u,wen2023real}, the initial learning rate is set to 0.001. The learning rate is then reduced by a factor of 0.9 every 5~epochs. We set the weighting factors ($\lambda_1$, $\lambda_2$ and $\lambda_3$) for the derivative terms to be 0.5, as suggested by \citet{wen2022u}. Each of the two trainings requires 121~hours (this corresponds to 120~epochs), for a total serial training time of 242~hours. As noted in the main paper, training times for the extended recurrent R-U-Net are 8~hours for pressure and 13~hours for saturation, for a total serial training time of 21~hours. The training of the pressure and saturation networks, for both surrogate models, could be done in parallel, which would reduce wall-clock training times to 121~hours and 13~hours.

FNO surrogate model evaluations for the full solution (both saturation and pressure) require 0.01~seconds on a single Nvidia A100 GPU. For the extended recurrent R-U-Net, each evaluation of the full solution requires 0.1~seconds with the same GPU. The observation that the FNO model is faster than a CNN-based model (such as the extended recurrent R-U-Net) in terms of evaluations, but requires more time to train, was also noted by \citet{wen2022u}.

The test-case relative errors for saturation and pressure are computed, for both FNO and the extended recurrent R-U-Net, using Eqs.~14 and 15 in the main paper. The FNO median relative errors for pressure and saturation, over the 500 test cases, are 2.2\% and 10.9\%, respectively. The relative errors for both methods are presented in Table~\ref{comparsion} below. The extended recurrent R-U-Net surrogate model provides more accurate predictions than FNO for the four-well, 3D problem considered in this work (using the training strategy described above). We do not claim this as a general finding -- it is possible that with more hyperparameter tuning and/or architecture modification the FNO method would provide better accuracy than the extended recurrent R-U-Net. These results indicate, however, that the extended recurrent R-U-Net is a reasonable choice for the problem at hand.

\bigskip
\begin{table}[!ht]
\begin{center}
\footnotesize
\caption{Median relative pressure and saturation errors for extended recurrent R-U-Net and FNO models}
\label{comparsion}
\renewcommand{\arraystretch}{1.2} 
\begin{tabular}{ c c c} 
\hline
\textbf{Surrogate model} & \textbf{Pressure median error} & \textbf{Saturation median error} \\
\hline
Extended recurrent R-U-Net  & 1.3\% & 4.5\% \\  
FNO  & 2.2\% & 10.9\% \\  
\hline
\end{tabular}
\end{center}
\end{table}

\section*{SI 4. Variance-based sensitivity analysis}
\renewcommand{\thesection}{SI \arabic{section}} 
\label{gsa}
The geomodels of the storage aquifer constructed during history matching are generated from the metaparameters and the PCA representation. Here we apply the variance-based sensitivity analysis of~\citet{sobol2001global} to evaluate the importance of the metaparameters and the PCA latent variables for particular data quantities. Specifically, we consider pressure and saturation in observation well~O1 in layer~1 (this is the layer where pressure is measured for history matching) at 3~years. The total-effect sensitivity index for the five metaparameters ($\mu_{\log k}$, $\sigma_{\log k}$, $a_r$, $d$, $e$) and for the PCA latent variables is computed. Note that we consider the PCA latent variables collectively, as a single variable $\bm{\xi}$. Thus, in the context of this analysis, the sensitivity to $\bm{\xi}$ indicates the importance of the particular realization, as opposed to the importance of the metaparameters that define the geological scenario from which it is drawn. 

The total-effect sensitivity index quantifies the variance contribution of each uncertain parameter over the variance of all samples. For uncertain parameter $i$ ($i = 1, \dots, 6$), denoted $S_i$, this quantity is computed using
\begin{equation} \label{total_effect}
S_{i} = \frac{\mathrm{E}_{\textbf{u}_{\sim i}} \left(\mathrm{Var}_{u_i}\left(v|\textbf{u}_{\sim i}\right)\right)}{\mathrm{Var}_{\textbf{u}}\left(v\right)},
\end{equation}
\noindent where $\textbf{u} = [\mu_{\log k}, \sigma_{\log k}, a_r, d, e, \bm{\xi}]$ is the set of uncertain parameters, $u_i$ denotes the uncertain parameter $i$, $\textbf{u}_{\sim i}$ ($\textbf{u}_{\sim i}$ = [$u_1$, ..., $u_{i-1}$, $u_{i+1}$, ..., $u_6$]) represents the set of all uncertain parameters except $u_i$, and $v$ is pressure or saturation at O1 in layer~1 at 3~years generated from the surrogate model. The variance in the numerator of Eq.~\ref{total_effect}, $\mathrm{Var}_{u_i}\left(v|\textbf{u}_{\sim i}\right)$, represents the variance of $v$ over all possible values of $u_i$, conditioned to a fixed $\textbf{u}_{\sim i}$. The expectation, $\mathrm{E}_{\textbf{u}_{\sim i}}\left(\mathrm{Var}_{u_i}\left(v|\textbf{u}_{\sim i}\right)\right)$, denotes the expectation of this variance over all possible values of $\textbf{u}_{\sim i}$. The quantity $\mathrm{E}_{\textbf{u}_{\sim i}}\left(\mathrm{Var}_{u_i}\left(v|\textbf{u}_{\sim i}\right)\right)$ thus represents the variance of pressure or saturation resulting from variations in a specific uncertain parameter, while considering all possible values of the other uncertain parameters. The denominator in Eq.~\ref{total_effect} represents the variance of $v$ over all samples of $\textbf{u}$, i.e., it quantifies the overall variability in pressure or saturation. Pressure or saturation is more sensitive to uncertain parameters with higher total-effect sensitivity index. For full details on the total-effect sensitivity index computation, please see \citet{saltelli2010variance}.

In this study, a total of 262,144 samples of $\textbf{u}$ are generated and evaluated. This number of samples ensures the convergence of the total-effect sensitivity index computation (note that this type of evaluation would be extremely time consuming using high-fidelity simulation). The horizontal correlation length is set to 1500~m, which is consistent with the value specified for the two true models used for history matching. The metaparameters ($\mu_{\log k}$, $\sigma_{\log k}$, $a_r$, $d$, $e$) are sampled from their prior ranges. The PCA latent variables, i.e., the components of $\bm{\xi}$, are sampled individually from $\mathcal{N}$(0, 1).

\begin{figure}[H]
\centering   
\subfloat[Total-effect index for pressure]{\label{fig:a}\includegraphics[width = 75mm]{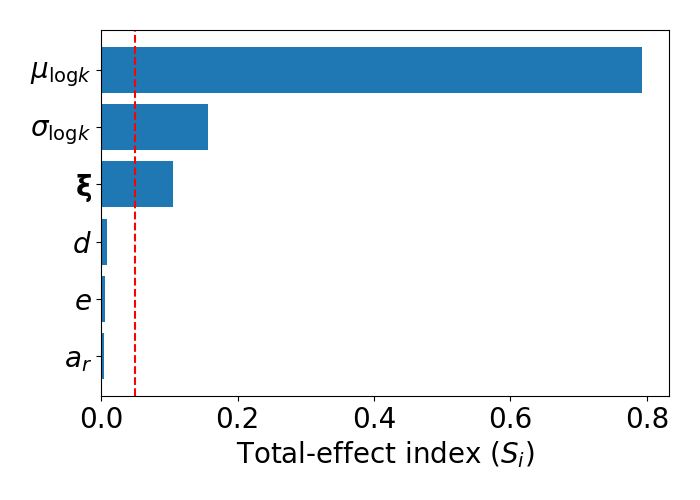}}
\subfloat[Total-effect index for saturation]{\label{fig:d}\includegraphics[width = 75mm]{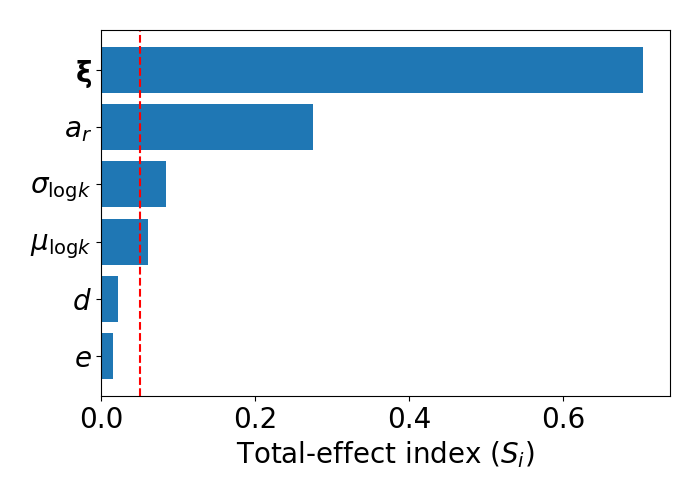}}\\
\caption{Total-effect sensitivity indices for pressure and saturation at observation well~O1 in layer~1 at 3~years. The red vertical dashed lines denote a cutoff value of 0.05.}
\label{gsa_total}
\end{figure}

The total-effect sensitivity indices, for pressure and saturation at observation well~O1 in layer~1 at 3~years, are shown in Fig.~\ref{gsa_total}. The red vertical dashed line displays a cutoff value of 0.05 -- parameters with a sensitivity index larger than this value are deemed important~\citep{zhang2015sobol}. The results indicate that both pressure and saturation (at this particular time and location) are sensitive to the mean and standard deviation of the log-permeability field and to the geological realization. Interestingly, pressure is most sensitive to $\mu_{\log k}$ while saturation is most sensitive to $\bm{\xi}$. In addition, pressure is insensitive to the permeability anisotropy ratio ($a_r$) while saturation is sensitive to this quantity (consistent with the results in Fig.~10 in the main text). Neither quantity is sensitive to the constants in the porosity-permeability relationship ($d$ and $e$). Similar results are observed for pressure and saturation in layer~1 at the other observation well locations at 3~years. These sensitivities will in general depend on the particular quantity (and location and time) considered.

The fact that observation well data are not sensitive to $d$ and $e$ suggests that these data will not lead to significant uncertainty reduction in $d$ and $e$. This expectation is consistent with the history matching results shown in Fig.~14 in the main text (and with the results presented later in this SI). Specifically, in our history matching results, we observe significant uncertainty reduction for $\mu_{\log k}$, $\sigma_{\log k}$ and $a_r$ (to which O1 data are sensitive), but only a small amount of uncertainty reduction for $d$ and $e$. We note finally that, in cases where correlation length was also considered to be uncertain, pressure and saturation at observation well~O1 in layer~1 at 3~years were found to be insensitive to the correlation length. This may be because the correlation lengths considered were in all cases much larger than the distance between injection wells and the closest observation well.

\section*{SI 5. History matching results for true model~2}
\renewcommand{\thesection}{SI \arabic{section}} 
\label{`true' model 2}
We now present history matching results for a second synthetic true model. True model~2 is shown in Fig.~9b in the main paper. For this true model, we set $l_h=1500$~m, $\mu_{\log k}=2.44$, $\sigma_{\log k} = 1.37$, $a_r=0.14$, $d=0.025$ and $e=0.074$. The hierarchical MCMC-based procedure requires 68,308 iterations to  converge in this case. A total of 14,500 metaparameter sets (corresponding to 14,500 geomodel realizations) are accepted during history matching. The acceptance rate is thus about 21\%.

\begin{figure}[H]
\centering   
\subfloat[$\mu_{\mathrm{log}k}$]{\label{fig:a}\includegraphics[width = 50mm]{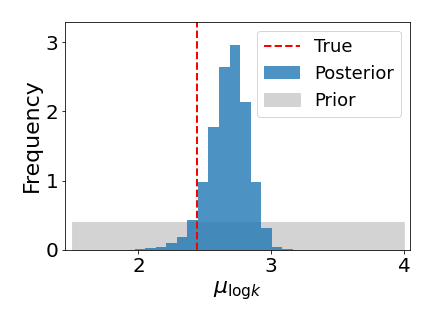}}
\hspace{2mm}
\subfloat[$\sigma_{\mathrm{log}k}$]{\label{fig:b}\includegraphics[width = 50mm]{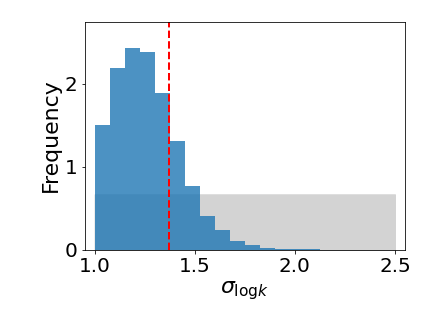}}
\hspace{2mm}
\subfloat[$\log_{10}(a_r)$]{\label{fig:d}\includegraphics[width = 50mm]{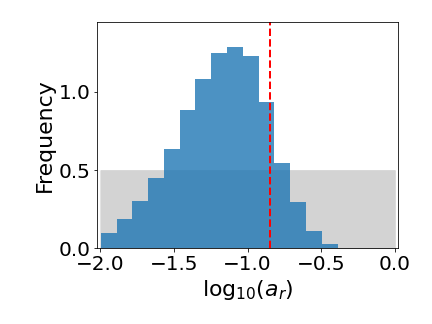}}\\
\subfloat[Parameter $d$]{\label{fig:d}\includegraphics[width = 50mm]{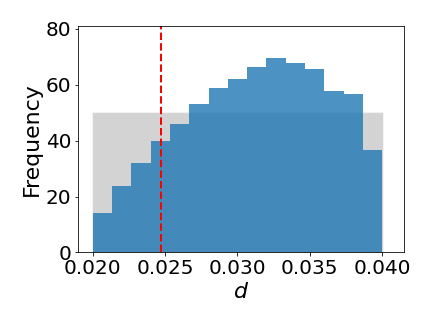}}
\hspace{5mm}
\subfloat[Parameter $e$]{\label{fig:e}\includegraphics[width = 50mm]{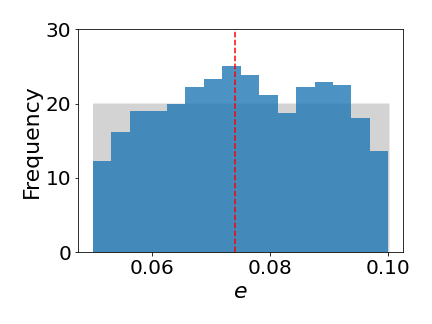}}\\
\subfloat[$\mu_{\phi}$]{\label{fig:d}\includegraphics[width = 50mm]{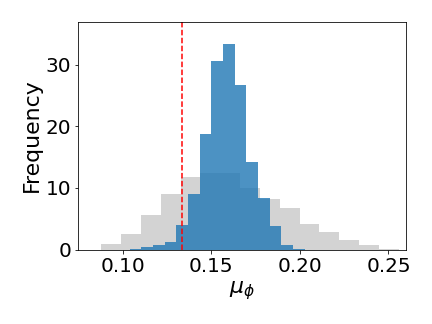}}
\hspace{5mm}
\subfloat[$\sigma_{\phi}$]{\label{fig:e}\includegraphics[width = 50mm]{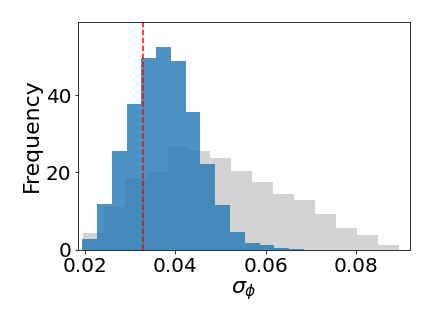}}
\caption{History matching results for the metaparameters for true model~2. Gray regions represent prior distributions, blue histograms are posterior distributions, and red vertical lines denote true values. Legend in (a) applies to all subplots.}
\label{meta_2}
\end{figure}

Prior and posterior results for the metaparameters are shown in Fig.~\ref{meta_2}. We again observe significant uncertainty reduction in the key metaparameters ($\mu_{\log k}$, $\sigma_{\log k}$ and $a_r$). The mean of log-permeability is slightly overestimated, though the true value does fall within the posterior distribution. Because the total covariance (Eq.~19 in the main text) is relatively large, the observations are not always as informative as might be expected. Consistent with our observations for true model~1, there is again little uncertainty reduction in metaparameters $d$ and $e$ individually. We see in Fig.~\ref{meta_2}g that the standard deviation of porosity is relatively accurate, while the mean of porosity (Fig.~\ref{meta_2}f) is overestimated.

\begin{figure}[H]
\centering   
\subfloat[Prior 1]{\label{fig:a}\includegraphics[width=26mm]{Figure/True_Model_1/Prior_Sw_Cluster_1.PNG}}
\hspace{1.5mm}
\subfloat[Prior 2]{\label{fig:b}\includegraphics[width=26mm]{Figure/True_Model_1/Prior_Sw_Cluster_2.PNG}}
\hspace{1.5mm}
\subfloat[Prior 3]{\label{fig:c}\includegraphics[width=26mm]{Figure/True_Model_1/Prior_Sw_Cluster_3.PNG}}
\hspace{1.5mm}
\subfloat[Prior 4]{\label{fig:b}\includegraphics[width=26mm]{Figure/True_Model_1/Prior_Sw_Cluster_4.PNG}}
\hspace{1.5mm}
\subfloat[Prior 5]{\label{fig:c}\includegraphics[width=26mm]{Figure/True_Model_1/Prior_Sw_Cluster_5.PNG}}
\includegraphics[width=4.5mm]{Figure/Test_Case/Sw_Scale.PNG}\\[1ex]
\subfloat[Posterior 1]{\label{fig:a}\includegraphics[width=26mm]{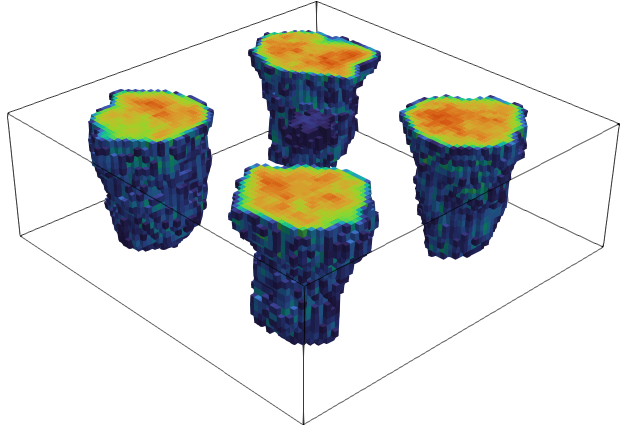}}
\hspace{1.5mm}
\subfloat[Posterior 2]{\label{fig:b}\includegraphics[width=26mm]{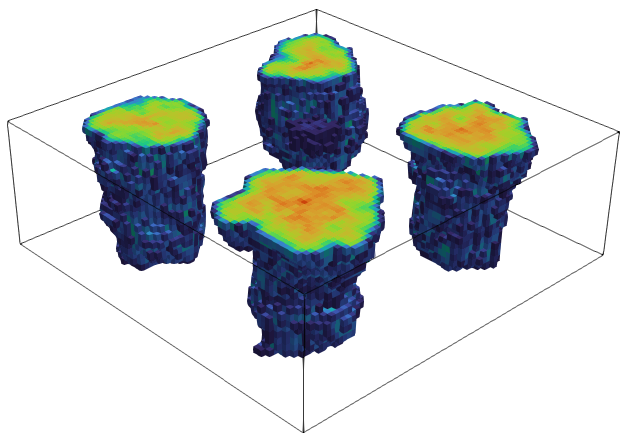}}
\hspace{1.5mm}
\subfloat[Posterior 3]{\label{fig:c}\includegraphics[width=26mm]{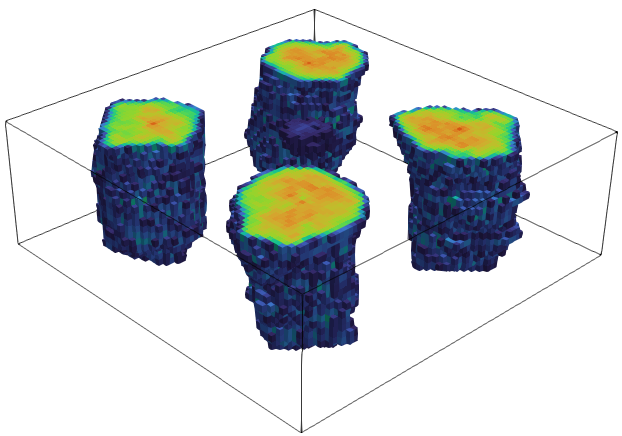}}
\hspace{1.5mm}
\subfloat[Posterior 4]{\label{fig:b}\includegraphics[width=26mm]{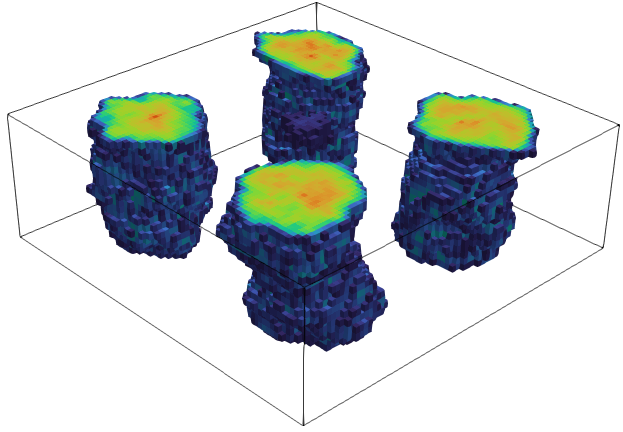}}
\hspace{1.5mm}
\subfloat[Posterior 5]{\label{fig:c}\includegraphics[width=26mm]{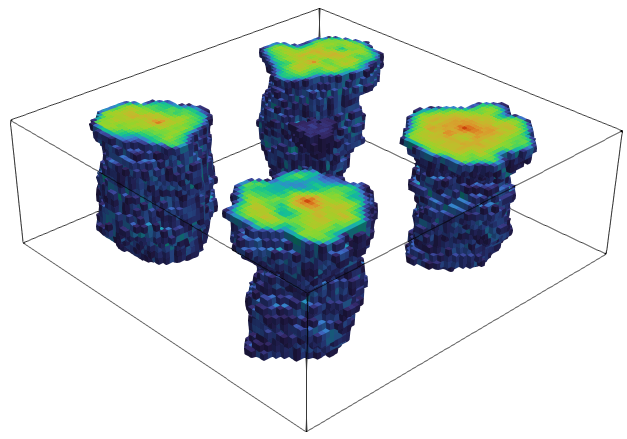}}
\includegraphics[width=4.5mm]{Figure/Test_Case/Sw_Scale.PNG}
\caption{Representative saturation plumes for prior geomodels (upper row) and posterior geomodels (lower row), for true model~2 at 30~years. True saturation field for this case shown in Fig.~10b in the main paper.}
\label{Prior_Posterior:Saturation_2}
\end{figure}

\begin{figure}[H]
\centering   
\subfloat[Prior 1]{\label{fig:a}\includegraphics[width=26mm]{Figure/True_Model_1/Prior_P_Cluster_1.PNG}}
\hspace{1.5mm}
\subfloat[Prior 2]{\label{fig:b}\includegraphics[width=26mm]{Figure/True_Model_1/Prior_P_Cluster_2.PNG}}
\hspace{1.5mm}
\subfloat[Prior 3]{\label{fig:c}\includegraphics[width=26mm]{Figure/True_Model_1/Prior_P_Cluster_3.PNG}}
\hspace{1.5mm}
\subfloat[Prior 4]{\label{fig:b}\includegraphics[width=26mm]{Figure/True_Model_1/Prior_P_Cluster_4.PNG}}
\hspace{1.5mm}
\subfloat[Prior 5]{\label{fig:c}\includegraphics[width=26mm]{Figure/True_Model_1/Prior_P_Cluster_5.PNG}}
\includegraphics[width=4.5mm]{Figure/True_Model_1/P_Prior_Scale.PNG}\\[1ex]
\subfloat[Posterior 1]{\label{fig:a}\includegraphics[width=26mm]{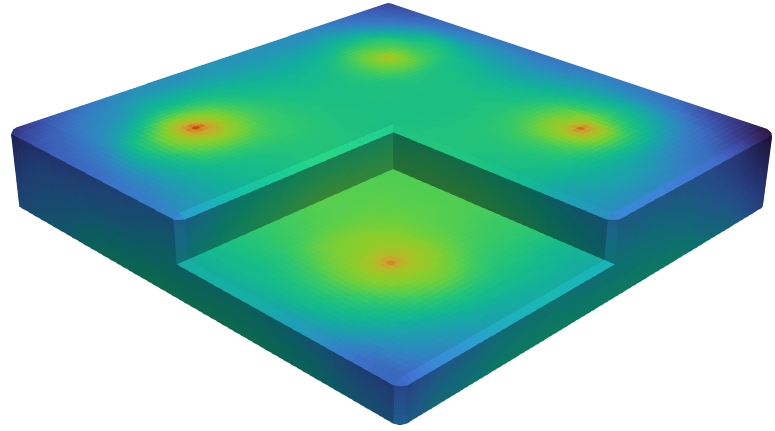}}
\hspace{1.5mm}
\subfloat[Posterior 2]{\label{fig:b}\includegraphics[width=26mm]{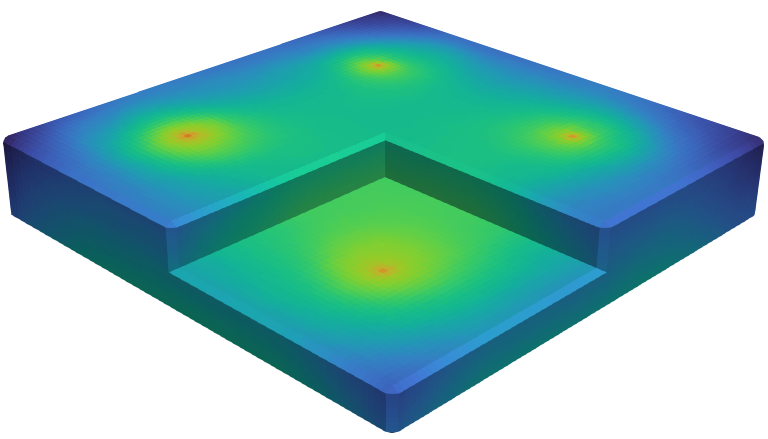}}
\hspace{1.5mm}
\subfloat[Posterior 3]{\label{fig:c}\includegraphics[width=26mm]{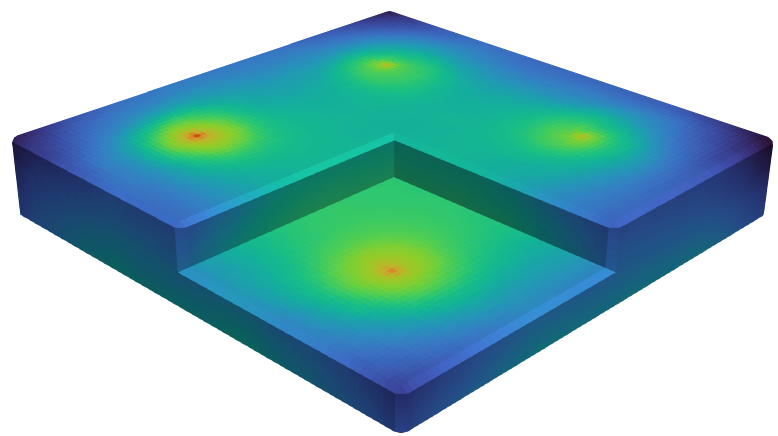}}
\hspace{1.5mm}
\subfloat[Posterior 4]{\label{fig:b}\includegraphics[width=26mm]{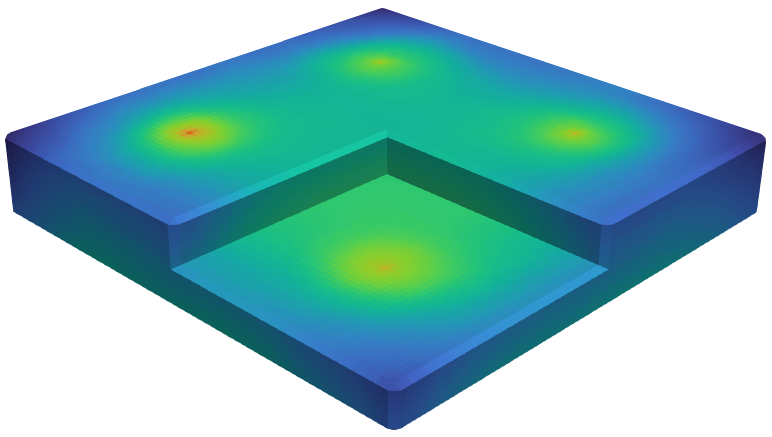}}
\hspace{1.5mm}
\subfloat[Posterior 5]{\label{fig:c}\includegraphics[width=26mm]{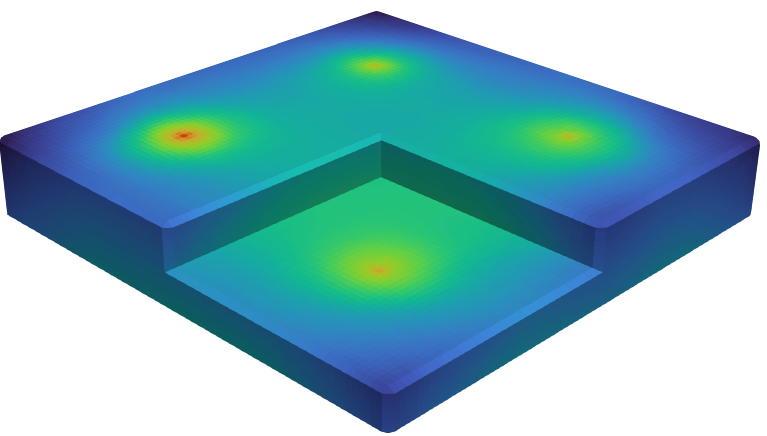}}
\includegraphics[width=4.5mm]{Figure/True_Model_1/P_Prior_Scale.PNG}
\caption{Representative pressure fields for prior geomodels (upper row) and posterior geomodels (lower row), for true model~2 at 30~years. True pressure field for this case shown in Fig.~11b in the main paper.}
\label{Prior_Posterior:Pressure_2}
\end{figure}

Representative prior and posterior saturation and pressure fields at 30~years are presented in Figs.~\ref{Prior_Posterior:Saturation_2} and \ref{Prior_Posterior:Pressure_2}. The true (simulated) saturation field for this case, shown in Fig.~10b in the main paper, exhibits near-cylindrical plumes around all four injectors. The posterior saturation fields generally display this geometry (this is particularly evident in Fig.~\ref{Prior_Posterior:Saturation_2}f, g and h). There is less variability in the posterior saturation plumes than in the priors, as would be expected. Similar observations apply to the pressure fields in Fig.~\ref{Prior_Posterior:Pressure_2}. These solutions again show reasonable agreement with the true model~2 pressure field shown in Fig.~11b in the main paper (note the scales differ between Fig.~\ref{Prior_Posterior:Pressure_2} and Fig.~11 in the main paper). 

It is clear from the results for both true models that substantial uncertainty reduction is achieved by history matching to monitoring well data. The amount of uncertainty reduction will vary by case and will additionally depend on the amount, type and quality (quantified in terms of error) of observation data collected. 

\section*{SI 6. History matching results using Gaussian-distributed priors}
\renewcommand{\thesection}{SI \arabic{section}} 
\label{prior}
Finally, we apply the hierarchical MCMC-based history matching method for cases with Gaussian-distributed priors. The mean of the Gaussian prior distribution for each metaparameter is taken to be the mid-range value of the uniform distribution used previously. The standard deviations for the Gaussian priors are 0.5 for $\mu_{\log k}$ and $\sigma_{\log k}$, 0.25 for $\log_{10} a_r$, 0.005 for $d$, and 0.017 for $e$. 

\begin{figure}[H]
\centering   
\subfloat[$\mu_{\mathrm{log}k}$]{\label{fig:a}\includegraphics[width = 50mm]{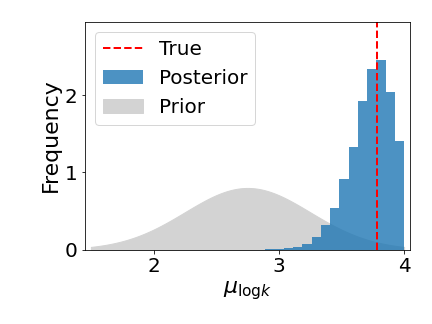}}
\hspace{2mm}
\subfloat[$\sigma_{\mathrm{log}k}$]{\label{fig:b}\includegraphics[width = 50mm]{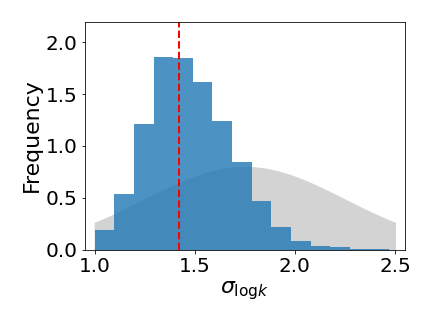}}
\hspace{2mm}
\subfloat[$\log_{10}(a_r)$]{\label{fig:d}\includegraphics[width = 50mm]{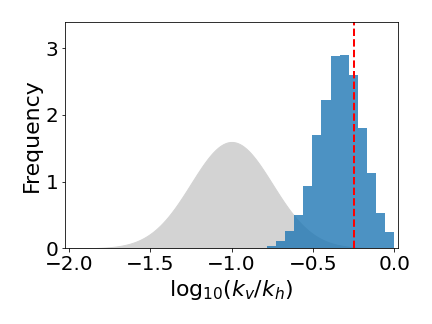}}\\
\subfloat[Parameter $d$]{\label{fig:d}\includegraphics[width = 50mm]{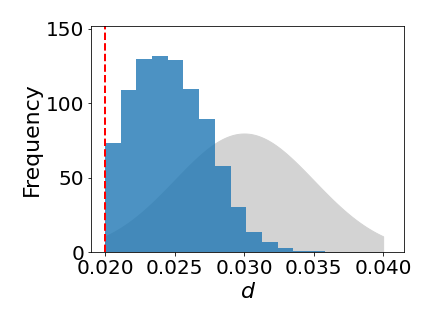}}
\hspace{5mm}
\subfloat[Parameter $e$]{\label{fig:e}\includegraphics[width = 50mm]{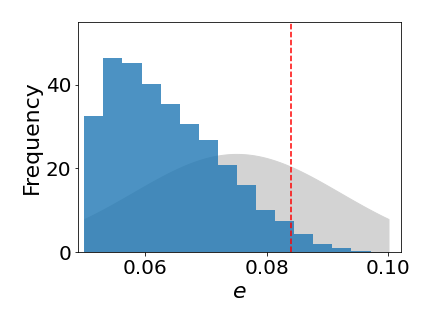}}\\
\subfloat[$\mu_{\phi}$]{\label{fig:d}\includegraphics[width = 50mm]{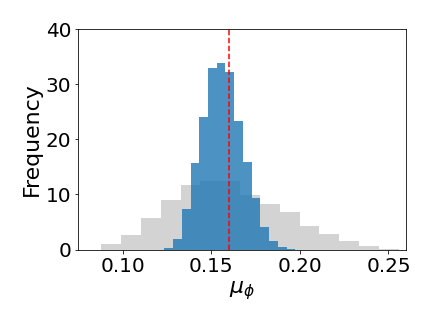}}
\hspace{5mm}
\subfloat[$\sigma_{\phi}$]{\label{fig:e}\includegraphics[width = 50mm]{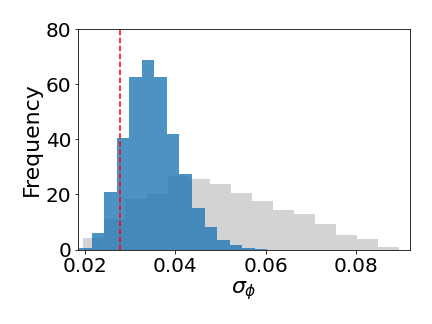}}
\caption{History matching results for the metaparameters for true model~1. Prior distributions of metaparameters are Gaussian in this case. Gray regions represent prior distributions, blue histograms are posterior distributions, and red vertical lines denote true values. Legend in (a) applies to all subplots.}
\label{meta_1_prior}
\end{figure}

\begin{figure}[H]
\centering   
\subfloat[$\mu_{\mathrm{log}k}$]{\label{fig:a}\includegraphics[width = 50mm]{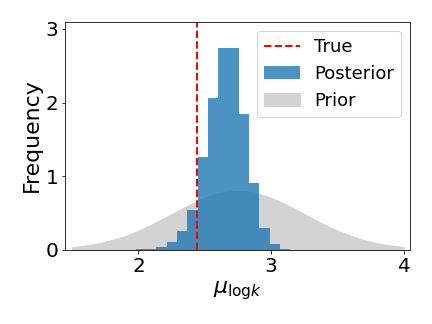}}
\hspace{2mm}
\subfloat[$\sigma_{\mathrm{log}k}$]{\label{fig:b}\includegraphics[width = 50mm]{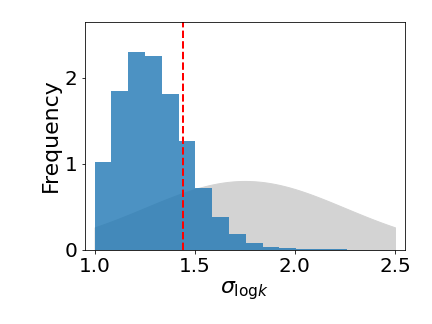}}
\hspace{2mm}
\subfloat[$\log_{10}(a_r)$]{\label{fig:d}\includegraphics[width = 50mm]{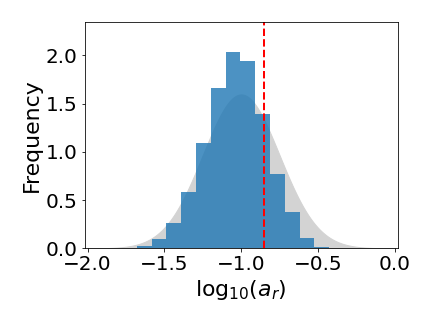}}\\
\subfloat[Parameter $d$]{\label{fig:d}\includegraphics[width = 50mm]{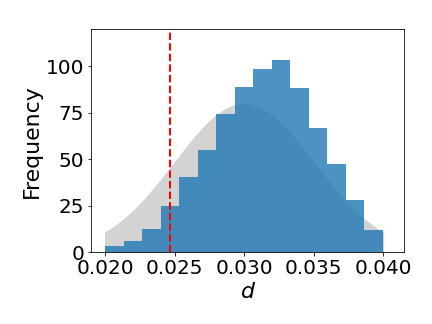}}
\hspace{5mm}
\subfloat[Parameter $e$]{\label{fig:e}\includegraphics[width = 50mm]{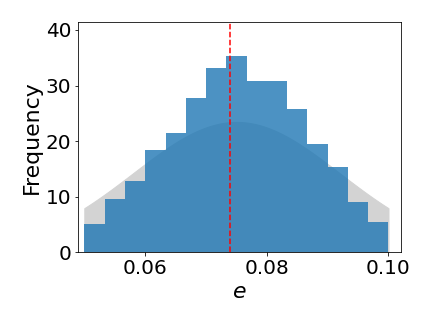}}\\
\subfloat[$\mu_{\phi}$]{\label{fig:d}\includegraphics[width = 50mm]{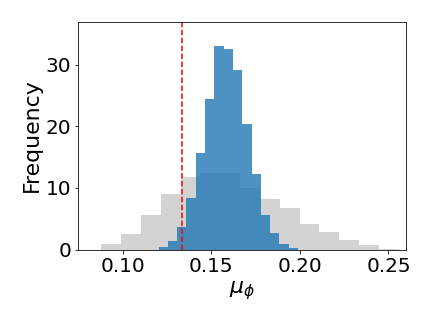}}
\hspace{5mm}
\subfloat[$\sigma_{\phi}$]{\label{fig:e}\includegraphics[width = 50mm]{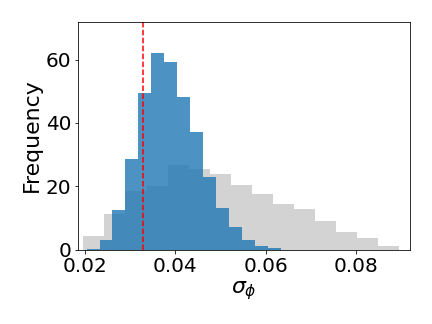}}
\caption{History matching results for the metaparameters for true model~2. Prior distributions of metaparameters are Gaussian in this case. Gray regions represent prior distributions, blue histograms are posterior distributions, and red vertical lines denote true values. Legend in (a) applies to all subplots.}
\label{meta_2_prior}
\end{figure}

The number of hierarchical MCMC iterations and the acceptance rates for these cases are similar to those for the uniform-prior cases. Specifically, to achieve convergence with Gaussian priors, true model~1 requires 58,598 iterations (with an acceptance rate of about 20\%) and true model~2 requires 76,936 iterations (acceptance rate of about 19\%). The resulting posterior distributions for the metaparameters, for both true models, are shown in Figs.~\ref{meta_1_prior} and~\ref{meta_2_prior}. These results are quite similar to those using uniform-distributed priors. When the true value of the metaparameter is well away from the prior mean, we can observe less uncertainty reduction than with the uniform priors (e.g., compare Fig.~14c in the main text to Fig.~\ref{meta_1_prior}c in SI). The opposite may be seen when the true value of the metaparameter is close to the prior mean (e.g., compare SI Figs.~\ref{meta_2}c and~\ref{meta_2_prior}c). In any event, it is clear from the results in this paper that the hierarchical MCMC-based method is able to treat different prior distributions.

\bibliographystyle{Main} 
\bibliography{ref}

\begin{thebibliography}{44}
\expandafter\ifx\csname natexlab\endcsname\relax\def\natexlab#1{#1}\fi
\providecommand{\url}[1]{\texttt{#1}}
\providecommand{\href}[2]{#2}
\providecommand{\path}[1]{#1}
\providecommand{\DOIprefix}{doi:}
\providecommand{\ArXivprefix}{arXiv:}
\providecommand{\URLprefix}{URL: }
\providecommand{\Pubmedprefix}{pmid:}
\providecommand{\doi}[1]{\href{http://dx.doi.org/#1}{\path{#1}}}
\providecommand{\Pubmed}[1]{\href{pmid:#1}{\path{#1}}}
\providecommand{\bibinfo}[2]{#2}
\ifx\xfnm\relax \def\xfnm[#1]{\unskip,\space#1}\fi
\bibitem[{Andrieu and Thoms(2008)}]{andrieu2008tutorial}
\bibinfo{author}{Andrieu, C.}, \bibinfo{author}{Thoms, J.}, \bibinfo{year}{2008}.
\newblock \bibinfo{title}{{A tutorial on adaptive MCMC}}.
\newblock \bibinfo{journal}{Stat. Comput.} \bibinfo{volume}{18}, \bibinfo{pages}{343--373}.
\bibitem[{Bui et~al.(2021)Bui, Hamon, Castelletto, Osei-Kuffuor, Settgast and White}]{bui2021multigrid}
\bibinfo{author}{Bui, Q.M.}, \bibinfo{author}{Hamon, F.P.}, \bibinfo{author}{Castelletto, N.}, \bibinfo{author}{Osei-Kuffuor, D.}, \bibinfo{author}{Settgast, R.R.}, \bibinfo{author}{White, J.A.}, \bibinfo{year}{2021}.
\newblock \bibinfo{title}{{Multigrid reduction preconditioning framework for coupled processes in porous and fractured media}}.
\newblock \bibinfo{journal}{Comput. Methods Appl. Mech. Eng.} \bibinfo{volume}{387}, \bibinfo{pages}{114111}.
\bibitem[{Chen et~al.(2020)Chen, Harp, Lu and Pawar}]{chen2020reducing}
\bibinfo{author}{Chen, B.}, \bibinfo{author}{Harp, D.R.}, \bibinfo{author}{Lu, Z.}, \bibinfo{author}{Pawar, R.J.}, \bibinfo{year}{2020}.
\newblock \bibinfo{title}{{Reducing uncertainty in geologic CO$_2$ sequestration risk assessment by assimilating monitoring data}}.
\newblock \bibinfo{journal}{Int. J. Greenh. Gas Control.} \bibinfo{volume}{94}, \bibinfo{pages}{102926}.
\bibitem[{Chen et~al.(2018)Chen, Dunlop, Papaspiliopoulos and Stuart}]{chen2018dimension}
\bibinfo{author}{Chen, V.}, \bibinfo{author}{Dunlop, M.M.}, \bibinfo{author}{Papaspiliopoulos, O.}, \bibinfo{author}{Stuart, A.M.}, \bibinfo{year}{2018}.
\newblock \bibinfo{title}{{Dimension-robust MCMC in Bayesian inverse problems}}.
\newblock \bibinfo{journal}{arXiv preprint} \bibinfo{volume}{arXiv:1803.03344}.
\bibitem[{Fang et~al.(2020)Fang, Fang and Demanet}]{fang2020deep}
\bibinfo{author}{Fang, Z.}, \bibinfo{author}{Fang, H.}, \bibinfo{author}{Demanet, L.}, \bibinfo{year}{2020}.
\newblock \bibinfo{title}{{Deep generator priors for Bayesian seismic inversion}}.
\newblock \bibinfo{journal}{Adv. Geophys.} \bibinfo{volume}{61}, \bibinfo{pages}{179--216}.
\bibitem[{Gamerman and Lopes(2006)}]{gamerman2006markov}
\bibinfo{author}{Gamerman, D.}, \bibinfo{author}{Lopes, H.F.}, \bibinfo{year}{2006}.
\newblock \bibinfo{title}{{Markov chain Monte Carlo: stochastic simulation for Bayesian inference}}.
\newblock \bibinfo{publisher}{CRC Press}.
\bibitem[{Gelman et~al.(1997)Gelman, Gilks and Roberts}]{gelman1997weak}
\bibinfo{author}{Gelman, A.}, \bibinfo{author}{Gilks, W.R.}, \bibinfo{author}{Roberts, G.O.}, \bibinfo{year}{1997}.
\newblock \bibinfo{title}{{Weak convergence and optimal scaling of random walk Metropolis algorithms}}.
\newblock \bibinfo{journal}{Ann. Appl. Probab.} \bibinfo{volume}{7}, \bibinfo{pages}{110--120}.
\bibitem[{Gelman et~al.(1996)Gelman, Roberts, Gilks et~al.}]{gelman1996efficient}
\bibinfo{author}{Gelman, A.}, \bibinfo{author}{Roberts, G.O.}, \bibinfo{author}{Gilks, W.R.}, et~al., \bibinfo{year}{1996}.
\newblock \bibinfo{title}{{Efficient Metropolis jumping rules}}.
\newblock \bibinfo{journal}{Bayesian Stat.} \bibinfo{volume}{5}, \bibinfo{pages}{42}.
\bibitem[{Gonz{\'a}lez-Nicol{\'a}s et~al.(2015)Gonz{\'a}lez-Nicol{\'a}s, Ba{\`u} and Alzraiee}]{gonzalez2015detection}
\bibinfo{author}{Gonz{\'a}lez-Nicol{\'a}s, A.}, \bibinfo{author}{Ba{\`u}, D.}, \bibinfo{author}{Alzraiee, A.}, \bibinfo{year}{2015}.
\newblock \bibinfo{title}{Detection of potential leakage pathways from geological carbon storage by fluid pressure data assimilation}.
\newblock \bibinfo{journal}{Adv. Water Resour.} \bibinfo{volume}{86}, \bibinfo{pages}{366--384}.
\bibitem[{Hastings(1970)}]{hastings1970monte}
\bibinfo{author}{Hastings, W.K.}, \bibinfo{year}{1970}.
\newblock \bibinfo{title}{{Monte Carlo sampling methods using Markov chains and their applications}}.
\newblock \bibinfo{journal}{Biometrika.} \bibinfo{volume}{57}, \bibinfo{pages}{97--109}.
\bibitem[{Jahandideh et~al.(2021)Jahandideh, Hakim-Elahi and Jafarpour}]{jahandideh2021inference}
\bibinfo{author}{Jahandideh, A.}, \bibinfo{author}{Hakim-Elahi, S.}, \bibinfo{author}{Jafarpour, B.}, \bibinfo{year}{2021}.
\newblock \bibinfo{title}{{Inference of rock flow and mechanical properties from injection-induced microseismic events during geologic CO$_2$ storage}}.
\newblock \bibinfo{journal}{Int. J. Greenh. Gas Control.} \bibinfo{volume}{105}, \bibinfo{pages}{103206}.
\bibitem[{Jiang and Durlofsky(2023a)}]{jiang2023history}
\bibinfo{author}{Jiang, S.}, \bibinfo{author}{Durlofsky, L.J.}, \bibinfo{year}{2023}a.
\newblock \bibinfo{title}{History matching for geological carbon storage using data-space inversion with spatio-temporal data parameterization}.
\newblock \bibinfo{journal}{arXiv preprint} \bibinfo{volume}{arXiv:2310.03228}.
\bibitem[{Jiang and Durlofsky(2023b)}]{jiang2023use}
\bibinfo{author}{Jiang, S.}, \bibinfo{author}{Durlofsky, L.J.}, \bibinfo{year}{2023}b.
\newblock \bibinfo{title}{Use of multifidelity training data and transfer learning for efficient construction of subsurface flow surrogate models}.
\newblock \bibinfo{journal}{J. Comput. Phys.} \bibinfo{volume}{474}, \bibinfo{pages}{111800}.
\bibitem[{Jung et~al.(2015)Jung, Zhou and Birkholzer}]{jung2015detection}
\bibinfo{author}{Jung, Y.}, \bibinfo{author}{Zhou, Q.}, \bibinfo{author}{Birkholzer, J.T.}, \bibinfo{year}{2015}.
\newblock \bibinfo{title}{{On the detection of leakage pathways in geological CO$_2$ storage systems using pressure monitoring data: Impact of model parameter uncertainties}}.
\newblock \bibinfo{journal}{Adv. Water Resour.} \bibinfo{volume}{84}, \bibinfo{pages}{112--124}.
\bibitem[{Krevor et~al.(2012)Krevor, Pini, Zuo and Benson}]{krevor2012}
\bibinfo{author}{Krevor, S.C.}, \bibinfo{author}{Pini, R.}, \bibinfo{author}{Zuo, L.}, \bibinfo{author}{Benson, S.M.}, \bibinfo{year}{2012}.
\newblock \bibinfo{title}{{Relative permeability and trapping of CO$_2$ and water in sandstone rocks at reservoir conditions}}.
\newblock \bibinfo{journal}{Water Resour. Res.} \bibinfo{volume}{48}, \bibinfo{pages}{W02532}.
\bibitem[{Kruschke(2014)}]{kruschke2015markov}
\bibinfo{author}{Kruschke, J.}, \bibinfo{year}{2014}.
\newblock \bibinfo{title}{Doing Bayesian data analysis: A tutorial with R, JAGS, and Stan}.
\newblock \bibinfo{publisher}{Academic Press}.
\bibitem[{Li and Laloui(2016)}]{li2016coupled}
\bibinfo{author}{Li, C.}, \bibinfo{author}{Laloui, L.}, \bibinfo{year}{2016}.
\newblock \bibinfo{title}{{Coupled multiphase thermo-hydro-mechanical analysis of supercritical CO$_2$ injection: Benchmark for the In Salah surface uplift problem}}.
\newblock \bibinfo{journal}{Int. J. Greenh. Gas Control.} \bibinfo{volume}{51}, \bibinfo{pages}{394--408}.
\bibitem[{Liu and Grana(2020)}]{liu2020petrophysical}
\bibinfo{author}{Liu, M.}, \bibinfo{author}{Grana, D.}, \bibinfo{year}{2020}.
\newblock \bibinfo{title}{{Petrophysical characterization of deep saline aquifers for CO$_2$ storage using ensemble smoother and deep convolutional autoencoder}}.
\newblock \bibinfo{journal}{Adv. Water Resour.} \bibinfo{volume}{142}, \bibinfo{pages}{103634}.
\bibitem[{Liu and Durlofsky(2021)}]{liu20213d}
\bibinfo{author}{Liu, Y.}, \bibinfo{author}{Durlofsky, L.J.}, \bibinfo{year}{2021}.
\newblock \bibinfo{title}{{3D CNN-PCA: A deep-learning-based parameterization for complex geomodels}}.
\newblock \bibinfo{journal}{Comput. Geosci.} \bibinfo{volume}{148}, \bibinfo{pages}{104676}.
\bibitem[{Malinverno and Briggs(2004)}]{malinverno2004expanded}
\bibinfo{author}{Malinverno, A.}, \bibinfo{author}{Briggs, V.A.}, \bibinfo{year}{2004}.
\newblock \bibinfo{title}{{Expanded uncertainty quantification in inverse problems: Hierarchical Bayes and empirical Bayes}}.
\newblock \bibinfo{journal}{Geophysics.} \bibinfo{volume}{69}, \bibinfo{pages}{1005--1016}.
\bibitem[{Mo et~al.(2019)Mo, Zhu, Zabaras, Shi and Wu}]{mo2019deep}
\bibinfo{author}{Mo, S.}, \bibinfo{author}{Zhu, Y.}, \bibinfo{author}{Zabaras, N.}, \bibinfo{author}{Shi, X.}, \bibinfo{author}{Wu, J.}, \bibinfo{year}{2019}.
\newblock \bibinfo{title}{Deep convolutional encoder-decoder networks for uncertainty quantification of dynamic multiphase flow in heterogeneous media}.
\newblock \bibinfo{journal}{Water Resour. Res.} \bibinfo{volume}{55}, \bibinfo{pages}{703--728}.
\bibitem[{Nicolaidou et~al.(2022)Nicolaidou, Birchwood, Prioul and Rodriguez-Herrera}]{ARMA-2022-0774}
\bibinfo{author}{Nicolaidou, E.}, \bibinfo{author}{Birchwood, R.A.}, \bibinfo{author}{Prioul, R.}, \bibinfo{author}{Rodriguez-Herrera, A.}, \bibinfo{year}{2022}.
\newblock \bibinfo{title}{{Stochastic inversion of wellbore stability models calibrated with hard and soft data}}, in: \bibinfo{booktitle}{U.S. Rock Mechanics/Geomechanics Symposium}, \bibinfo{organization}{American Rock Mechanics Association}.
\bibitem[{Remy et~al.(2009)Remy, Boucher and Wu}]{remy2009applied}
\bibinfo{author}{Remy, N.}, \bibinfo{author}{Boucher, A.}, \bibinfo{author}{Wu, J.}, \bibinfo{year}{2009}.
\newblock \bibinfo{title}{{Applied geostatistics with SGeMS: A user's guide}}.
\newblock \bibinfo{publisher}{Cambridge University Press}.
\bibitem[{Saadatpoor et~al.(2010)Saadatpoor, Bryant and Sepehrnoori}]{saadatpoor2010new}
\bibinfo{author}{Saadatpoor, E.}, \bibinfo{author}{Bryant, S.L.}, \bibinfo{author}{Sepehrnoori, K.}, \bibinfo{year}{2010}.
\newblock \bibinfo{title}{{New trapping mechanism in carbon sequestration}}.
\newblock \bibinfo{journal}{Transp. Porous Media.} \bibinfo{volume}{82}, \bibinfo{pages}{3--17}.
\bibitem[{Saltelli et~al.(2010)Saltelli, Annoni, Azzini, Campolongo, Ratto and Tarantola}]{saltelli2010variance}
\bibinfo{author}{Saltelli, A.}, \bibinfo{author}{Annoni, P.}, \bibinfo{author}{Azzini, I.}, \bibinfo{author}{Campolongo, F.}, \bibinfo{author}{Ratto, M.}, \bibinfo{author}{Tarantola, S.}, \bibinfo{year}{2010}.
\newblock \bibinfo{title}{{Variance based sensitivity analysis of model output. Design and estimator for the total sensitivity index}}.
\newblock \bibinfo{journal}{Comput. Phys. Commun.} \bibinfo{volume}{181}, \bibinfo{pages}{259--270}.
\bibitem[{Sobol(2001)}]{sobol2001global}
\bibinfo{author}{Sobol, I.M.}, \bibinfo{year}{2001}.
\newblock \bibinfo{title}{{Global sensitivity indices for nonlinear mathematical models and their Monte Carlo estimates}}.
\newblock \bibinfo{journal}{Math. Comput. Simul.} \bibinfo{volume}{55}, \bibinfo{pages}{271--280}.
\bibitem[{Sun and Durlofsky(2019)}]{sun2019data}
\bibinfo{author}{Sun, W.}, \bibinfo{author}{Durlofsky, L.J.}, \bibinfo{year}{2019}.
\newblock \bibinfo{title}{{Data-space approaches for uncertainty quantification of CO$_2$ plume location in geological carbon storage}}.
\newblock \bibinfo{journal}{Adv. Water Resour.} \bibinfo{volume}{123}, \bibinfo{pages}{234--255}.
\bibitem[{Tang et~al.(2022b)Tang, Fu, Jo, Jiang, Sherman, Hamon, Azzolina and Morris}]{2022deep}
\bibinfo{author}{Tang, H.}, \bibinfo{author}{Fu, P.}, \bibinfo{author}{Jo, H.}, \bibinfo{author}{Jiang, S.}, \bibinfo{author}{Sherman, C.S.}, \bibinfo{author}{Hamon, F.}, \bibinfo{author}{Azzolina, N.A.}, \bibinfo{author}{Morris, J.P.}, \bibinfo{year}{2022b}.
\newblock \bibinfo{title}{{Deep learning-accelerated 3D carbon storage reservoir pressure forecasting based on data assimilation using surface displacement from InSAR}}.
\newblock \bibinfo{journal}{Int. J. Greenh. Gas Control.} \bibinfo{volume}{120}, \bibinfo{pages}{103765}.
\bibitem[{Tang et~al.(2022a)Tang, Ju and Durlofsky}]{tang2022deep}
\bibinfo{author}{Tang, M.}, \bibinfo{author}{Ju, X.}, \bibinfo{author}{Durlofsky, L.J.}, \bibinfo{year}{2022a}.
\newblock \bibinfo{title}{{Deep-learning-based coupled flow-geomechanics surrogate model for CO$_2$ sequestration}}.
\newblock \bibinfo{journal}{Int. J. Greenh. Gas Control.} \bibinfo{volume}{118}, \bibinfo{pages}{103692}.
\bibitem[{Tang et~al.(2020)Tang, Liu and Durlofsky}]{tang2020deep}
\bibinfo{author}{Tang, M.}, \bibinfo{author}{Liu, Y.}, \bibinfo{author}{Durlofsky, L.J.}, \bibinfo{year}{2020}.
\newblock \bibinfo{title}{A deep-learning-based surrogate model for data assimilation in dynamic subsurface flow problems}.
\newblock \bibinfo{journal}{J. Comput. Phys.} \bibinfo{volume}{413}, \bibinfo{pages}{109456}.
\bibitem[{Tang et~al.(2021)Tang, Liu and Durlofsky}]{tang2021deep}
\bibinfo{author}{Tang, M.}, \bibinfo{author}{Liu, Y.}, \bibinfo{author}{Durlofsky, L.J.}, \bibinfo{year}{2021}.
\newblock \bibinfo{title}{{Deep-learning-based surrogate flow modeling and geological parameterization for data assimilation in 3D subsurface flow}}.
\newblock \bibinfo{journal}{Comput. Methods Appl. Mech. Eng.} \bibinfo{volume}{376}, \bibinfo{pages}{113636}.
\bibitem[{Tavakoli et~al.(2013)Tavakoli, Yoon, Delshad, ElSheikh, Wheeler and Arnold}]{tavakoli2013comparison}
\bibinfo{author}{Tavakoli, R.}, \bibinfo{author}{Yoon, H.}, \bibinfo{author}{Delshad, M.}, \bibinfo{author}{ElSheikh, A.H.}, \bibinfo{author}{Wheeler, M.F.}, \bibinfo{author}{Arnold, B.W.}, \bibinfo{year}{2013}.
\newblock \bibinfo{title}{{Comparison of ensemble filtering algorithms and null-space Monte Carlo for parameter estimation and uncertainty quantification using CO$_2$ sequestration data}}.
\newblock \bibinfo{journal}{Water Resour. Res.} \bibinfo{volume}{49}, \bibinfo{pages}{8108--8127}.
\bibitem[{Wang et~al.(2021)Wang, Chang and Zhang}]{wang2021efficient}
\bibinfo{author}{Wang, N.}, \bibinfo{author}{Chang, H.}, \bibinfo{author}{Zhang, D.}, \bibinfo{year}{2021}.
\newblock \bibinfo{title}{Efficient uncertainty quantification for dynamic subsurface flow with surrogate by theory-guided neural network}.
\newblock \bibinfo{journal}{Comput. Methods Appl. Mech. Eng.} \bibinfo{volume}{373}, \bibinfo{pages}{113492}.
\bibitem[{Wen et~al.(2021)Wen, Hay and Benson}]{wen2021ccsnet}
\bibinfo{author}{Wen, G.}, \bibinfo{author}{Hay, C.}, \bibinfo{author}{Benson, S.M.}, \bibinfo{year}{2021}.
\newblock \bibinfo{title}{{CCSNet: a deep learning modeling suite for CO$_2$ storage}}.
\newblock \bibinfo{journal}{Adv. Water Resour.} \bibinfo{volume}{155}, \bibinfo{pages}{104009}.
\bibitem[{Wen et~al.(2022)Wen, Li, Azizzadenesheli, Anandkumar and Benson}]{wen2022u}
\bibinfo{author}{Wen, G.}, \bibinfo{author}{Li, Z.}, \bibinfo{author}{Azizzadenesheli, K.}, \bibinfo{author}{Anandkumar, A.}, \bibinfo{author}{Benson, S.M.}, \bibinfo{year}{2022}.
\newblock \bibinfo{title}{{U-FNO—An enhanced Fourier neural operator-based deep-learning model for multiphase flow}}.
\newblock \bibinfo{journal}{Adv. Water Resour.} \bibinfo{volume}{163}, \bibinfo{pages}{104180}.
\bibitem[{Wen et~al.(2023)Wen, Li, Long, Azizzadenesheli, Anandkumar and Benson}]{wen2023real}
\bibinfo{author}{Wen, G.}, \bibinfo{author}{Li, Z.}, \bibinfo{author}{Long, Q.}, \bibinfo{author}{Azizzadenesheli, K.}, \bibinfo{author}{Anandkumar, A.}, \bibinfo{author}{Benson, S.M.}, \bibinfo{year}{2023}.
\newblock \bibinfo{title}{{Real-time high-resolution CO$_2$ geological storage prediction using nested Fourier neural operators}}.
\newblock \bibinfo{journal}{Energy Environ. Sci.} \bibinfo{volume}{16}, \bibinfo{pages}{1732--1741}.
\bibitem[{White et~al.(2014)White, Chiaramonte, Ezzedine, Foxall, Hao, Ramirez and McNab}]{white2014geomechanical}
\bibinfo{author}{White, J.A.}, \bibinfo{author}{Chiaramonte, L.}, \bibinfo{author}{Ezzedine, S.}, \bibinfo{author}{Foxall, W.}, \bibinfo{author}{Hao, Y.}, \bibinfo{author}{Ramirez, A.}, \bibinfo{author}{McNab, W.}, \bibinfo{year}{2014}.
\newblock \bibinfo{title}{{Geomechanical behavior of the reservoir and caprock system at the In Salah CO$_2$ storage project}}.
\newblock \bibinfo{journal}{Proc. Natl. Acad. Sci.} \bibinfo{volume}{111}, \bibinfo{pages}{8747--8752}.
\bibitem[{Xiao et~al.(2022)Xiao, Zhang, Ma, Jin and Zhou}]{xiao2022model}
\bibinfo{author}{Xiao, C.}, \bibinfo{author}{Zhang, S.}, \bibinfo{author}{Ma, X.}, \bibinfo{author}{Jin, J.}, \bibinfo{author}{Zhou, T.}, \bibinfo{year}{2022}.
\newblock \bibinfo{title}{Model-reduced adjoint-based inversion using deep-learning: Example of geological carbon sequestration modeling}.
\newblock \bibinfo{journal}{Water Resour. Res.} \bibinfo{volume}{58}, \bibinfo{pages}{e2021WR031041}.
\bibitem[{Xiao et~al.(2021)Xiao, Xu, Reuschen, Nowak and Hendricks~Franssen}]{xiao2021bayesian}
\bibinfo{author}{Xiao, S.}, \bibinfo{author}{Xu, T.}, \bibinfo{author}{Reuschen, S.}, \bibinfo{author}{Nowak, W.}, \bibinfo{author}{Hendricks~Franssen, H.J.}, \bibinfo{year}{2021}.
\newblock \bibinfo{title}{{Bayesian inversion of multi-Gaussian log-conductivity fields with uncertain hyperparameters: An extension of preconditioned Crank-Nicolson Markov chain Monte Carlo with parallel tempering}}.
\newblock \bibinfo{journal}{Water Resour. Res.} \bibinfo{volume}{57}, \bibinfo{pages}{e2021WR030313}.
\bibitem[{Yan et~al.(2022a)Yan, Chen, Harp, Jia and Pawar}]{yan2022robust}
\bibinfo{author}{Yan, B.}, \bibinfo{author}{Chen, B.}, \bibinfo{author}{Harp, D.R.}, \bibinfo{author}{Jia, W.}, \bibinfo{author}{Pawar, R.J.}, \bibinfo{year}{2022a}.
\newblock \bibinfo{title}{{A robust deep learning workflow to predict multiphase flow behavior during geological CO$_2$ sequestration injection and post-injection periods}}.
\newblock \bibinfo{journal}{J. Hydrol.} \bibinfo{volume}{607}, \bibinfo{pages}{127542}.
\bibitem[{Yan et~al.(2022b)Yan, Harp, Chen, Hoteit and Pawar}]{yan2022gradient}
\bibinfo{author}{Yan, B.}, \bibinfo{author}{Harp, D.R.}, \bibinfo{author}{Chen, B.}, \bibinfo{author}{Hoteit, H.}, \bibinfo{author}{Pawar, R.J.}, \bibinfo{year}{2022b}.
\newblock \bibinfo{title}{A gradient-based deep neural network model for simulating multiphase flow in porous media}.
\newblock \bibinfo{journal}{J. Comput. Phys.} \bibinfo{volume}{463}, \bibinfo{pages}{111277}.
\bibitem[{Zhang et~al.(2015)Zhang, Trame, Lesko and Schmidt}]{zhang2015sobol}
\bibinfo{author}{Zhang, X.Y.}, \bibinfo{author}{Trame, M.N.}, \bibinfo{author}{Lesko, L.J.}, \bibinfo{author}{Schmidt, S.}, \bibinfo{year}{2015}.
\newblock \bibinfo{title}{Sobol sensitivity analysis: a tool to guide the development and evaluation of systems pharmacology models}.
\newblock \bibinfo{journal}{CPT: Pharmacomet. Syst. Pharmacol.} \bibinfo{volume}{4}, \bibinfo{pages}{69--79}.
\bibitem[{Zhong et~al.(2019)Zhong, Sun and Jeong}]{zhong2019predicting}
\bibinfo{author}{Zhong, Z.}, \bibinfo{author}{Sun, A.Y.}, \bibinfo{author}{Jeong, H.}, \bibinfo{year}{2019}.
\newblock \bibinfo{title}{{Predicting CO$_2$ plume migration in heterogeneous formations using conditional deep convolutional generative adversarial network}}.
\newblock \bibinfo{journal}{Water Resour. Res.} \bibinfo{volume}{55}, \bibinfo{pages}{5830--5851}.
\bibitem[{Zhou and Tartakovsky(2021)}]{zhou2021markov}
\bibinfo{author}{Zhou, Z.}, \bibinfo{author}{Tartakovsky, D.M.}, \bibinfo{year}{2021}.
\newblock \bibinfo{title}{{Markov chain Monte Carlo with neural network surrogates: application to contaminant source identification}}.
\newblock \bibinfo{journal}{Stoch. Environ. Res. Risk Assess.} \bibinfo{volume}{35}, \bibinfo{pages}{639--651}.

\end{thebibliography}

\end{document}